\documentclass[10pt,journal,compsoc]{IEEEtran}
\usepackage{amsmath,amssymb,amsfonts, amsthm}
\usepackage{float}  %设置图片浮动位置的宏包
\usepackage{subfigure}  %插入多图时用子图显示的宏包
\usepackage{graphicx}
\usepackage{textcomp}
\usepackage{xcolor}
\usepackage{caption}
\usepackage[ruled,linesnumbered]{algorithm2e}
\usepackage[noend]{algorithmic}
\usepackage{nccbbb}
\usepackage[export]{adjustbox}
\usepackage{booktabs}
\usepackage{multirow}

\newtheorem {definition}{Definition}

\usepackage{makecell}
\usepackage{comment}
\usepackage{ragged2e}
\usepackage{balance}
\usepackage{flushend}

\DeclareMathOperator*{\argmin}{argmin}

\newcommand{\qin}[1]{\textcolor{violet}{[qin:#1]}}
\newcommand{\ie}{\textit{i}.\textit{e}.}

\begin{document}
%\title{A Survey on Graph Learning with Distribution Shifts: Domain Adaptation, Out-of-distribution, and Continual Learning}
\title{Graph Learning under Distribution Shifts: A Comprehensive Survey on Domain Adaptation, Out-of-distribution, and Continual Learning}

\author{Man Wu, Xin Zheng, Qin Zhang, Xiao Shen, Xiong Luo,~\textit{Senior Member, IEEE}, Xingquan Zhu,~\textit{Fellow, IEEE}, Shirui Pan,~\textit{Senior Member, IEEE}
\IEEEcompsocitemizethanks{
\IEEEcompsocthanksitem 
Man Wu and Xiong Luo are with the School of Computer and Communication Engineering, University of Science and Technology Beijing, Beijing, China.
Email: \{man.wu,xluo\}@ustb.edu.cn.
\IEEEcompsocthanksitem Xin Zheng is with the Department of Software System and Cybersecurity (SSC), Monash University, Melbourne, Australia. E-mail: xin.zheng@monash.edu.
\IEEEcompsocthanksitem 
Qin Zhang is with the College of Computer Science and Software Engineering, Shenzhen University, Guangdong, China.
Email: qinzhang@szu.edu.cn.
\IEEEcompsocthanksitem
Xiao Shen is with the School of Computer Science and Technology, Hainan University, Haikou, China.
E-mail: xshen@hainanu.edu.cn.
\IEEEcompsocthanksitem Xingquan Zhu is with the Dept. of Electrical Engineering \& Computer Science, Florida Atlantic University, USA.
Email: xzhu3@fau.edu.
\IEEEcompsocthanksitem Shirui Pan is with the School of Information and Communication Technology and Institute for Integrated and Intelligent Systems (IIIS), Griffith University, Queensland, Australia. 
Email: s.pan@griffith.edu.au.
}}

% note the % following the last \IEEEmembership and also \thanks - 
% these prevent an unwanted space from occurring between the last author name
% and the end of the author line. i.e., if you had this:
% 
% \author{....lastname \thanks{...} \thanks{...} }
%                     ^------------^------------^----Do not want these spaces!
%
% a space would be appended to the last name and could cause every name on that
% line to be shifted left slightly. This is one of those "LaTeX things". For
% instance, "\textbf{A} \textbf{B}" will typeset as "A B" not "AB". To get
% "AB" then you have to do: "\textbf{A}\textbf{B}"
% \thanks is no different in this regard, so shield the last } of each \thanks
% that ends a line with a % and do not let a space in before the next \thanks.
% Spaces after \IEEEmembership other than the last one are OK (and needed) as
% you are supposed to have spaces between the names. For what it is worth,
% this is a minor point as most people would not even notice if the said evil
% space somehow managed to creep in.

% The paper headers
\markboth{Journal of \LaTeX\ Class Files,~Vol.~14, No.~8, August~2015}%
{Shell \MakeLowercase{\textit{et al.}}: Bare Demo of IEEEtran.cls for IEEE Journals}
% The only time the second header will appear is for the odd numbered pages
% after the title page when using the twoside option.
% 
% *** Note that you probably will NOT want to include the author's ***
% *** name in the headers of peer review papers.                   ***
% You can use \ifCLASSOPTIONpeerreview for conditional compilation here if
% you desire.

% If you want to put a publisher's ID mark on the page you can do it like
% this:
%\IEEEpubid{0000--0000/00\$00.00~\copyright~2015 IEEE}
% Remember, if you use this you must call \IEEEpubidadjcol in the second
% column for its text to clear the IEEEpubid mark.

% use for special paper notices
%\IEEEspecialpapernotice{(Invited Paper)}

\IEEEtitleabstractindextext{
\justifying
\begin{abstract}
Graph learning plays a pivotal role and has gained significant attention in various application scenarios, from social network analysis to recommendation systems, for its effectiveness in modeling complex data relations represented by graph structural data.
In reality, the real-world graph data typically show dynamics over time, with changing node attributes and edge structure, leading to the severe graph data distribution shift issue. 
This issue is compounded by the diverse and complex nature of distribution shifts, which can significantly impact the performance of graph learning methods in degraded generalization and adaptation capabilities, posing a substantial challenge to their effectiveness.
In this survey, we provide a comprehensive review and summary of the latest approaches, strategies, and insights that address distribution shifts within the context of graph learning.
Concretely, according to the \textit{observability} of distributions in the inference stage and the \textit{availability} of sufficient supervision information in the training stage, we categorize existing graph learning methods into several essential scenarios, including graph domain adaptation learning, graph out-of-distribution learning, and graph continual learning. 
For each scenario, a detailed taxonomy is proposed, with specific descriptions and discussions of existing progress made in distribution-shifted graph learning.
Additionally, we discuss the potential applications and future directions for graph learning under distribution shifts with a systematic analysis of the current state in this field.
The survey is positioned to provide general guidance for the development of effective graph learning algorithms in handling graph distribution shifts, and to stimulate future research and advancements in this area.
\end{abstract}
% Note that keywords are not normally used for peerreview papers.
}
% make the title area
\maketitle
%\IEEEtitleabstractindextext{

%}
% As a general rule, do not put math, special symbols or citations
% in the abstract or keywords.
% For peer review papers, you can put extra information on the cover
% page as needed:
% \ifCLASSOPTIONpeerreview
% \begin{center} \bfseries EDICS Category: 3-BBND \end{center}
% \fi
%
% For peerreview papers, this IEEEtran command inserts a page break and
% creates the second title. It will be ignored for other modes.
%\IEEEpeerreviewmaketitle

\section{Introduction}
\IEEEPARstart{G}{raph} structural data is ubiquitous in various real-world application domains, including social networks~\cite{Wang2020SGNN, Wei2019LRL,pilanci2020domain,dai2019network,corrabs190510095}, biological networks~\cite{wang2022test,chen2022graphtta,jin2022empowering,li2022ood}, road networks~\cite{you2020graph,li2022graphde,zheng2023spatio}, and computer networks~\cite{GPIL2022Tan,zhang2022hierarchical,mancini2019adagraph}. 
In these diverse domains, complex relationships among nodes, intricately woven through edges, harbor valuable information within entities, graph structures, and overarching graph data patterns.
In this case, graph learning~\cite{xia2021graph} techniques have arisen as responses to better analyzing and understanding various graph types, serving a wide range of graph-related tasks with promising inference abilities, covering drug discovery~\cite{jin2022empowering,ji2023drugood}, knowledge graph exploration~\cite{qiu2018deepinf,luo2023normalizing}, social network analysis~\cite{wu2023graph,zheng2022multi}, recommender systems~\cite{wang2017graph,jin2023dual}, and physical movement prediction\cite{yu2018spatio}, etc.

Despite the success of prevalent graph learning approaches, the presence of \textit{distribution shifts} in graph data poses a substantial constraint on the capabilities of current methods~\cite{li2022out}. 
This is due to the dynamic and evolving nature of real-world graph data. For example, social networks evolve with new users and shifting relationships over time, leading to significant variations on node features and edge connections, so that the performance of graph learning models would degrade on recommendation systems or trend analysis~\cite{pomeroy2020dynamics}.
%This dynamic nature can lead to distribution shifts, affecting the accuracy of models used for tasks like recommendation systems or trend analysis~\cite{pomeroy2020dynamics}. 
%
In financial networks, the relationships between entities (like stocks, commodities, or institutions) change due to market trends, economic policies, or global events~\cite{bardoscia2021physics, acemoglu2015systemic}. A graph learning model trained on historical market data might underperform when the underlying relationships between entities shift. In biology, networks representing interactions between proteins, genes, or species in an ecosystem can change due to mutations, environmental pressures, or disease outbreaks~\cite{wu2023molecular}. Models predicting disease spread or gene interactions must adapt to these shifts. 
In addition, transportation systems (like road networks, air traffic, or public transit systems) experience shifts due to factors like urban development, changes in travel patterns, or infrastructure modifications. These shifts impact models used for optimizing routes, predicting congestion, or planning new infrastructure~\cite{ercan2017public}.
These shifts in statistical distributions of graph data observed in nodes, edges, and different graphs considerably complicate the graph learning process, posing challenges for effective model deployment and application in real-world scenarios.

To this end, in this work, our focus is on engaging in \textbf{graph learning under distribution shifts},
%Therefore, these shifts arise when the statistical distribution of the graph data utilized during the training phase diverges from what is encountered in the testing phase. 
%This introduces further complexities in conducting \textbf{graph learning with distribution shifts}, 
specifically referring to the scenarios where the disparities of graph data probability distributions might encompass all aspects of graph components, including node features, graph structures, and label distributions. 
Consequently, graph learning models encounter difficulties in achieving precise generalization to previously unseen graph data distributions during the testing phase.

\begin{figure}[!t]
    \centering
\includegraphics[width=0.49\textwidth]{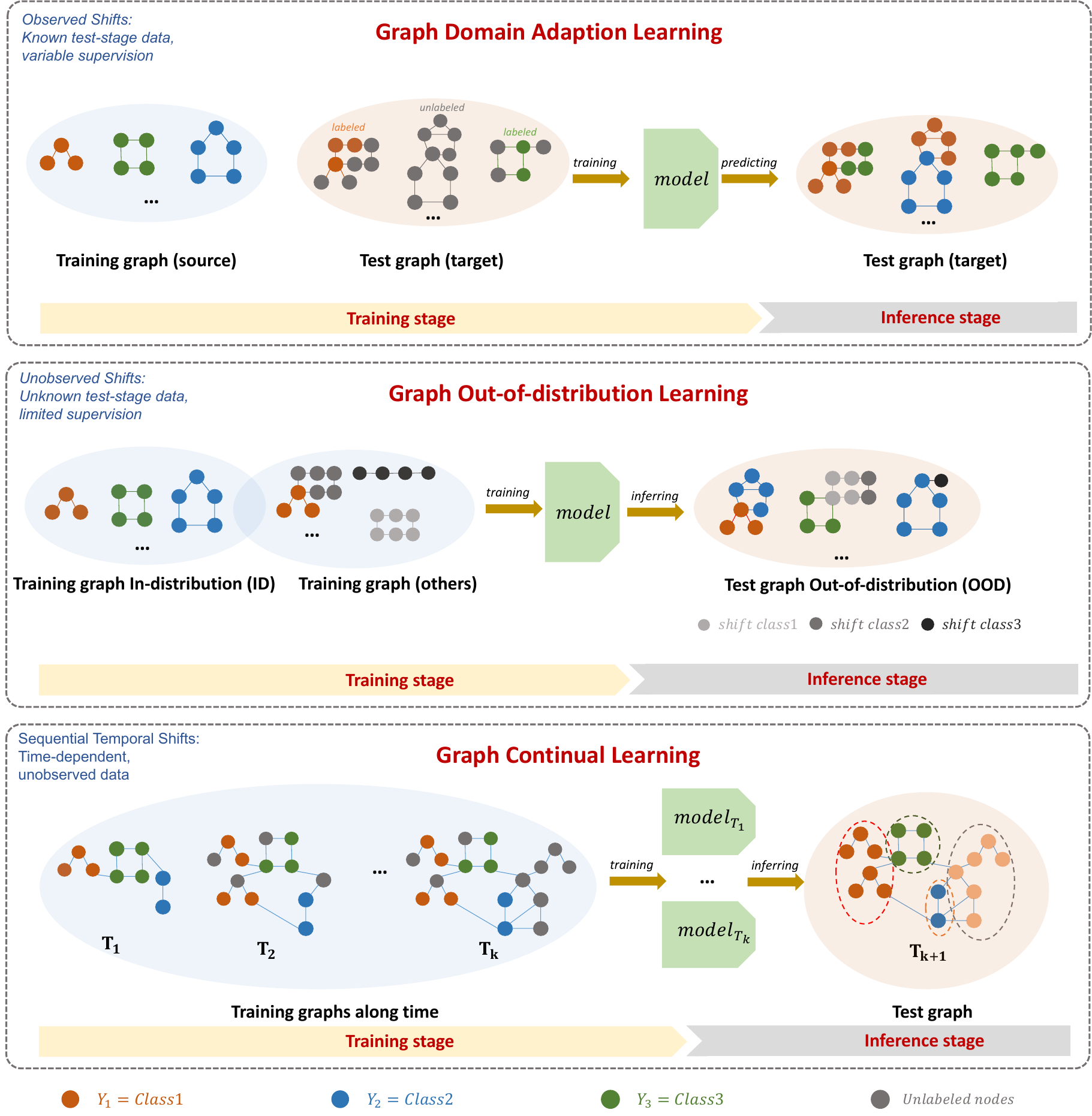}
    \caption{Overview of graph data distribution shifts and corresponding graph learning methods.}
    \label{fig:shiftall}
\end{figure}

In recent years, there has been a growing interest in exploring the paradigms of graph learning under distribution shifts~\cite{alam2018domain, pilanci2020domain, dai2019network, wu2019domain,li2022out, chen2022learning, li2022learning, baranwal2021graph,MALTONI201956, biesialska-etal-2020-continual, pmlrELLA2013, DBLP2016, Core502017}, to enable models to comprehend complex scenes, objects, and concepts in both static and dynamic scenarios for graph data.
Concretely, the graph data distribution shift scenarios can be categorized based on \textit{whether the test-stage distributions are observable} and \textit{whether the available supervision information is sufficient}.
Therefore, they can be classified into three main categories, as shown in Fig.~\ref{fig:shiftall}:
\begin{itemize}
\item \textbf{Observed Shifts: Known test-stage data, variable supervision}, where observed test-stage graph data that may exhibit potential unknown distribution shifts from the training stage, available supervisions could be either sufficient or insufficient
\item \textbf{Unobserved Shifts: Unknown test-stage data, limited supervision}, where unobserved test-stage graph data distributions with diverse and multiple shift types, available supervisions typically tend to be insufficient
\item \textbf{Sequential Temporal Shifts: Time-dependent, unobserved data}, where unobserved test-stage graph data distributions that arrive sequentially with dynamic temporal shifts over time.
\end{itemize}
%(1) observed test-stage graph data that may exhibit potential unknown distribution shifts from the training stage, available supervisions could be either sufficient or insufficient; (2) unobserved test-stage graph data distributions with diverse and multiple shift types, available supervisions typically tends to be insufficient; (3) unobserved test-stage graph data distributions that arrive sequentially with dynamic temporal shifts over time.}
Based on these three types of graph data distribution shift scenarios, in this work, we conduct a comprehensive review and provide a systematic taxonomy of existing graph learning methods that deal with the challenges of distribution shift learning in graph data. 
Concretely, current graph learning methods can also be classified into three categories, with each category corresponding to a particular distribution shift scenario:  

\begin{itemize}
    \item \textbf{Graph Domain Adaptation Learning}, aims to transfer graph learning models from a training (source) domain to a test (target) domain and require their proficient performance on the target domain with differing graph data distributions \cite{alam2018domain, das2018unsupervised, zhang2018structural, pilanci2020domain, wu2020unsupervised, dai2019network}. $\rightarrow$ \textit{Observed Shifts}.
    \item \textbf{Graph Out-of-distribution Learning}, aims to empower the graph learning models to effectively learn test graph data with different distributions from the training data and potential new classes unseen in training \cite{yang2021oodsurvey,li2022out}, enabling models' well generalization capabilities. $\rightarrow$ \textit{Unobserved Shifts}.
    \item \textbf{Graph Continual Learning}, aims to enable graph learning models to assimilate new information contingent upon changes in the evolving graph data distributions, while concurrently refining pre-existing knowledge and addressing nascent and previously unseen tasks~\cite{yuan2023continual, febrinanto2023graph}. $\rightarrow$ \textit{Sequential Temporal Shifts}.
\end{itemize}

Therefore, in this survey, our goal is to provide an in-depth understanding of graph data distribution shifts and delve into various graph learning models and approaches that tackle distribution shift challenges.
We will examine key concepts, critical challenges, previous limitations, and evaluation protocols associated with graph learning under distribution shifts. 
Moreover, we will discuss the potential real-world applications and highlight the promising future directions within this research area.
To the best of our knowledge, this is the first comprehensive survey on graph learning with general distribution shift scenarios.
Nevertheless, it is worth establishing connections between our work and several prior studies that concentrate on specific aspects of graph distribution shift: Li et al.~\cite{li2022out} reviewed the graph out-of-distribution generalization methods and Yang et al. \cite{yang2021oodsurvey} reviewed the graph out-of-distribution detection methods, which are two types of out-of-distribution tasks under the unobserved graph data shift scenario, respectively.
Liu et al.~\cite{liu2024generalization} reviewed the graph out-of-distribution adaptation methods under the observed/unobserved graph data shift scenario. 
Yuan et al.~\cite{yuan2023continual} and Febrinanto et al.~\cite{febrinanto2023graph} reviewed the graph continual learning methods and benchmarks under the sequential temporal graph data shift scenario.
In contrast, our emphasis is distinctly placed on a broader perspective of graph learning under distribution shifts. This encompasses a more extensive and fresh method review %, including publications up to 2023Q2, 
to furnish a more comprehensive research roadmap.

We envision that this survey will serve as an important resource for researchers and practitioners interested in graph learning under distribution shifts, provide insights into the latest developments and future research efforts, and facilitate informed decision-making in the areas of model selection, architectural design, and evaluation strategies for the continuous development of graph learning models, with potential benefits for both academic and industrial applications.

In summary, the core contributions of this survey work can be presented as follows:
\begin{itemize}
    \item \textbf{Broad-ranging Graph Distribution Shift Scenarios.} To the best of our knowledge, this is the first survey work in the field of graph learning covering broad-ranging graph distribution shift scenarios, facilitating in-depth understanding and analysis for dealing with diverse and complex graph data distribution shift cases.
    \item \textbf{Comprehensive Review and Taxonomy.} We provide a systematic taxonomy of existing graph learning advancements with various distribution shifts. This taxonomy includes three key categories: graph domain adaptation learning for observable shifts with known test-stage graphs, graph out-of-distribution learning for unobserved shifts with unknown test-stage graphs, as well as graph continual learning for sequential temporal shifts with time-evolving graphs.
    \item \textbf{Highlighted Practical Applications and Future Direction.} We
highlight the practical utility of graph learning in addressing distribution shifts, spanning from applications in scientific discovery to personalized daily life recommendations. Additionally, we identify several promising future research directions, intending to inspire and propel the progress of this research field.
\end{itemize}

\section{Problem Definition}
In this section, we outline the background of graph neural networks, list commonly used notations, and define graph related concepts.

Given a graph $G = (V, E)$ where $V = \{v_1, \cdots, v_{N} \}$ is the node set with $N$  nodes and $E=\{e_{i,j} | i,j = 1,\ldots,N\} \in V \times V$ is the edge set. The neighbor set of node $v$ is denoted as $\mathcal{N}(v) = \{u: (v, u) \in E  \}$. The node features are represented by a feature matrix $\mathbf{X} \in \mathbb{R}^{N \times d}$, where the $i$-th row $\mathbf{x}_i \in  \mathbb{R}^{d}$ is the feature vector of node $v_i$ and $d$ is the number of feature dimensions. The topological structure of $G$ is represented by the adjacency matrix $\mathbf{A}\in \{0,1\}^{n\times n}$, where $\mathbf{A}_{i,j}=1$ if the nodes $v_i$ and $v_j$ are connected, \ie, $\exists e_{i,j} \in E$, otherwise $\mathbf{A}_{i,j}= 0$. $\mathbf{Y} \in \mathbb{R}^{N \times C}$ is the node label matrix of $G$, where $C$ is the already-known classes used for training with $\mathcal{Y}= \{1,\cdots, C\}$. 
Therefore, a graph is composed of node features, topological structure, as well as node labels, denoted as $G = (\mathbf{X}, \mathbf{A}, \mathbf{Y})$.

Moreover, given a set of graphs with $M$ graphs, we have $\mathbb{G}=\{(G_1,y_1),\ldots, (G_M,y_M)\}$, and $\mathbb{Y} = \{y_i\in \mathbb{R}^{1 \times K}\}_{i=1}^{M}$ is the label set of the graph set. Here, each label $y_i$ corresponds to one of the $K$ classes to which the graph $G_i$ belongs. 
In this section, we primarily focus on illustrating the concepts of various graph distribution learning methodologies through the lens of node-level graph learning tasks. These principles can be readily extended to encompass edge-level and graph-level learning scenarios as well.

%If node $v_i \in V$ is associated with a label $c$, $y_{i,c} = 1$, otherwise, $y_{i,c} = 0$.

%{Consider a \textsl{graph} denoted  as $G=(V,E,X)$, where $V=\{v_i | {i = 1,\ldots,N}\}$ is a set of $N$ nodes in the graph. $E=\{e_{i,j} | i,j = 1,\ldots,N.$ $i\neq j \}$ is a set of  edges between pairs of nodes $v_i$ and $v_j$. $X\in \mathbb{R}^{N\times d}$ denotes the feature matrix of  nodes, and $d$ is the dimension of node features. $x_i \in X$ indicates the feature vector associated with each node $v_i$. The topological structure of $G$ is represented as  an adjacency matrix $A \in \mathbb{R}^{N \times N}$, where $A_{i,j}=1$ if the nodes $v_i$ and $v_j$ are connected,~\ie, $\exists e_{i,j} \in E$, otherwise $A_{i,j}= 0$. $Y \in \mathbb{R}^{N \times C}$ is the label matrix of $G$, where $C$ is the already-known classes used for training. If node $v_i \in V$ is associated with a label $c$, $y_{i,c} = 1$, otherwise, $y_{i,c} = 0$. }
% Let $\mathbb{X}$ be the input space and $\mathbb{Y}$ be the label space.
% A graph function $f_\theta:\mathbb{X}\rightarrow \mathbb{Y}$ with parameter $\theta$ maps the input instance $X\in \mathbb{X}$, i.e., node/link/graph for node-level/link-level/graph-level task, into the label $Y\in \mathbb{Y}$. 
% A loss function $\mathcal{L}$ measures the distance between prediction and ground-truth label. 

\begin{definition}[\textbf{Conventional Closed-set Graph Learning}]
Given a graph $G = (\mathbf{X}, \mathbf{A}, \mathbf{Y})$, a GNN model $f_{\boldsymbol{\theta}}(\cdot)$ takes the node attributes $\mathbf{X}$ and graph structure $\mathbf{A}$ as inputs.
%is composed of a feature encoder $f_{\boldsymbol{\theta}_g}(\cdot)$ for learning representative node embeddings $\mathbf{Z}$, and a classifier $f_{\boldsymbol{\theta}_c}(\cdot)$ for classifying the nodes into $C$ already-known classes, where $\boldsymbol{\theta}=\left[\boldsymbol{\theta}_g, \boldsymbol{\theta}_c\right]$. 
%We have $\mathbf{Z}=f_{\boldsymbol{\theta}_g}(\mathbf{X},\mathbf{A})$, and $\mathbf{\hat{Y}}=f_{\boldsymbol{\theta}_c}(\mathbf{Z})$ is the output prediction of the GNN model.
With the assumption that test data distribution $\mathcal{P}_{te}$ and train data distribution $\mathcal{P}_{tr}$ share the same feature space and label space, for typical \textsl{closed-set graph learning} problem, the feature encoder and the classifier of the GNN model are optimized to minimize the expected risk~\cite{yu2017open-holder38} as 
\begin{equation}
    \label{eq:closedrisk}
    f_{\boldsymbol{\theta}^{*}}= arg\min_{\boldsymbol{\theta}} \mathbb{E}_{G \sim \mathcal{P}_{tr}}[\mathcal{L} (f_{\boldsymbol{\theta}}(\mathbf{X},\mathbf{A}), \mathbf{Y})]
\end{equation}
where $f_{\boldsymbol{\theta}^{*}}$ is the optimal GNN model trained on the training graph that achieves the minimal loss. And the loss function $\mathcal{L} (\cdot)$ measures the discrepancy between the output prediction $\mathbf{\hat{Y}}=f_{\boldsymbol{\theta}}(\mathbf{X},\mathbf{A})$ and the ground-truth node labels $\mathbf{Y}$. Generally, $\mathcal{L} (\cdot)$ can be the cross-entropy function to discriminate between known classes.

In this case, on the test graph $G_{te} = (\mathbf{X}_{te},\mathbf{A}_{te}) \sim \mathcal{P}_{te}$, we have $\mathbf{\hat{Y}}_{te} = f_{\boldsymbol{\theta}^{*}}(\mathbf{X}_{te},\mathbf{A}_{te})$ for making inference of the test node labels.
\begin{comment}
For a typical \textsl{closed-set graph learning} problem, a GNN encoder $f_{\theta_g}$ takes node features $X$ and adjacency matrix $A$ as input, aggregates the neighborhood information and outputs representations.
Then, a classifier $f_{\theta_c}$ is used to classify the nodes into $C$ already-known classes.
The GNN encoder and the classifier are optimized to minimize the expected risk ~\cite{yu2017open-holder38} in Eq. \eqref{eq:closedrisk}, with the assumption that test data distribution $\mathcal{P}_{te}$ and train data distribution $\mathcal{P}_{tr}$ share the same feature space and label space, \ie,
\begin{equation}
    \label{eq:closedrisk}
    f^*_\theta = arg\min_{{f_\theta}} \mathbb{E}_{(X,Y) \sim P_{te}}[\mathcal{L} (f_\theta(X,A), Y)]
\end{equation}
% \begin{equation}
%     \label{eq:closedrisk}
%     f^*_\theta = arg\min_{f\in \mathcal{H}} \mathbb{E}_{(X,Y) \sim \mathcal{D}_{te}} \mathbb{I}(y\neq f(\theta_g, \theta_c; x,A))
% \end{equation}
where $f^*_\theta$ is an optimal graph function on the test data that can achieve the minimal loss, $\theta=(\theta_g, \theta_c)$.
%
The loss function $\mathcal{L} (f_\theta(X), Y)$ measures the discrepancy between the predicted value $f_\theta(X)$ and the actual value $Y$.
%%
% $\mathcal{H}$ is the hypothesis space, $\mathbb{I}(\cdot)$ is the indicator function which outputs 1 if the expression holds and 0 otherwise. 
Generally, it can be optimized with cross-entropy to discriminate between known classes. 
\end{comment}
\end{definition}
\begin{definition}[\textbf{Graph Domain Adaptation Learning}]
Given a graph $G_{src} = (\mathbf{X}_{src}, \mathbf{A}_{src}, \mathbf{Y}_{src}) \sim \mathcal{P}_{src}$ with sufficient labeled instances that are drawn from the source distribution $\mathcal{P}_{src}$ called the source domain, and a target graph $G_{tgt} = (\mathbf{X}_{tgt}, \mathbf{A}_{tgt}, \mathbf{Y}_{tgt}^{\dag}) \sim \mathcal{P}_{tgt}$ with a small number of labels (subscript$^\dag$ denotes partially observed labels) or no labels $\mathbf{Y}_{tgt}^{\dag}=\emptyset$ that are drawn from the target distribution $\mathcal{P}_{tgt}$ called the target domain, where $\mathcal{P}_{src}\neq \mathcal{P}_{tgt}$ but all nodes in ${G}_{src}$ and ${G}_{tgt}$ are constrained in the same $C$-classes as $\{\mathbf{Y}_{src}, \mathbf{Y}_{tgt}\} \in \mathcal{Y}= \{1,\cdots, C\}$.

The goal of graph domain adaptation is to learn an optimal graph learning model $f_{\boldsymbol{\theta}^{*}}$ with both the source and target domains, so that it can achieve the minimal loss on the target domain, and the marginal distributions satisfy:
\begin{equation}
\begin{aligned}
        f_{\boldsymbol{\theta}^{*}} = \argmin _{{\boldsymbol{\theta}}} \mathbb{E}_{G_{src}\sim \mathcal{P}_{src},G_{tgt}\sim \mathcal{P}_{tgt}}  [\mathcal{L}_{GDA} (f_{\boldsymbol{\theta}}(G_{src},G_{tgt})].
\end{aligned}
\end{equation}
In graph domain adaptation (GDA), both the source distribution $\mathcal{P}_{src}$ and the target distribution $\mathcal{P}_{tgt}$ are available during training. 
For different application scenarios of GDA research, the loss function $\mathcal{L}_{GDA}(\cdot)$ would show different formulations. 
For instance, in unsupervised GDA, the given target domain is totally unlabeled with $\mathbf{Y}_{tgt}^{\dag}=\emptyset$, $\mathcal{L}_{GDA}(\cdot)$ would be an unsupervised learning objective without using the target domain labels.
Moreover, in semi-supervised GDA, the given target domain contains a small number of labels with $\mathbf{Y}_{tgt}^{\dag} \neq \emptyset$, $\mathcal{L}_{GDA}(\cdot)$ could use partial target domain labels to optimize graph machine learning models.

\end{definition}
\begin{comment}
    
In graph domain adaptation (GDA), both the source distribution $\mathcal{P}_{src}$ and the target distribution $\mathcal{P}_{tgt}$ are available during training.
%
There are different settings in the research for GDA.
%
In unsupervised GDA, the given target dataset is totally unlabeled.
%
In semi-supervised GDA, the given target set contains a small number of labels.
%
Differently, the problem of graph out-of-distribution learning (G-OOD) is stronger than GDA. During the training of G-OOD, the test set is unavailable and the test set may contain new classes unseen in the training.
\end{comment}

\begin{figure}
    \centering
    \includegraphics[width=0.5\textwidth]{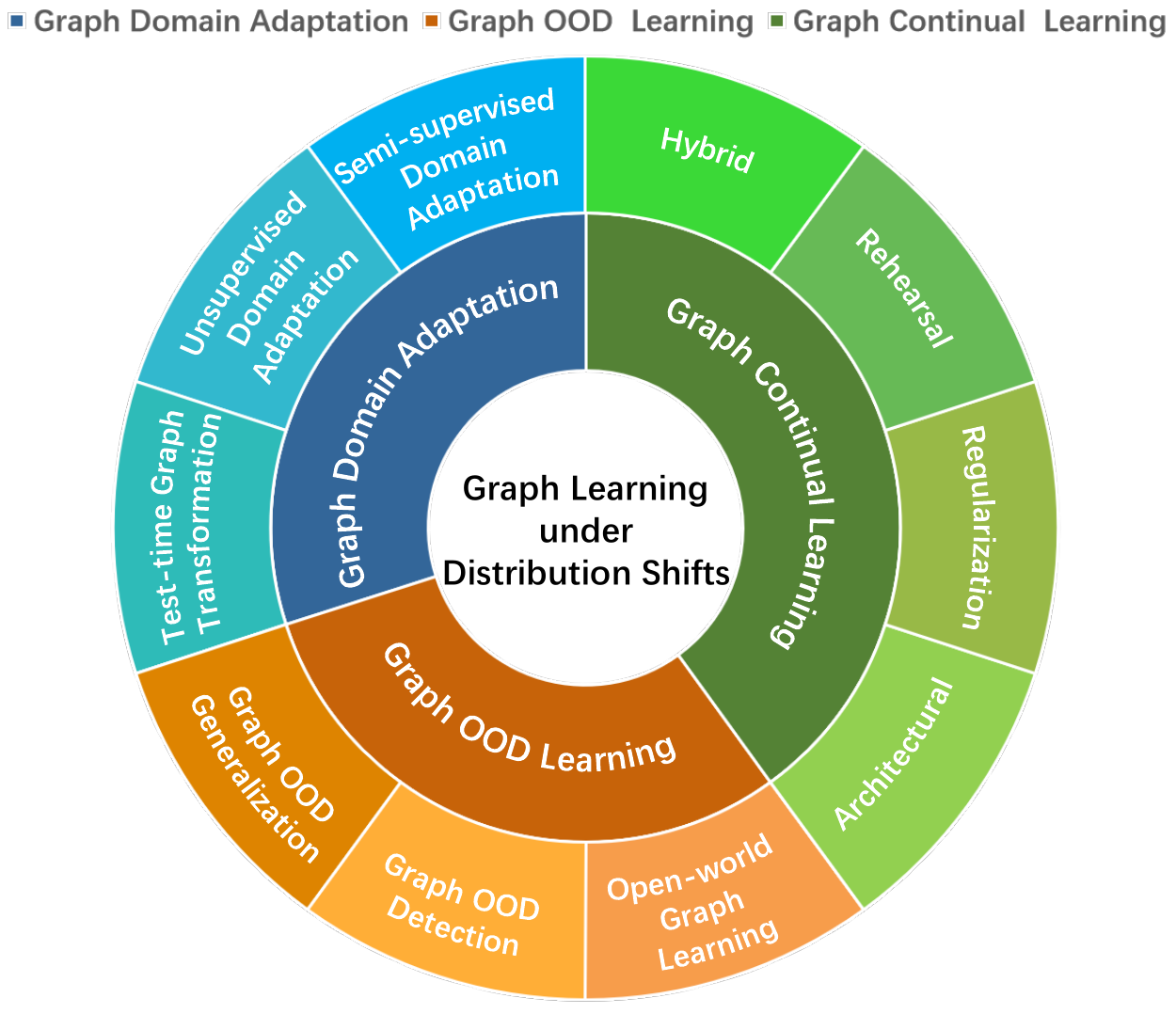}
    \caption{Overview of the general taxonomy of graph learning under distribution shifts.}
    \label{fig: overall}
\end{figure}

%The OOD graph learning problem is defined as:
\begin{definition}[\textbf{Graph Out-of-distribution Learning}]

Given a training graph $G_{tr} = (\mathbf{X}_{tr}, \mathbf{A}_{tr}, \mathbf{Y}_{tr})\sim \mathcal{P}_{tr}$ where $\mathbf{Y}_{tr} \in \mathcal{Y}_{tr}= \{1,\ldots, C\}$, and a test graph $G_{te} = (\mathbf{X}_{te}, \mathbf{A}_{te}, \mathbf{Y}_{te})\sim \mathcal{P}_{te}$ where $\mathbf{Y}_{te} \in \mathcal{Y}_{te}= \{1,\ldots, C, C+1, \ldots\}$, denoting there exists unseen node classes from $\{C+1, \ldots\}$ in the test graph, and $\mathcal{P}_{tr} \neq \mathcal{P}_{te}$, 
% graph out-of-distribution learning (G-OOD) 
\textbf{open-world graph learning} 
aims to learn  an optimal graph function $f_{\boldsymbol{\theta}^{*}}$ that can achieve the minimal loss on the training distribution $P_{tr}$ and also achieve optimal performance on data from the test distribution $P_{te}$ that may contain  new label space $\overline{\mathcal{Y}}= \{1,\ldots, C, unknown\}$, where
$\mathcal{Y}_{te} \supseteq \overline{\mathcal{Y}} \supset \mathcal{Y}_{tr}$
%$X\in\mathbb{X}, Y\in \mathbb{Y}_{te}\supset\mathbb{Y}_{tr}$, and $P_{te}\neq P_{tr}$, 
with the minimization of the expected risk \cite{yu2017open-holder38}:

% {\color{blue}graph out-of-distribution learning (G-OOD) aims to learn a $(C+1)$-$class$ classifier $f_{\overline{\boldsymbol{\theta}}_c}$ that maps the test graph nodes into the new label space $\overline{\mathcal{Y}}= \{1,\ldots, C, unknown\}$,} with the minimization of the expected risk \cite{yu2017open-holder38}:
% \begin{equation}
%     \label{eq:openrisk}
%     f_{\boldsymbol{\theta}^{*}} = arg\min_{
%     \boldsymbol{\theta}} \mathbb{E}_{(x,y) \sim \mathcal{\overline{D}}_{te}} \mathbb{I}(y\neq f(\theta_g, \overline{\theta}_c; x,\overline{A}))
% \end{equation}

\begin{equation}
    \label{eq:openrisk}
    f_{\boldsymbol{\theta}^{*}} = arg\min_{
    \boldsymbol{\theta}} \mathbb{E}_{G_{tr}\sim \mathcal{P}_{tr},G_{te}\sim \mathcal{P}_{te}}  [\mathcal{L}_{G-OOD} (f_{\boldsymbol{\theta}}(G_{tr})].
\end{equation}
Here, in open-world graph learning, only the training graph distribution $\mathcal{P}_{tr}$ is available during training, and the test distribution $\mathcal{P}_{te}$ is unavailable.
The predicted class $unknown\in\overline{\mathcal{Y}}$ contains a group of novel categories, which may contain more than one class. 

When only focusing on the detection of the unseen-class nodes, the problem degenerates into \textit{Graph OOD detection} problem, i.e.,  a binary-class problem to learn whether a test example is from seen classes $\mathcal{Y}_{tr}$ or not.
In addition, When $Y_{te} = Y_{tr}$, i.e. there is no unseen class during test, the model only concentrates on the feature distribution shift (i.e. the difference between the $\mathcal{P}_{te}$ and $\mathcal{P}_{tr}$), the problem  will degenerate to \textbf{graph OOD generalization} problem. 
We refer to \textbf{graph OOD generalization, graph OOD detection} and \textbf{open-world graph learning} collectively as \textbf{graph out-of-distribution (G-OOD) learning}. A more detailed taxonomy is illustrated in section \ref{subsec:g-ood} and Fig. \ref{fig: overall}.

% In addition, the predicted class $unknown\in\overline{\mathcal{Y}}$ contains a group of novel categories, which may contain more than one class. 
% Thus, the overall risk aims to classify known classes while also detecting the samples from unseen categories as class $unknown$.

% Given the training set $\mathcal{D} = \{(X_i,Y_i)\}_{i=1}^N$ of $N$ instances (i.e., nodes, links, or graphs) that are drawn from the training distribution $P_{tr}(X,Y)$, where $X_i\in\mathbb{X}$ and $Y_i\in \mathbb{Y}_{tr}\subset\mathbb{Y}$.
% %
% The goal of OOD graph learning is to learn an optimal graph function $f^*_\theta$ that can achieve the minimal loss on the training set and also achieve optimal performance on data from the test distribution $P_{te}(X,Y)$ that may contain new classes, where $X\in\mathbb{X}, Y\in \mathbb{Y}_{te}\supset\mathbb{Y}_{tr}$, and $P_{te}\neq P_{tr}$:
% \begin{equation}
%     f^*_\theta = \argmin _{f_\theta} \mathbb{E}_{(X,Y)\sim P_{tr}}[\mathcal L (f_\theta(X), Y)].
% \end{equation}
\end{definition}

%\qin{==Here, the definition of "out-of-distribution graph learning" seems to be the definition of "open-set graph learning". Few researchers use this expression "out-of-distribution graph learning". "out-of-distribution generalization"/"out-of-distribution detection"/"open-set graph learning"/"open-world graph learning" are more commonly used.  }

Both GDA and G-OOD aim to transfer the learned knowledge from a training set to a test set with a different data distribution.
%as $f_{\boldsymbol{\theta}^{*}}(G_{te}) = Y_{te}$ when $(X_{te},Y_{te})\sim P_{te}$.
%
Differently, graph continual learning (GCL) aims to adapt to the dynamics in real-world applications by continually training the graph learning model on a sequence of training sets with evolving data distributions.

\begin{definition}[\textbf{Graph Continual Learning}]
Given a sequence of training graphs $\mathcal{D}_{tr} = \{G^{(t)}_{tr}\}_{t=t_{0}}^{t_{k}}$ that are drawn from the training distribution $\mathcal{P}_{tr}$, where $t$ is the index of the training task at the $t$-th time step. And when $\forall t\neq t'$, we have $\mathcal{P}^{(t)}_{tr} \neq \mathcal{P}^{(t')}_{tr}$. 
In this case, the goal of graph continual learning (GCL) is to learn an optimal graph learning model $f_{\boldsymbol{\theta}^{*}}$ incrementally from the sequential training graph set. At the learning session $t$, only the training graph $G^{(t)}_{tr}\sim\mathcal{P}^{(t)}_{tr}$ can be accessed as:
\begin{equation}
    f_{\boldsymbol{\theta}^{(t)*}} = \argmin _{\boldsymbol{\theta}^{(t)}} \mathbb{E}_{G^{(t)}_{tr}\sim P^{(t)}_{tr}}[\mathcal{L}_{GCL} (f_{\boldsymbol{\theta}^{(t)}}(G^{(t)}_{tr}))].
\end{equation}
$f_{\boldsymbol{\theta}^{(t)*}}$ is expected to achieve the minimal loss on the data drawn from the test distribution $G_{te}^{(t')} \sim \mathcal{P}^{(t')}_{te}$ after any learning session $t'>=t$, as $f_{\boldsymbol{\theta}^{(t)*}} (G_{te}^{(t')}) = \mathbf{Y}_{te}^{(t')}$

\end{definition}
\noindent
%And after training, the optimal graph learning model is anticipated to perform well as $f_{\boldsymbol{\theta}^{*}} (G_{te}^{t'}) = Y_{te}^{t'}$ when $(X_{te}^{t'},Y_{te}^{t'})\sim P_{te}^{t}$ in the inference stage.

Overall, the summary of the information used by these three research problems in both the training and inference stages is presented in Table ~\ref{tab:inforsum}.
Specifically, graph domain adaptation learning (GDA) and graph out-of-distribution learning (G-OOD) are more concerned with the distribution shifts of static data. The main difference between them is that GDA deals with observed shifts, meaning it handles known test-stage data, whereas G-OOD tackles unobserved shifts, or unknown test-stage data.
In contrast, graph continual learning (GCL) targets the distribution shifts of dynamic data, focusing on sequential temporal shifts.
\begin{table}[!bt]
\centering
\renewcommand\arraystretch{1.5}
\caption{A summary of information used in graph domain adaption learning, graph out-of-distribution learning, and graph continual learning in both training and inference stages.}
\label{tab:inforsum}
\resizebox{0.48\textwidth}{!}{
\begin{tabular}{llll}
\toprule
                                   & Training                 & Inference          & Labels          \\\midrule
Graph Domain Adaption Learning     & $G_{src} (tr)$, $G_{tgt} (te)$ & $G_{tgt} (te)$       & $\mathbf{Y}_{tgt} \subseteq \mathbf{Y}_{src}$ \\
Graph Out-of-distribution Learning & $G_{tr}$                   & $G_{te}$              & $\mathbf{Y}_{tr} \subseteq \mathbf{Y}_{te}$ \\
Graph Continual Learning           & $G_{tr}^{(t)}|_{t=t_{0}}^{t_{k}}$ & $G_{te}^{(t')}|_{t'=t_{k+1}}^{t_{m}}$ & $\mathbf{Y}_{tr} \subseteq \mathbf{Y}_{te}$ \\\bottomrule
\end{tabular}
}
\end{table}
%\xin{It would be better if we summarize the difference between three categories with a table, be like, what information can we access in the training and inference, and the optimization process.}

%\qin{There is another concern. This article aims to do survey on "distribution shift learning on graphs", and we include three categories of works introduced above. However, these three categories of studies all focus on the inconsistency of the label space. Maybe we should also involve the works about  distribution shift w.r.t. feature space and the hybrid.}

\section{Categorization And Frameworks}

% {\color{red}Fig. \ref{fig:overall} summarizes the general taxonomy of interpretable cross-modal reasoning (I-CMR) and the related methods reviewed in this paper.
% %
% The taxonomy is inspired by the modalities and design details of the provided explanations for CMR.
% %
% To begin with, according to the modality of the explanations, methods for I-CMR can be classified into five categories: visual explanation, textual explanation, graph explanation, symbol explanation, and multimodal explanation.
% %
% We further consider the design details of different methods and classify them in a hierarchical manner for ease of understanding and comparative analysis.}

\begin{figure*}
    \centering
    \includegraphics[width=\textwidth]{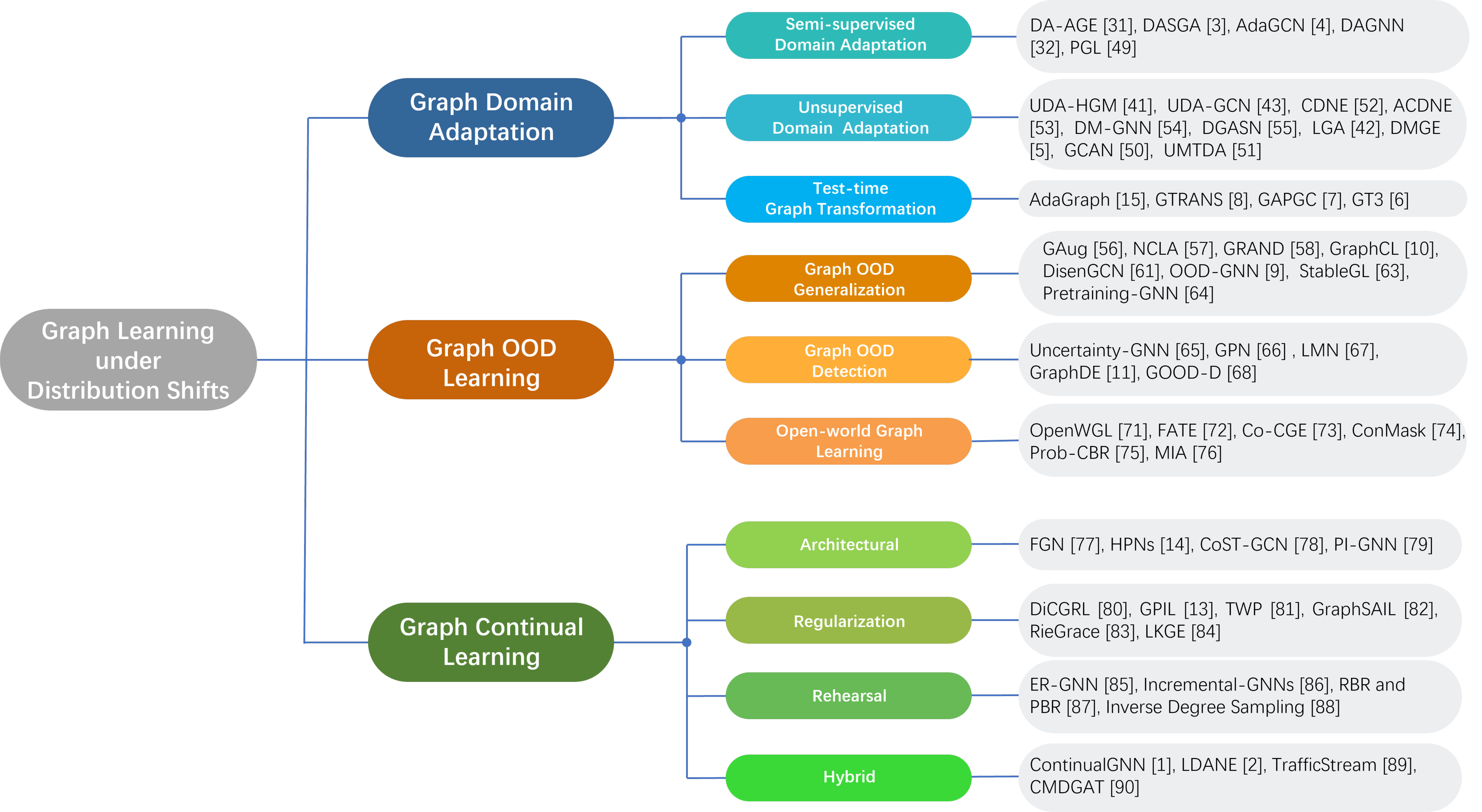}
    \caption{Hierarchical classification of graph learning under distribution shifts.}
    \label{fig:Hierarchical_classification}
\end{figure*}
Fig. \ref{fig:Hierarchical_classification} summarizes the general taxonomy of graph learning under distribution shifts and the related methods reviewed in this paper.
The taxonomy is centered on the shifting scenarios and design details of the graph learning algorithms under distribution shifts.
To begin with, according to the shifting scenarios of the explanations, methods for graph learning under distribution shifts can be classified into three categories: Graph domain adaptation, out-of-distribution graph learning and continual graph learning.
We further consider the design details of different methods and classify them in a hierarchical manner for ease of understanding and comparative analysis.

%In this section, we present our taxonomy of graph neural networks (GNNs), as shown in Table II. We categorize graph neural networks (GNNs) into recurrent graph neural networks (RecGNNs), convolutional graph neural networks (ConvGNNs), graph autoencoders (GAEs), and spatial-temporal graph neural networks (STGNNs). Figure 2 gives examples of various model architectures. In the following, we give a brief introduction to each category.

% \subsection{Taxonomy of Graph Neural Networks (GNNs)}
\subsection{Taxonomy of Graph Domain Adaptation Learning}
Domain adaptation addresses the issue of declining model performance due to differences in data distributions across different domains, aiming to enhance the model's generalization capability on the target domain. Domain adaptation methods can be further categorized into three types: semi-supervised, unsupervised, and test-time graph transformation.

\textbf{Semi-supervised Domain Adaptation.} These approaches focus on training the model using labeled data from the source domain as well as unlabeled data from both the source and target domains, aiming to ensure a strong performance in the target domain.

\textbf{Unsupervised Domain Adaptation.} These methods concentrate on model transfer without the availability of labeled target domain data. The core idea is to reduce the feature distribution differences between the source and target domains through feature alignment, thereby enhancing the model's generalization ability to the target domain.

\textbf{Test-time Graph Transformation (Adaptation).} These approaches are centered around adjusting the graph data of the target domain during testing to match the characteristics of the target domain, enhancing the model's adaptability to the target domain.

\subsection{Taxonomy of Graph Out-of-distribution Learning}\label{subsec:g-ood}
Out-of-distribution (OOD) graph learning addresses the challenge of learning from graphs with different features than those seen during training. OOD graph learning can be further categorized into three types: Graph Out-of-Distribution Generalization, Graph Out-of-Distribution Detection, and Open-world Graph Learning.

\textbf{Graph Out-of-distribution Generalization.} This type of OOD graph learning focuses on developing models that can generalize well to graphs with different features than the ones seen in training. The goal is to ensure that the model's performance remains satisfactory even when confronted with graphs from previously unseen distributions.

\textbf{Graph Out-of-distribution Detection.} In this scenario, the focus is on identifying or detecting graphs that belong to out-of-distribution categories. The goal is to design models capable of flagging or distinguishing graphs that significantly deviate from the training data distribution, potentially indicating new or unfamiliar graph patterns.

\textbf{Open-world Graph Learning.} Open-world graph learning deals with the challenge of learning in situations where the set of possible graph classes is not known in advance. This means that the model must not only adapt to new graph classes but also make decisions about classifying data into known classes or identifying data as belonging to a novel class.

\subsection{Taxonomy of Graph Continual Learning}
Continual graph learning addresses the challenge of acquiring knowledge from a streaming  of graph data that arrives over time and continuously evolves. The methods of continual graph learning can be further classified into four categories: architectural, regularization, rehearsal, and hybrid.

\textbf{Architectural Approach}. These approaches focus on modifying the specific architecture of networks, activation functions, or layers of algorithms to address a new task and prevent the forgetting of previous tasks.

\textbf{Regularization Approach}. These approaches consolidate the learned knowledge by adding a regularization item to the loss function, constraining the neural weights from updating in a direction that compromises performance on prior tasks.

\textbf{Rehearsal Approach}. These approaches maintain a memory buffer preserving the information of prior tasks and replay it when learning new tasks to mitigate catastrophic forgetting.

\textbf{Hybrid Approach}. These approaches combine more than one continual learning approach to take advantage of multiple approaches and improve the performance of models.

% \subsection{Frameworks}

\section{Graph Domain Adaptation Learning}
\begin{figure*}[!ht]
	\includegraphics[width=0.85\linewidth]{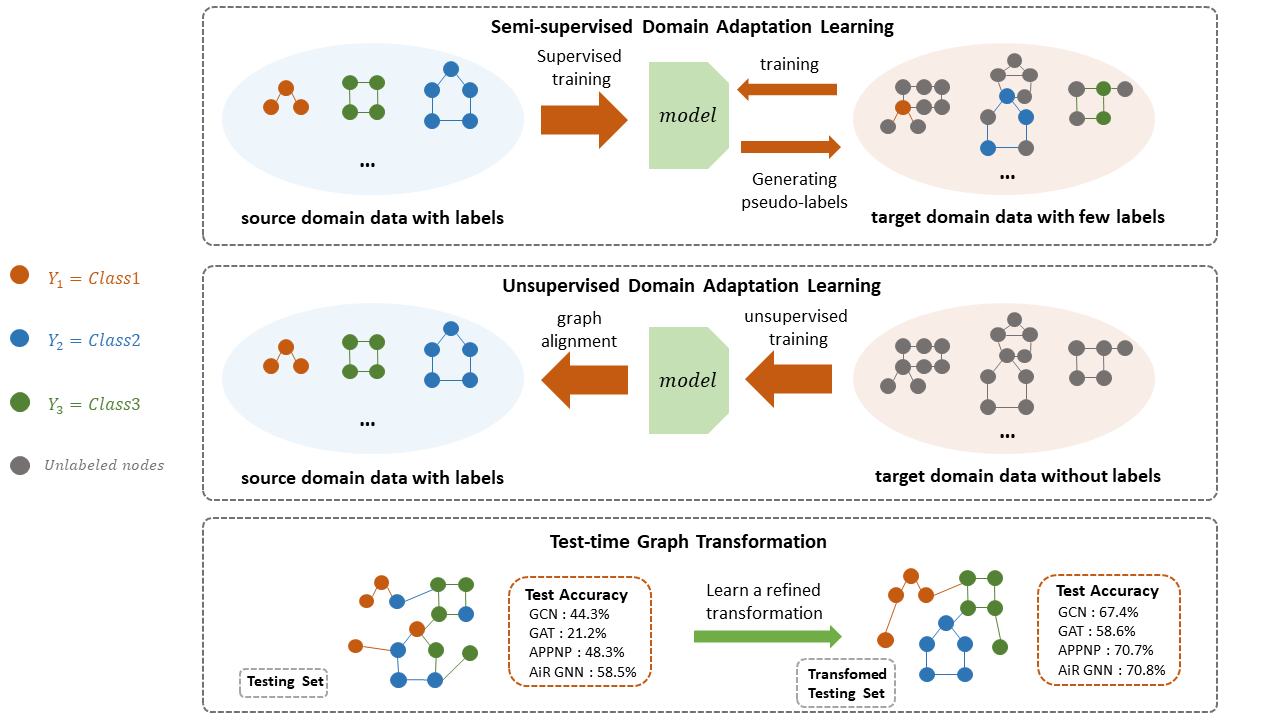}
		\centering
	\caption{The overview of graph domain adaptation learning.}
	\label{fig:Graph_domain_adaption}
\end{figure*}

Domain adaptation is the task of transferring knowledge learned from a source domain to a target domain where the distribution of the data is different. In the context of graph learning, domain adaptation refers to the task of adapting graph-based learning models to new domains where the graph structure, node features, or edge features may be different.
The requirement for domain adaptation in graph learning arises in many real-world applications where the graph data may change over time, or where models trained on one dataset need to be applied to a new dataset with different characteristics. For example, in social network analysis, models trained on one social network may not generalize well to a different social network due to differences in the graph structure and user behavior. In drug discovery, models trained on one set of molecules may not generalize well to a different set of molecules due to differences in the chemical properties of the molecules.

The methods for graph domain adaptation can be roughly divided into three groups: semi-supervised domain adaptation \cite{alam2018domain, pilanci2020domain, dai2019network, wu2019domain, luo2020progressive}, unsupervised domain adaptation \cite{das2018unsupervised, wu2020unsupervised, zhang2018structural, corrabs190510095, ma2019gcan, yang2020heterogeneous}, and test-time graph transformation (adaptation) \cite{mancini2019adagraph, jin2022empowering, chen2022graphtta, wang2022test}. The categories of graph domain adaptation are shown in Fig. \ref{fig:Graph_domain_adaption}.
\begin{itemize}
    \item 
    \textbf{Semi-supervised Domain Adaptation}. These approaches involve adapting from a source domain to a target domain using a small amount of labeled data and unlabeled data from the target domain.

    \item 
    \textbf{Unsupervised Domain Adaptation}. These approaches aim to enhance the model's cross-domain generalization performance by reducing the distribution discrepancy between the source and target domains through feature alignment, without relying on any labeled target data.

    \item 
    \textbf{Test-time Graph Transformation (Adaptation)}. These approaches refer to adjusting the graph data of the target domain during testing to accommodate the characteristics of the target domain. This approach aims to enhance the performance of a model by fine-tuning its behavior based on the specifics of the target domain during the testing phase.
    
\end{itemize}

\subsection{Semi-supervised Domain Adaptation}
Semi-supervised domain adaptation denotes a circumstance wherein labeled data is accessible within the source domain, whereas the target domain possesses a limited fraction of labeled data. 
The primary aim of this paradigm is to bolster the model's efficacy within the target domain through adept utilization of these labeled data instances from the source domain, along with any potentially available labeled data within the target domain. At its essence, this approach amalgamates the principles of supervised and unsupervised learning. This synergy is directed towards orchestrating a more potent adaptation to the data distribution inherent within the target domain. The ultimate aspiration is to engender robust predictive outcomes within this new domain context.

\textbf{DA-AGE} \cite{alam2018domain} is a model based on adversarial domain adaptation, designed to address distribution shift issues. It combines graph-based semi-supervised learning techniques within the framework of deep learning to leverage unlabeled data. This model is capable of training deep neural network (DNN) models using both labeled and unlabeled data, thereby tackling challenges arising from data distribution differences between the source and target domains.

\textbf{DASGA} \cite{pilanci2020domain} is a method used for graph domain adaptation. This algorithm is based on scenarios where the source graph has a larger number of annotated nodes. It achieves this by learning the label function spectrum in the source graph and transferring it to the target graph, even when the target graph has fewer labeled nodes. This process relies on transformations between Fourier bases to address the lack of one-to-one correspondence between Fourier basis vectors in two independently constructed graph domains, thus enabling flexible alignment between these two graphs.

\begin{figure}
    \centering
    \includegraphics[width=0.8\linewidth]{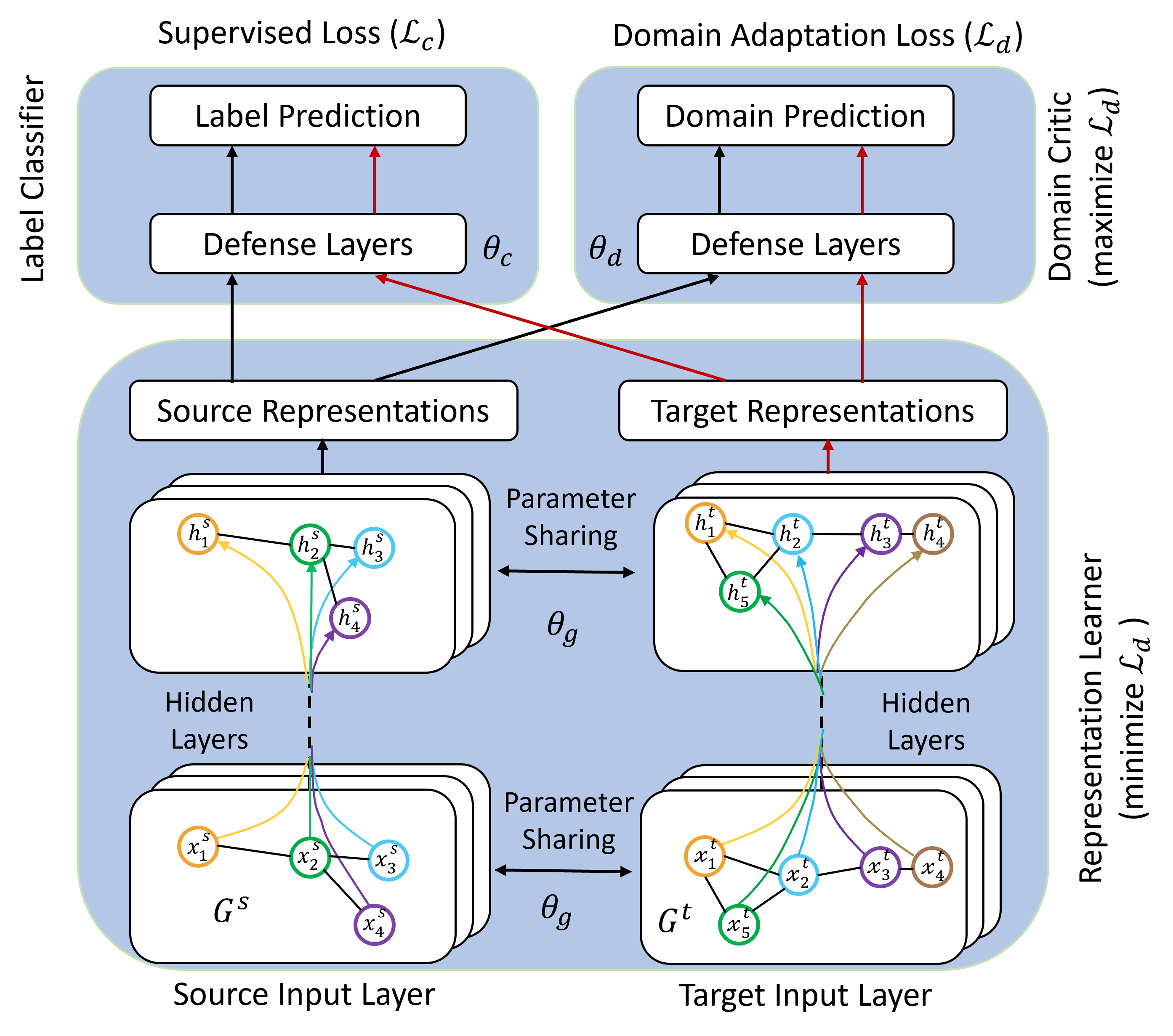}
    \caption{AdaGCN incorporates adversarial domain adaptation and graph convolution methods.}
    \label{fig:AdaGCN-img}
\end{figure}
\textbf{AdaGCN} \cite{dai2019network} is a graph transfer learning framework that combines adversarial domain adaptation and graph convolution techniques. It consists of two main components: a semi-supervised learning part and an adversarial domain adaptation part. The former aims to utilize the label information available in both the source and target networks to learn representations capable of distinguishing nodes of different categories. The role of the latter is to alleviate distribution differences between the source and target domains, thereby promoting the effectiveness of knowledge transfer.

\textbf{DAGNN} \cite{wu2019domain} is an end-to-end framework for cross-domain text classification. It models documents as graphs, employs domain-adversarial training for feature optimization, and captures non-consecutive semantics. By leveraging knowledge from different domains with hierarchical graph neural networks, DAGNN improves task performance by utilizing relationships and patterns across domains.

\begin{figure}
    \centering
    \includegraphics[width=1\linewidth]{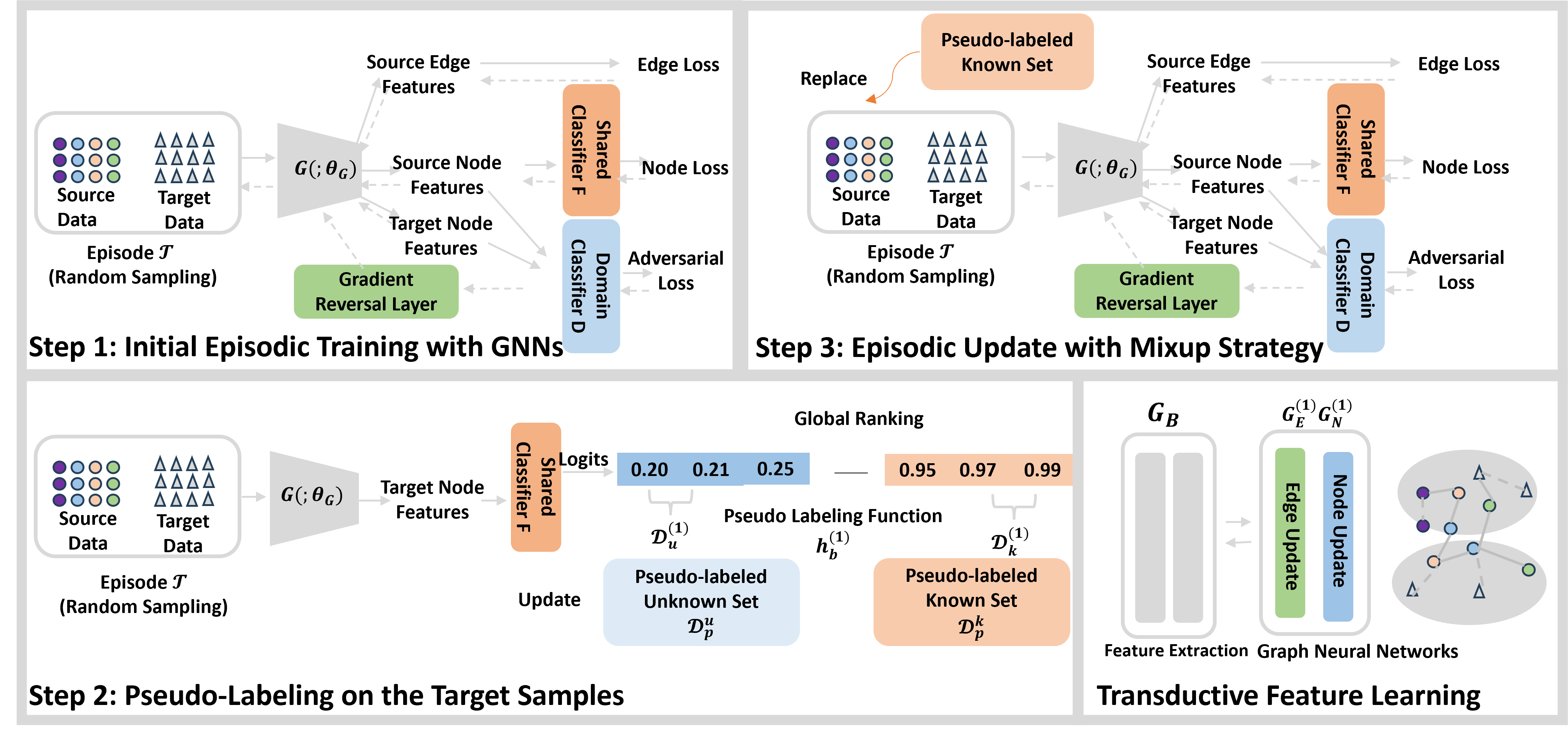}
    \caption{PGL is an end-to-end progressive graph learning framework that integrates graph neural networks and episodic training.}
    \label{fig:PGL-img}
\end{figure}
\textbf{PGL} \cite{luo2020progressive} is an end-to-end progressive graph learning framework known for integrating graph neural networks with occasional training. This design helps mitigate potential mode collapse and employs adversarial learning to reduce the disparity between the source data distribution and the target data distribution. The framework is aimed at addressing more realistic open-set domain adaptation problems, where the target data may include additional categories not present in the source data.

% add discussion
\textbf{Discussion.}  
The success of semi-supervised domain adaption relies not only on the generalization capabilities of transfer models but also depends on the in-depth analysis and application of limited labeled data in the target domain. This integrated approach is key for future research and practice.

\subsection{Unsupervised Domain Adaptation}
Unsupervised Domain Adaptation is centered on the task of transferring a model's knowledge when there exists an absence of labeled data within the target domain. The principal goal of this paradigm is to augment the model's efficacy within the target domain despite the unavailability of labeled target domain data. This is realized through the strategic alignment of the feature distributions inherent to both the source and target domains. The crux of this approach rests upon the harmonization of features, with the intention of mitigating the dissimilarities existing between the source and target domain features. This alignment is instrumental in empowering the model to adeptly adapt to the distinctive characteristics present within the target domain.

\textbf{UDA-HGM} \cite{das2018unsupervised} is an unsupervised domain adaptation method that focuses on unlabeled data in the target domain, emphasizing the correspondence between samples from the source and target domains. The research employs hypergraph representations of the source and target domains, utilizing first-order, second-order, and third-order similarities to achieve hypergraph matching with class regularization. To enhance computational efficiency, the research begins by selecting subsets of samples to construct graphs and then employs a customized optimization method for graph matching using conditional gradients and alternating direction method of multipliers for the matching process.

\textbf{UDA-GCN} \cite{wu2020unsupervised} is an unsupervised domain adaptation graph convolutional network that focuses on domain adaptation learning for graph data. In its primary stage, it constructs a dual graph convolutional network component that operates on the basis of local and global consistencies to aggregate features. Further, an attention mechanism is applied to establish consistent representations for nodes across different graph domains. Additionally, a domain adaptation learning module is introduced to jointly optimize three different loss functions, facilitating cross-domain knowledge transfer.

\textbf{CDNE} \cite{shen2020network} a cross-network deep network embedding model which incorporates domain adaptation into deep network embedding to learn label-discriminative and network-invariant node vector representations. For deep network embedding, CDNE leverages two stacked autoencoders to reconstruct the network proximity matrix of the source network and the target network respectively. Additionally, CDNE designs pairwise constraint to map more strongly connected nodes within each network closer in the embedding space. For domain adaptation, the marginal and class-conditional MMD constraints have been incorporated into CDNE to reduce the distribution shifts between the source and target networks.

\textbf{ACDNE} \cite{shen2020adversarial} is an adversarial cross-network deep network embedding model, which integrates adversarial domain adaptation with deep network embedding. On one hand, a deep network embedding module is designed by ACDNE, by utilizing two feature extractors to learn the embeddings from each node’s own attributes and its neighbors’ attributes respectively. On the other hand, an adversarial domain adaptation module is employed by ACDNE to make a domain discriminator compete against the deep network embedding module, so as to learn network-invariant node embeddings.

\textbf{DM-GNN} \cite{shen2023domain} is a domain-adaptive message passing graph neural network which integrates graph neural network with conditional adversarial domain adaptation. DM-GNN separates ego-embedding learning from neighbor-embedding learning via dual feature extractors. DM-GNN devises a label-aware propagation scheme to promote intra-class propagation while avoiding inter-class propagation, making source embeddings more label-discriminative. To match the class-conditional distributions between the source and the target networks, DM-GNN adopts conditional adversarial domain adaptation.

\textbf{DGASN} \cite{shen2023CNEC} is a domain-adaptive graph attention-supervised network designed to address the cross-network homophilous and heterophilous edge classification problem. DGASN proposes to apply direct supervision on graph attention learning guided by the edge labels observed from the source network. Specifically, DGASN assigns lower (higher) attention weights to heterophilous (homophilous) edges during neighborhood aggregation, so as to guarantee more label-discriminative embeddings to separate nodes from different classes.

\textbf{LGA} \cite{zhang2018structural} is a fresh model for enhancing distributional and structural similarities during the adaptation process. It involves embedding data from two domains into a latent space, aligning their distributions by minimizing the maximum mean discrepancy metric. Additionally, the transformed manifold is represented using graphs to maximize graphical structural similarities within the embedding space. This is accomplished by minimizing the spectrum distance between graph Laplacians of the embeddings. The spectrum retains intrinsic manifold information and remains unchanged under data reordering and eigenspace unitary transformations. 

\begin{figure}
    \centering
    \includegraphics[width=1\linewidth]{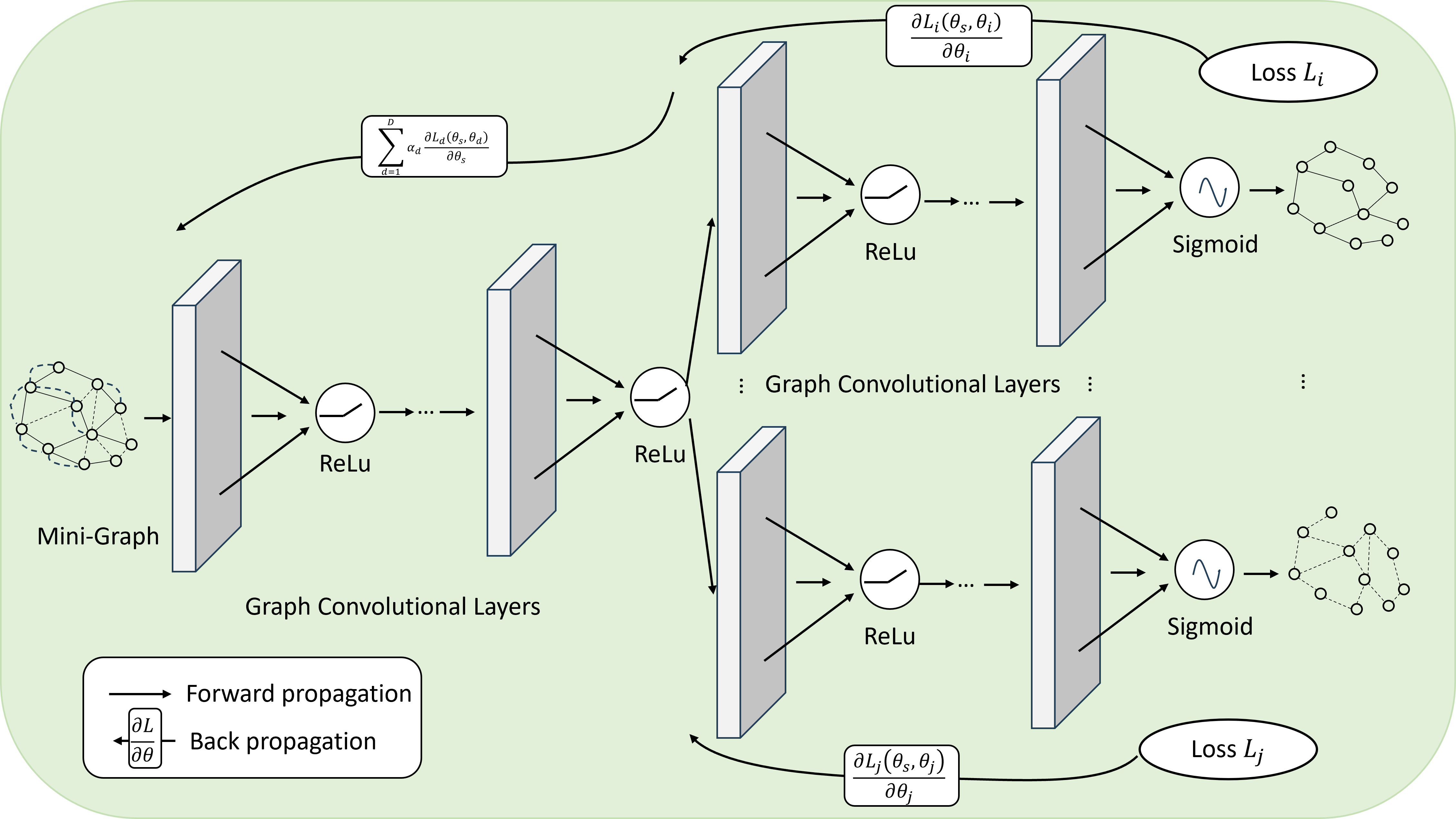}
    \caption{DMGE is a model that utilizes knowledge from different domains.}
    \label{fig:DMGE-o}
\end{figure}
\textbf{DMGE} (Deep Multi-Graph Embedding) \cite{corrabs190510095} is a model enhancing task performance by leveraging knowledge from different domains. It constructs a multi-graph based on the behaviors of users from different domains and utilizes a multi-graph neural network to learn cross-domain representation in an unsupervised manner. Particularly, a multiple gradient descent optimizer is used for efficiently training the model.

% \begin{figure}
%     \centering
%     \includegraphics[width=1\linewidth]{figure/DA-Gcan.png}
%     \caption{Gcan is an end-to-end graphical convolutional adversarial network for joint modeling of multiple data. The figure is from reference \cite{ma2019gcan}.}
%     \label{fig:Gcan-img}
% \end{figure}
\textbf{GCAN} \cite{ma2019gcan} is an end-to-end graph convolutional adversarial network that achieves unsupervised domain adaptation by jointly modeling data structures, domain labels, and class labels within a unified deep framework. Additionally, the model incorporates three effective alignment mechanisms: structure-aware alignment, domain alignment, and class centroid alignment. These alignment mechanisms aid in learning domain-specific features and semantic representations, thereby reducing disparities between different domains and enabling effective domain adaptation.

\begin{figure}
    \centering
    \includegraphics[width=1\linewidth]{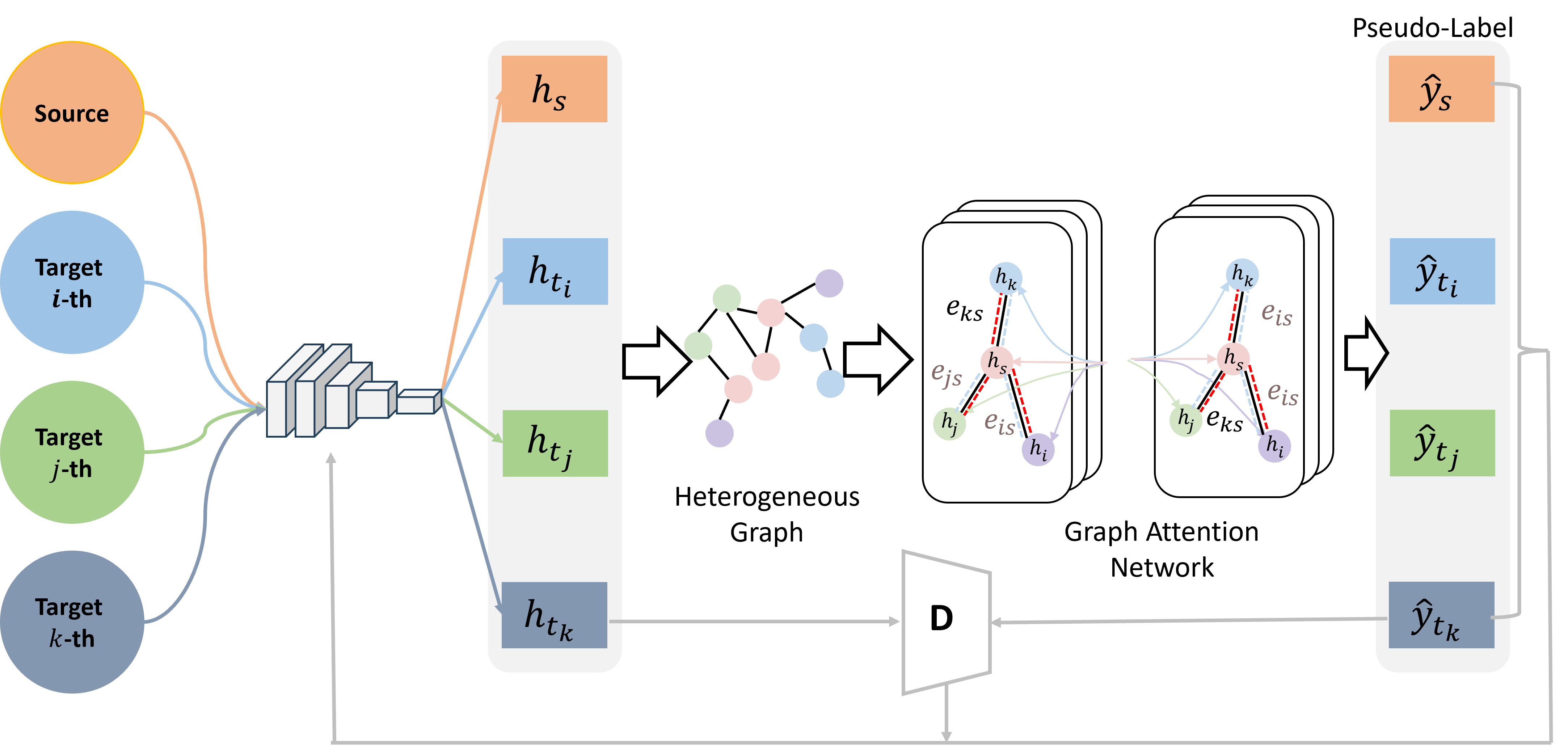}
    \caption{UMTDA enables the learning of domain-invariant representations by sharing similar feature representations through heterogeneous graph attention networks.}
    \label{fig:UMTDA-img}
\end{figure}
\textbf{UMTDA} \cite{yang2020heterogeneous} is a method designed for the scenario of multiple unlabeled target domains and one labeled source domain, aiming to perform deep semantic information propagation. Through a heterogeneous graph attention network, it strives to learn a shared unified subspace that allows all domains to share similar feature representations. In this process, attention mechanisms are used to optimize relationships between multiple domain samples, facilitating enhanced semantic propagation. Subsequently, by leveraging the graph attention network, it predicts pseudo-labels for the target domains. By aligning the centroids of the labeled source domain and the pseudo-label target domain, it achieves the learning of domain-invariant representations.

\textbf{Discussion.} 
A key challenge in unsupervised domain adaption is ensuring that models can generalize well from the source to the target domain without losing the critical information encoded in the graph structures. Future research should focus on enhancing these models to better handle the nuances of graph data, exploring more adaptive and flexible techniques that can account for varying degrees of distribution shifts.

\subsection{Test-time Graph Transformation (Adaptation)}
Test-time graph transformation (adaptation) aims to adapt models based on test samples in the presence of distributional shifts. Test time adaptation can be further subdivided into test time training and fully test time adaptation according to whether it can access the source data and alter the training of the source model. 
Here, we introduce several typical existing methods.

\textbf{AdaGraph} \cite{mancini2019adagraph} presents the first deep architecture for Predictive Domain Adaptation. AdaGraph leverages metadata information to build a graph where each node represents a domain, while the strength of an edge models the similarity among two domains according to their metadata. Then in order to exploit the graph for the purpose of DA and AdaGraph has a  novel domain-alignment layers. This framework yields the new state of the art on standard PDA benchmarks. 

\textbf{GTRANS} \cite{jin2022empowering} which optimizes a contrastive surrogate loss to transform graph structure and node features, and provide theoretical analysis with deeper discussion to understand this framework. Experimental results on distribution shift, abnormal features and adversarial attack have demonstrated the effectiveness of GTRANS. 

\begin{figure}[ht]
	\includegraphics[width=1\linewidth]{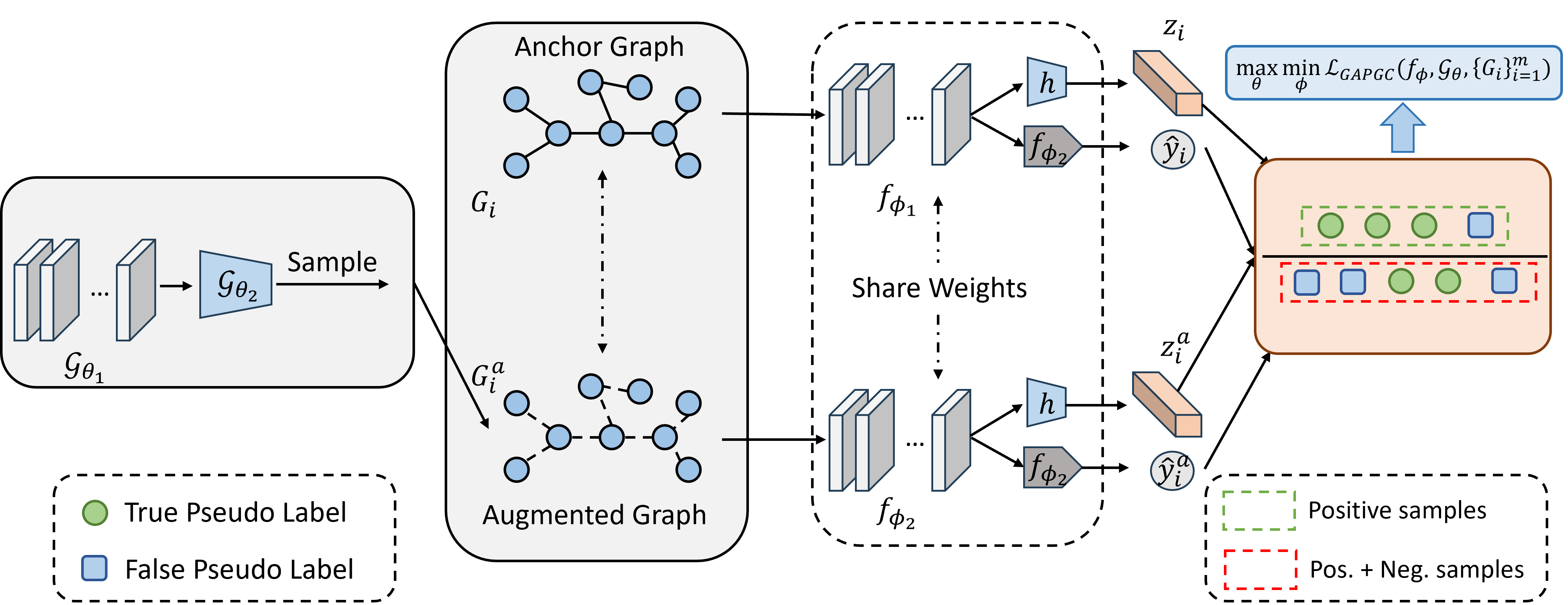}
		\centering
	\caption{GAPGC is a test-time training method designed for GNNs with contrast loss variants as self-supervised targets during testing.}
	\label{fig: GAPGC}
\end{figure}

\textbf{GAPGC} (Graph Adversarial Pseudo Group Contrast) \cite{chen2022graphtta} is a test-time training method designed for GNNs with a contrastive loss variant as the self-supervised objective during testing. The framework of GAPGC is shown in Fig. \ref{fig: GAPGC}. Recently, the effectiveness of test-time training has been validated to improve the performance on OOD test data, where some self-supervised auxiliary tasks are proposed. The authors argue that the simple augmentations in self-supervised training (e.g., randomly dropping nodes or edges) could harm the label-related critical information in graph representations. Therefore, GAPGC generates relatively reliable pseudo-labels, avoiding the severe shifts caused by the incorrect positive samples. The proposed adversarial learnable augmenter and group pseudo-positive samples can promote the relevance between the self-supervised task and the main task, to enhance the performance of the main task. The theoretical evidence is also derived to show that GAPGC can capture minimal sufficient information for the main task from an information theory perspective, which benefits the predictions on the OOD testing data. 

\textbf{GT3} (Graph Test-Time Training with Constraint) \cite{wang2022test} is another test-time training method on graphs, which proposes a hierarchical self-supervised learning framework. Specifically, it first introduces the global contrastive learning strategy to encourage node representations to capture the global information of the whole graph. Global contrastive learning is based on maximizing the mutual information between the local node representation and the global graph representation. Then, it presents the local contrastive learning for distinguishing different nodes from different augmented views of a graph, so that the node representation can capture more local information. Besides, an additional constraint is proposed to encourage that the representations of testing samples are close to the representations of the training samples. The model’s generalization capacity for the graph classification task can be enhanced based on this test time training strategy with self-supervised learning. 

\textbf{Discussion.}
Test-time Graph Transformation (Adaptation) is a sophisticated approach aimed at enhancing the adaptability of graph-based machine learning models to new, potentially diverse target domains by adjusting the graph data during the testing phase.

% \begin{figure}[ht]
% 	\includegraphics[width=1\linewidth]{figure/GT3.png}
% 		\centering
% 	\caption{The hierarchical self-supervised learning framework of GT3.The figure is from reference \cite{wang2022test}}
% 	\label{fig:GT3}
% \end{figure}

\begin{table}[ht]
\caption{A summary of graph domain adaptation (GDA) learning methods.}
\label{tab: Graph Domain Adaptation methods}
\centering
\resizebox{0.48\textwidth}{!}{
\begin{tabular}{p{5cm}p{3cm}<{\centering}} 
\toprule
Category                                                                                          & Method    \\ 
\midrule
\multirow{7}{*}{Semi-supervised GDA Learning}           & DA-AGE \cite{alam2018domain}   \\ 
\cmidrule(r){2-2}
& DASGA \cite{pilanci2020domain}  \\ 
\cmidrule(r){2-2}
& AdaGCN \cite{dai2019network}   \\ 
\cmidrule(r){2-2}
& DAGNN \cite{wu2019domain}    \\ 
\cmidrule(r){2-2}
& PGL \cite{luo2020progressive}      \\ 
\cmidrule(r){1-2}
\multirow{14}{*}{Unsupervised GDA Learning}              & UDA-HGM \cite{das2018unsupervised}  \\ 
\cmidrule(r){2-2}
& UDA-GCN \cite{wu2020unsupervised}  \\ 
\cmidrule(r){2-2}
& CDNE \cite{shen2020network}    \\ 
\cmidrule(r){2-2}
& ACDNE \cite{shen2020adversarial}     \\ 
\cmidrule(r){2-2}
& DM-GNN \cite{shen2023domain}     \\ 
\cmidrule(r){2-2}
& DGASN \cite{shen2023CNEC}    \\ 
\cmidrule(r){2-2}
& LGA \cite{zhang2018structural}     \\ 
\cmidrule(r){2-2}
& DMGE \cite{corrabs190510095}    \\ 
\cmidrule(r){2-2}
& GCAN \cite{ma2019gcan}    \\ 
\cmidrule(r){2-2}
& UMTDA \cite{yang2020heterogeneous}    \\ 
\cmidrule(r){1-2}
\multirow{5}{*}{\begin{tabular}[c]{@{}c@{}}Test-time Graph Transformation\\~(Adaptation)\end{tabular}} & AdaGraph \cite{mancini2019adagraph} \\ 
\cmidrule(r){2-2}
& GTRANS \cite{jin2022empowering}   \\ 
\cmidrule(r){2-2}
& GAPGC \cite{chen2022graphtta}    \\ 
\cmidrule(r){2-2}
& GT3 \cite{wang2022test}      \\
\bottomrule
\end{tabular}
}
\end{table}

\section{Graph Out-of-distribution Learning}
Out-of-distribution (OOD) graph learning refers to the problem of learning from graphs that have different characteristics than those seen during training. Specifically, in graph-based machine learning, a common assumption is that the training and test data come from the same distribution. However, in many real-world scenarios, this assumption does not hold, and the test graphs may have different statistical properties than the training graphs. In such cases, a model trained on the training graphs may not generalize well to the test graphs.
OOD graph learning addresses this problem by developing models that can handle the distribution shift between the training and test graphs. One way to achieve this is by learning representations that are invariant to the differences between the training and test distributions. Another approach is to explicitly model the distribution shift between the training and test graphs and adapt the model accordingly.

OOD graph learning has several applications, including social network analysis, recommender systems, and drug discovery. In these applications, the graphs may have different characteristics depending on the population or context, and it is important to develop models that can handle these differences to achieve good performance in the target application.

\begin{figure*}[h]
\includegraphics[width=0.85\linewidth]{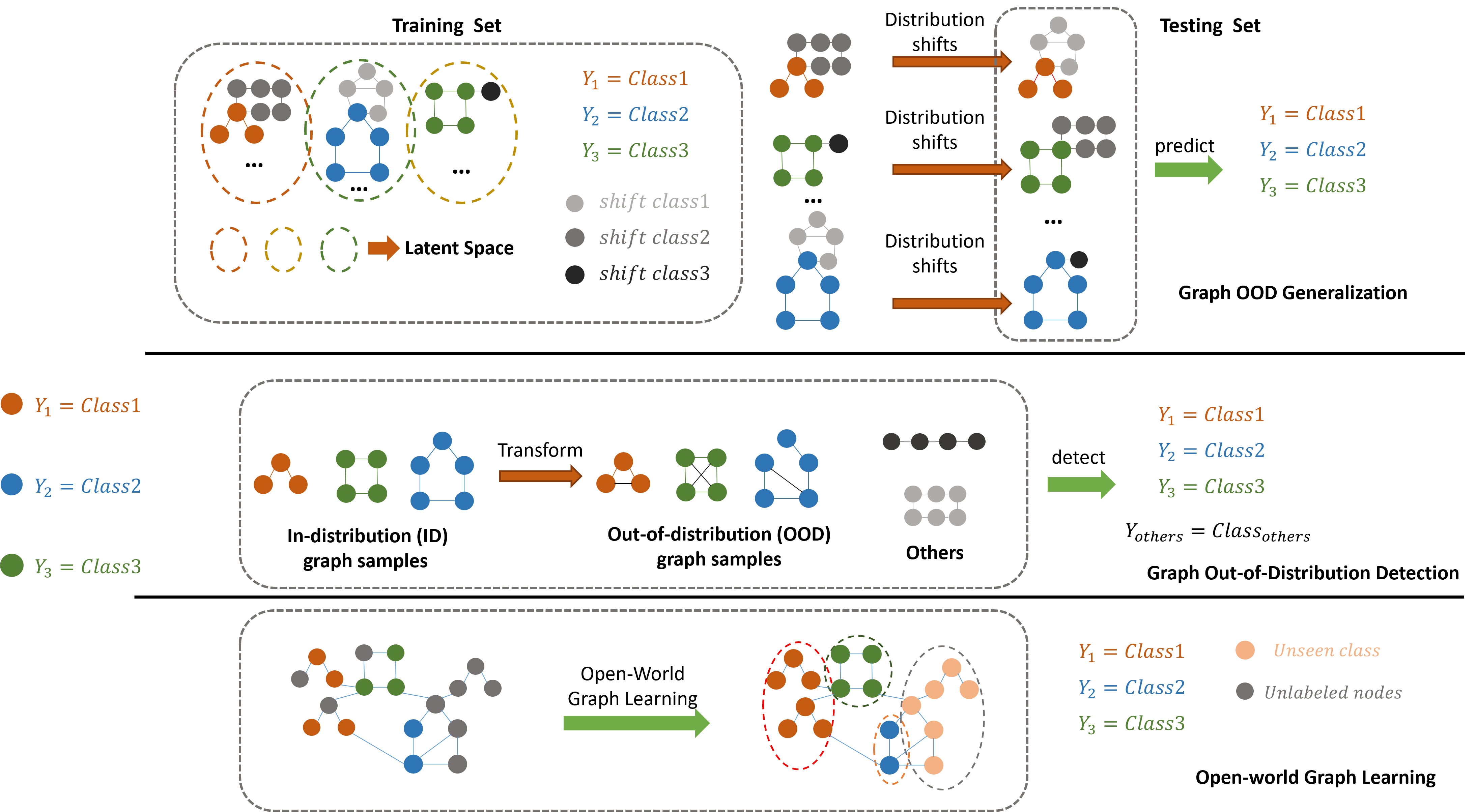}
		\centering
	\caption{The overview of graph out-of-distribution learning.}
	\label{fig:Graph_OOD}
\end{figure*}

Fig. \ref{fig:Graph_OOD} depicts an overview of the mentioned domains, in which the differences are shown visually.

\subsection{Graph Out-of-distribution Generalization}
% \cite{li2022out, chen2022learning, li2022learning, baranwal2021graph}
% \begin{definition}[\textbf{Graph Out-of-Distribution Generalization}]
% Given the training set $\mathcal{D} = \{(X_i,Y_i)\}_{i=1}^N$ of $N$ instances (i.e., nodes, links, or graphs) that are drawn from the training distribution $P_{tr}(X,Y)$, the goal is to learn an optimal graph predictor $f^*_\theta$ that can achieve the best generalization on the data drawn from test distribution $P_{te}(X,Y)$,where $P_{te}(X,Y)\neq P_{tr}(X,Y)$:
% \begin{equation}
%     f^*_\theta = \argmin _{f_\theta} \mathbb{E}_{(X,Y)\sim P_{te}}[\mathcal{L} (f_\theta(X), Y)].
% \end{equation}
% \end{definition}

To tackle the challenges brought by unknown distribution shifts and solve the graph OOD generalization problem, considerable efforts have been made in literature, which can be divided into three categories. We select some typical examples for each category to introduce.

\subsubsection{Data Augmentation}
 This category of methods aims to manipulate the input graph data, i.e., graph augmentation. By systematically generating more training samples to increase the quantity and diversity of the training set, graph augmentation techniques are effective in improving the OOD generalization performance. %Here we mainly review the representative structure-wise graph data augmentation approaches.
 
\textbf{GAug} (Graph Augmentation) \cite{DBLP:conf/aaai/0003LNW0S21} is a method for feature-wise graph augmentation. It utilizes a differentiable edge predictor to generate augmented graphs by encoding class-homophilic structure, which promotes intra-class edges and reduces inter-class edges to enhance prediction accuracy and generalization ability. GAug modifies the input graph using an edge prediction module and learns to generate new possible edges.

\textbf{NCLA} (Neighbor Contrastive Learning on Learnable Graph Augmentation) \cite{shen2023neighbor} is an end-to-end automatic graph contrastive learning method, applying neighbor contrastive learning on learnable graph augmentation. NCLA automatically generates multiple graph augmented views with adaptive topology by the multi-head graph attention mechanism, where each attention head corresponds to each augmented view. The attention-based learnable graph augmentation avoids improper modification of the original topology and can be compatible with various graph datasets without prior domain knowledge.

\textbf{GRAND} (Graph Random Neural Network) \cite{feng2020graph} is a framework for enhancing generalization in GNNs by combining feature-wise graph augmentation methods. First, it randomly drops node features and propagates perturbed features. Then in accordance with the graph homophily assumption\cite{mcpherson2001birds}, stochastic augmentation is employed to generate diverse augmented representations for individual nodes.
  
  \begin{figure}
       \centering
       \includegraphics[width=1\linewidth]{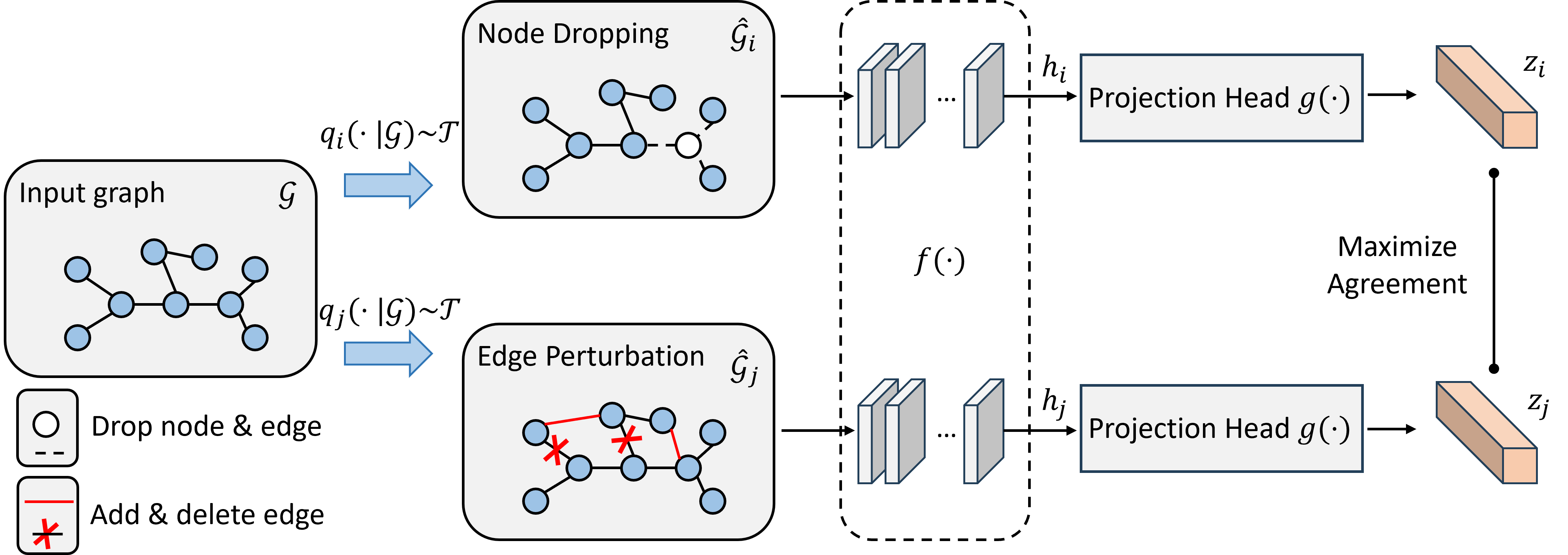}
       \caption{A framework of GraphCL. GraphCL is a mixed-type graph data augmentation method, which provides four data augmentations, including node dropping, edge perturbation, attribute masking, and subgraph sampling.}
       \label{fig:GraphCL-o}
   \end{figure}
\textbf{GraphCL} (Graph Contrastive Learning) \cite{you2020graph} GraphCL is a mixed-type graph data augmentation method, which provides four data augmentations, including node dropping, edge perturbation, attribute masking, and subgraph sampling. As depicted in Fig. \ref{fig:GraphCL-o}, two categories of augmentations are randomly sampled from a pool of augmentations and implemented on the input graph. A GNN encoder and a projection head are trained with the objective of maximizing the consensus between their respective representations\cite{ding2021closer}.
%After obtaining the augmented samples of the input, it makes the graph encoder maximizes representation consistency under augmentations and has shown good OOD generalization ability in graph classification\cite{ding2021closer}. 
   
\subsubsection{Model Develoment}
This category of methods aims to propose new graph models for learning OOD generalized graph representations. 

\textbf{DisenGCN} \cite{ma2019disentangled} is a method that utilizes the DisenConv, a disentangled multichannel convolutional layer, to learn disentangled representations of nodes. Within DisenConv, a dynamic process determines the neighboring nodes that are influenced by specific latent factors and directs them to the corresponding channels for information extraction and convolution. The disentangled nature of these latent factors contributes to improved generalization performance when dealing with out-of-distribution data.

\textbf{OOD-GNN} \cite{li2022ood} is a GNN designed backed by confounder balancing theory\cite{kuang2018stable} in causality. It is designed for poor generalization on unseen testing graphs with different distributions. It utilizes a nonlinear graph representation decorrelation method using random Fourier features to encourage independence between relevant and irrelevant graph representations, eliminating spurious correlations. OOD-GNN optimizes sample graph weights and the graph encoder to enhance discriminative graph representations, resulting in improved predictions.

\subsubsection{Learning Strategy}
This category of methods focuses on exploiting the training schemes with tailored optimization objectives and constraints to enhance the OOD generalization capability.%, including graph invariant learning, graph adversarial training, and graph self-supervised learning. %Here are several examples showcasing the aforementioned methods.

\textbf{StableGL} \cite{zhang2021stable} is a framework for Graph Invariant Learning in GNNs that addresses unstable inference in different test environments caused by distribution shifts. It consists of two essential components: Locally stable learning captures stable node properties by reweighting the neighborhood aggregation process, while globally stable learning reduces training losses in different environments, ensuring the generalization of GNNs.

% {\color{red}\textbf{DAGNN} (Domain Adversarial GNN) \cite{wu2019domain} is a method inspired by DANNmotivated by DANN \cite{ganin2016domain} that aims to learn domain-invariant graph representations. It employs domain-adversarial learning between a domain classifier and an encoder. The method focuses on minimizing the classification loss on source domain data and improving differentiation between source and target domains. By utilizing a graph adversarial training strategy, DAGNN maximizes the use of domain information for OOD generalized predictions in text classification tasks.}
 
\textbf{Pretraining-GNN} \cite{hu2019strategies} refers to a self-supervised technique employed in the pre-training of GNNs. The crux of this approach lies in the pre-training of a highly expressive GNN on both individual nodes and complete graphs, enabling the simultaneous acquisition of valuable local and global representations. Consequently, this method effectively sidesteps negative transfer phenomena and substantially enhances the model's generalization capabilities across various downstream tasks.

\textbf{Discussion.}
In the graph OOD generalization, while the Data Augmentation addresses graph OOD generalization by diversifying the training graph data, Model Development focuses on building inherently more adaptable models, and Learning Strategy emphasizes the training schemes with tailored optimization objectives and constraints that enhance the model's ability to generalize to new distributions. Each approach tackles the challenge of Graph OOD Generalization from a different angle, offering a comprehensive capability for improving model performance on unseen graph data.

% \begin{table*}
%     \caption{A summary of graph OOD generalization methods.}
%     \centering
%     \label{tab:graph OOD generalization}
%     \centering
% \begin{tabular}{|l|l|l|}

% \hline
    
% \multirow{3}{*}{Data}              & Structure-wise Graph Data Augmentation & GAug \cite{DBLP:conf/aaai/0003LNW0S21}  \\ \cline{2-3} 
%                                    & Feature-wise Graph Data Augmentation   & GRAND \cite{feng2020graph} \\ \cline{2-3} 
%                                    & Mixed-type Graph Data Augmentation     & GraphCL \cite{you2020graph} \\ \hline
% \multirow{2}{*}{Model}             & Disentanglement-based Graph Models     & DisenGCN  \cite{ma2019disentangled} \\ \cline{2-3} 
%                                    & Causality-based Graph Models           & OOD-GNN \cite{li2022ood} \\ \hline
% \multirow{3}{*}{Learning Strategy} & Graph Invariant Learning               & StableGL \cite{zhang2021stable} \\ \cline{2-3} 
%                                    & Graph Adversarial Training             & DAGNN \cite{wu2019domain} \\ \cline{2-3} 
%                                    & Graph Self-supervised Learning         & Pretraining-GNN \cite{hu2019strategies} \\ \hline
% \end{tabular}
% \end{table*}

\subsection{Graph Out-of-distribution Detection}
Graph Out-of-Distribution detection (Graph OOD detection) aims to identify test-times samples that are semantically different from the training data categories, and therefore should not be predicted into the known classes. Here, we select some typical examples for the introduction.

\textbf{Uncertainty-GNN} \cite{zhao2020uncertainty} is an innovative framework that integrates multiple uncertainty sources to enhance the performance of GNNs in semi-supervised node classification tasks. It effectively improves predictions and accuracy by leveraging deep learning-based and belief/evidence theory-based uncertainties. Uncertainty-GNN improves GNN performance in node classification tasks by effectively addressing misclassification and out-of-distribution detection.

\textbf{GPN} (Graph Posterior Networks) \cite{stadler2021graph} is a model that combines node classification and OOD detection, using variational inference and a latent variable to approximate the true posterior distribution and differentiate between in-distribution and OOD nodes. LMN mitigates cross-distribution mixing, learns from neighboring nodes without explicit OOD labels, and reduces overfitting with bi-level optimization. Overall, LMN provides a comprehensive solution for accurate node classification and reliable OOD detection.

\textbf{LMN} \cite{huang2022end} is a model that combines node classification and OOD detection, using variational inference and a latent variable to approximate the true posterior distribution and differentiate between in-distribution and OOD nodes. LMN mitigates cross-distribution mixing, learns from neighboring nodes without explicit OOD labels, and reduces overfitting with bi-level optimization. Overall, LMN provides a comprehensive solution for accurate node classification and reliable OOD detection.

\textbf{GraphDE} \cite{li2022graphde} is a unified framework for graph OOD detection and debiased learning. It utilizes a latent environment variable and generative models to address outliers in the training data. An OOD detector is also included to distinguish between in-distribution and OOD samples, enhancing the robustness of graph representation learning. Overall, GraphDE is a promising solution for reliable graph representation learning in real-world applications.

% \begin{figure}[h]
% 	\includegraphics[width=1\linewidth]{figure/GraphDE.PNG}
% 		\centering
% 	\caption{The framework of GraphDE\cite{li2022graphde}.\textbf{Left}: The training data is processed in two steps. Initially, the recognition model infers the environment variable from the data. This inferred variable is then used by the generative models to compute the overall training loss. \textbf{Right}: The testing data undergoes a similar procedure. First, it is inputted into the structure estimation model to detect OOD samples. Subsequently, the OOD-detected data is passed to the classification model to estimate class probabilities.}
% 	\label{fig:GrapgDE}
% \end{figure}

\textbf{GOOD-D} (Graph Out-Of Distribution Detection) \cite{liu2023good} is an unsupervised method for detecting OOD graphs. 
% As shown in Fig. \ref{fig:GOOD-D}, 
It utilizes a hierarchical graph contrastive learning framework to capture attributive and structural patterns from in-distribution (ID) training data. By maximizing agreement between feature and structure graph views at multiple levels, GOOD-D effectively detects OOD graphs. It incorporates a self-adaptive mechanism to balance learning objectives and OOD scores.

\textbf{Discussion.}
Graph OOD Detection aims to equip models with the ability to identify and detect graphs that represent new, unknown, or significantly different patterns from those seen during training. This capability is essential in numerous fields where understanding and reacting to evolving or novel graph structures is crucial.

\begin{figure}[t]
\includegraphics[width=1\linewidth]{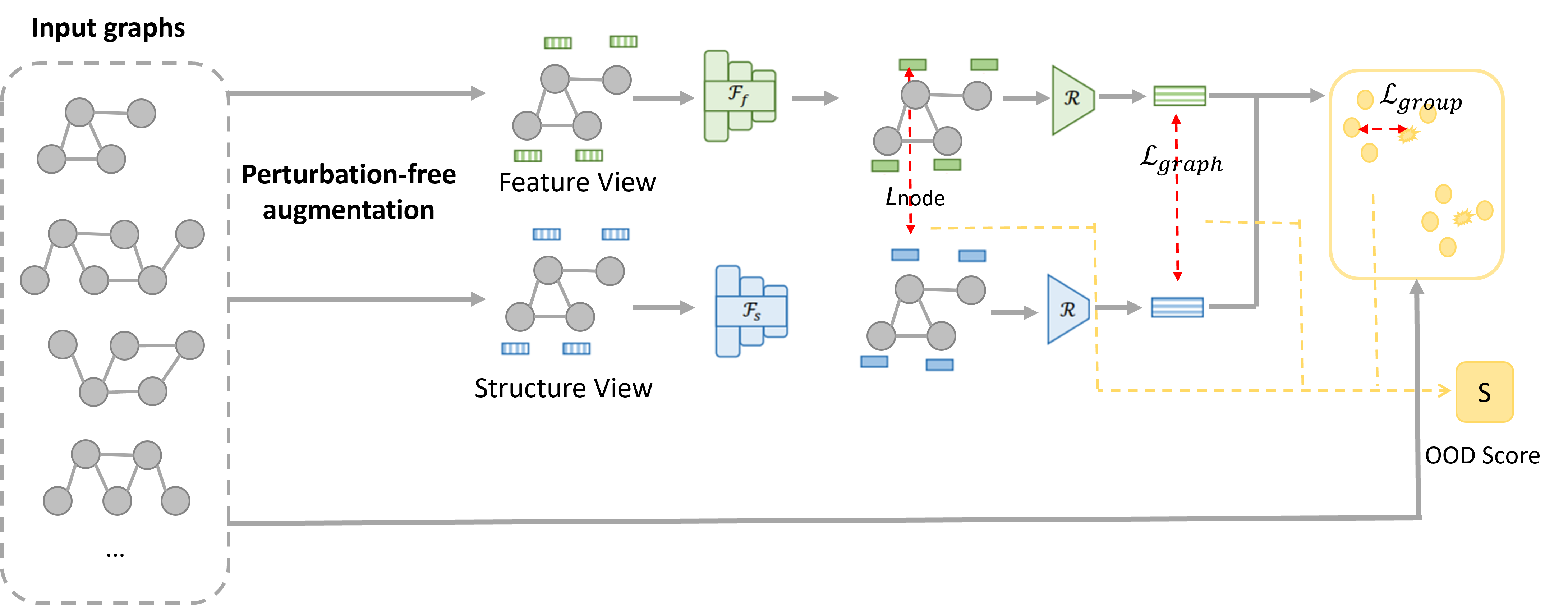}
		\centering
	\caption{The framework of GOOD-D.}
	\label{fig:GOOD-D}
\end{figure}
% \subsubsection{Graph Outer Detection}

% \subsubsection{Novel Class Detection}

\subsection{Open-world Graph Learning}
Open-world Graph Learning is a task where models are trained in a setting without prior knowledge of the number of classes or their training samples. The challenge lies in distinguishing unseen classes without labeled samples from existing classes. This applies to both graph structures and knowledge graphs. The literature can be categorized into graph structure and knowledge graph studies.

\subsubsection{Graph Structure}

In the open-world learning of graph structure, the model needs to better learn the feature representation of graph data, classify nodes belonging to known classes into correct groups through threshold judgment or classifier output, etc., and also classify nodes not belonging to existing classes into invisible classes \cite{zhang2023g2pxy,zhang2022dynamic,wu2020openwgl}.

\begin{figure}[t]
	\includegraphics[width=1\linewidth]{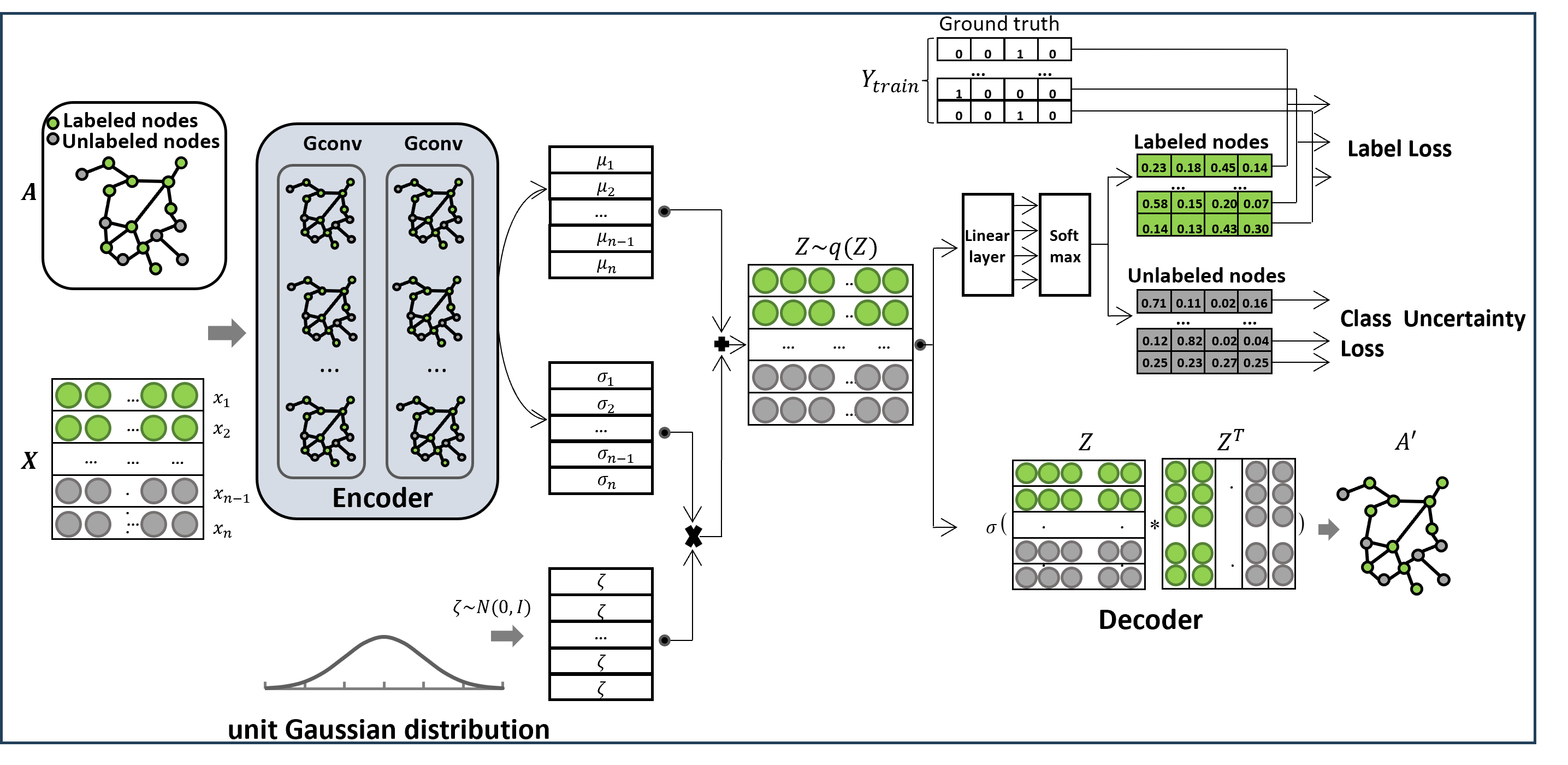}
		\centering
	\caption{The framework of OpenWGL. OpenWGL captures uncertainty using a graph variational autoencoder for node embeddings and differentiates between seen and unseen class nodes through label and class uncertainty loss minimization with an automatic threshold.}
	\label{fig:OpenGL}
\end{figure}

\textbf{OpenWGL} \cite{wu2020openwgl} tackles open-world graph learning with two key components: Node Uncertainty Representation Learning and Open-world Classifier Learning. As shown in Fig. \ref{fig:OpenGL}, it captures uncertainty using a graph variational autoencoder for node embeddings and differentiates between seen and unseen class nodes through label and class uncertainty loss minimization with an automatic threshold. This approach provides a comprehensive solution for accurate classification in open-world scenarios.

\textbf{FATE} (Federated Learning via Adapted Transformer) \cite{wu2021towards} is a federated learning framework that handles open-world feature extrapolation using graph representation and learning. It utilizes a backbone network and a graph neural network to handle new features without retraining. FATE includes two training strategies for better extrapolation and reduced overfitting. The framework is effective and scalable, supported by theoretical analysis and experiments.

\textbf{Co-CGE} \cite{mancini2022learning} is an approach for Compositional Zero-Shot Learning (CZSL) in an open-world setting that aims to recognize unseen compositions of known state-object primitives. It utilizes a graph convolutional neural network to model dependencies between states, objects, and compositions. By incorporating feasibility scores, Co-CGE enhances CZSL performance by considering the feasibility of unseen compositions.

\subsubsection{Knowledge Graph}
Knowledge graph completion (KGC) aims to fill in missing relationships in a Knowledge Graph (KG). Traditional closed-world approaches for KGC are limited in their ability to address a small number of missing relations. However, recent advancements in KGC involve incorporating knowledge from open-world resources, like online encyclopedias and newswire corpus, to enhance completion capabilities.

\textbf{ConMask} \cite{shi2018open} is a model specifically designed for open-world Knowledge Graph Completion (KGC) tasks. It consists of Relationship-Dependent Content Masking, Target Fusion, and Target Entity Resolution components. By masking irrelevant information, extracting entity embeddings, and generating similarity rankings, ConMask effectively connects new entities to knowledge graphs, improving their completeness in an open-world setting.

\begin{figure}[ht]
	\centering
 \includegraphics[width=0.9\linewidth]{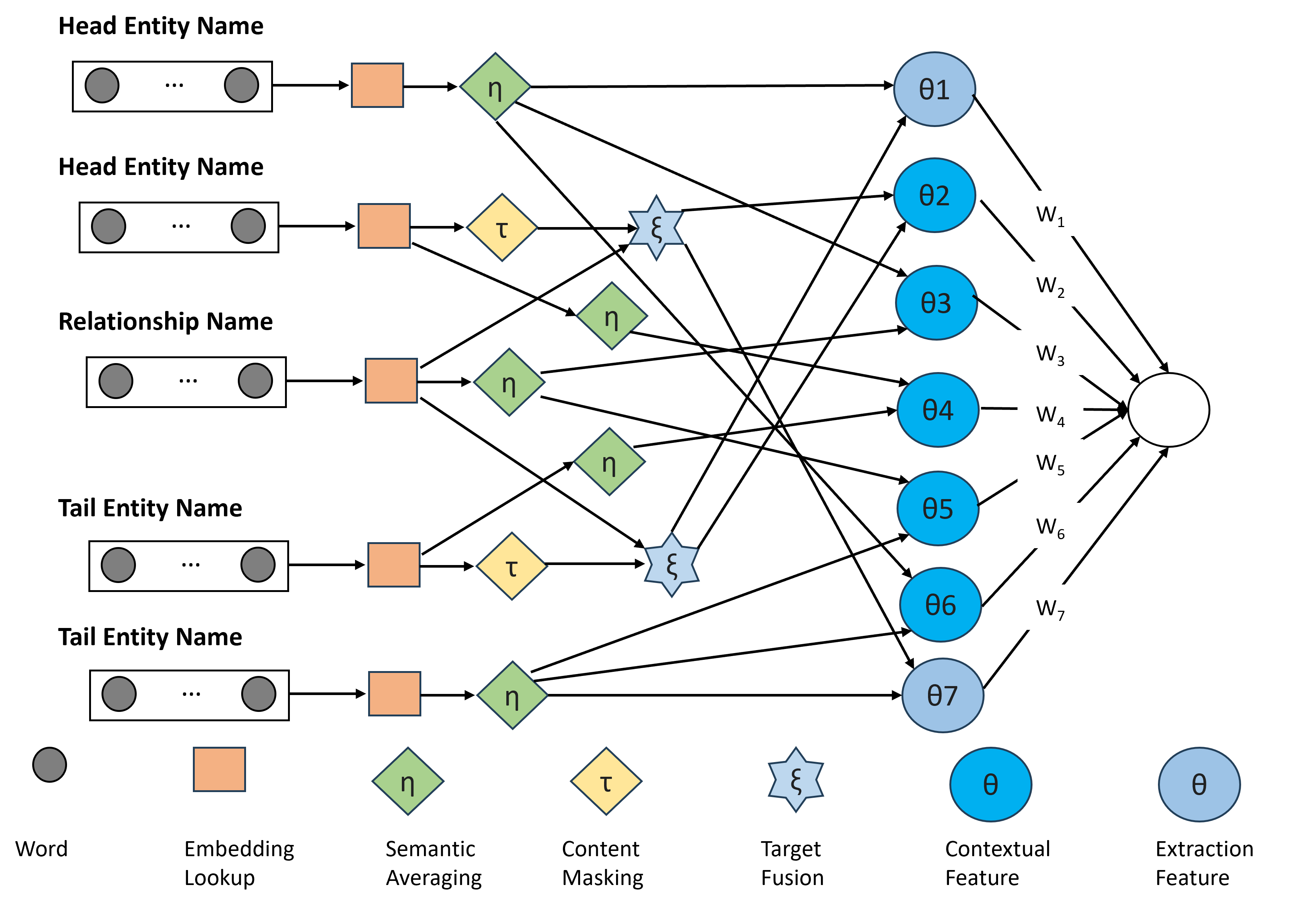}
	\caption{The framework of ConMask. It consists of Relationship-Dependent Content Masking, Target Fusion, and Target Entity Resolution components.}
	\label{fig:ConMask}
\end{figure}

\textbf{Prob-CBR} (Probabilistic Case-based Reasoning) \cite{das2020probabilistic} is an approach for open-world graph completion. It combines probabilistic and case-based reasoning principles to handle uncertainty in knowledge graphs. By retrieving similar cases, Prob-CBR predicts attributes using reasoning paths. Its probabilistic model estimates query effectiveness with simple path statistics. It dynamically grows with new entities and outperforms other approaches.

\begin{figure}[!t]
	\includegraphics[width=1\linewidth]{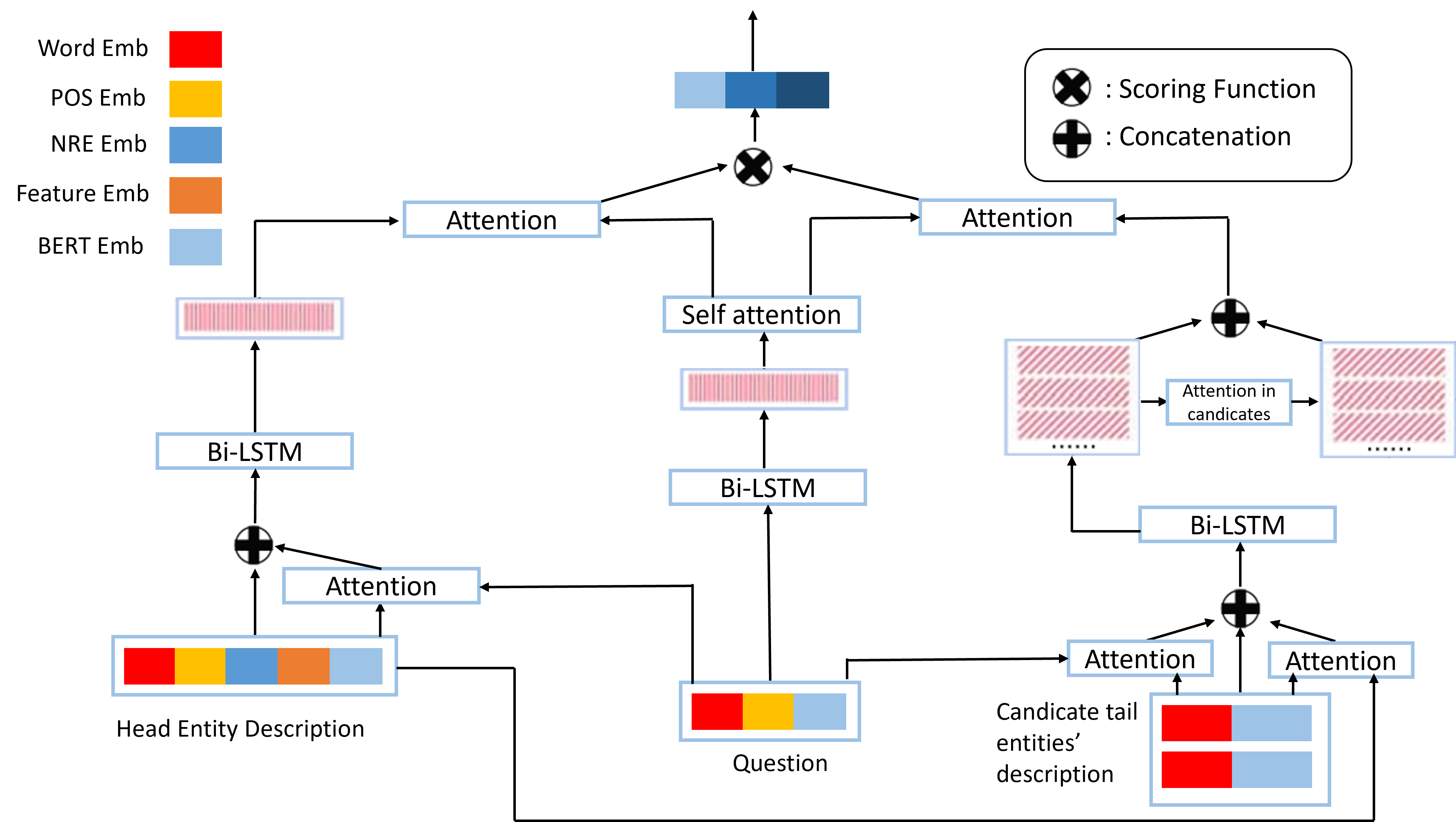}
		\centering
	\caption{The framework of MIA. MIA improves representations by incorporating additional textual features from head entity descriptions.}
	\label{fig:MIA}
\end{figure}

\textbf{MIA} \cite{niu2021open} is an open-world KGC model that utilizes a word-level attention mechanism to simulate interactions between entity descriptions, names, relationship names, and candidate tail entity descriptions. As shown in Fig. \ref{fig:MIA}, it improves representations by incorporating additional textual features from head entity descriptions. However, its effectiveness in predicting the tail entity relies on the availability of relationship-related information in entity descriptions.

\textbf{Discussion.}
Open-world Graph Learning is a dynamic and challenging research area, requiring models to go beyond traditional classification tasks to recognize and categorize previously unseen classes. In graph structures, it involves advanced feature representation learning and the innovative classification of nodes into both known and unseen classes. In the realm of knowledge graphs, the shift towards leveraging open-world resources for KGC marks a significant advancement in making these graphs more complete and reflective of the ever-changing real world. Both areas highlight the need for models that are adaptable, flexible, and capable of handling the complexities and uncertainties inherent in real-world data.

\begin{table}[ht]
 \caption{A summary of graph out-of-distribution (OOD) learning methods.}
    \label{tab:Out-Of-Distribution Graph methods}
\resizebox{0.48\textwidth}{!}{
\begin{tabular}{lll}
\toprule
\textbf{Category}                                                    & \textbf{Subcategory}                       & \textbf{Method}          \\ \midrule \multirow{11}{*}{Graph OOD Generalization}   & \multirow{4}{*}{Data Augmentation} & GAug  \cite{DBLP:conf/aaai/0003LNW0S21}           \\ \cmidrule(r){3-3} 
&                   & NCLA  \cite{shen2023neighbor}        \\ \cmidrule(r){3-3}
&                   & GRAND  \cite{feng2020graph}        \\ \cmidrule(r){3-3} 
&           & GraphCL  \cite{you2020graph}       \\ \cmidrule(r){2-3} 
& \multirow{2}{*}{Model Development}              & DisenGCN \cite{ma2019disentangled}       \\ \cmidrule(r){3-3}
&                    & OOD-GNN \cite{li2022ood}        \\ \cmidrule(r){2-3}
& \multirow{2}{*}{Learning Strategy} & StableGL \cite{zhang2021stable}        \\ \cmidrule(r){3-3}
&          & Pretraining-GNN \cite{hu2019strategies}\\ \midrule
\multirow{7}{*}{Graph OOD Detection}        & \multirow{5}{*}{-}                  & Uncertainty-GNN \cite{zhao2020uncertainty}\\ \cmidrule(r){3-3} 
&                 & GPN \cite{stadler2021graph}            \\ \cmidrule(r){3-3} 
&                   & LMN \cite{huang2022end}            \\ \cmidrule(r){3-3}  
&          & GraphDE \cite{li2022graphde}        \\ \cmidrule(r){3-3} 
&         & GOOD-D \cite{liu2023good}         \\ \midrule 
\multirow{8}{*}{Open-world Graph Learning}                  & \multirow{3}{*}{Graph Structure}   & OpenWGL \cite{wu2020openwgl}        \\ \cmidrule(r){3-3}  
&         & FATE \cite{wu2021towards}           \\ \cmidrule(r){3-3}  
&         & Co-CGE \cite{mancini2022learning}         \\ \cmidrule(r){2-3}
& \multirow{3}{*}{Knowledge Graph}   & ConMask \cite{shi2018open}         \\\cmidrule(r){3-3}
&          & Prob-CBR \cite{das2020probabilistic}       \\ \cmidrule(r){3-3}
&         & MIA \cite{niu2021open}            \\ \bottomrule
                                                    % & \multirow{4}{*}{Test-time Graph Transformation (Adaptation)} & \multirow{4}{*}{}                  & AdaGraph \cite{mancini2019adagraph}        \\ \cline{4-4} 
                                                    % &               &          & GTRANS \cite{jin2022empowering}         \\ \cline{4-4} 
                                                    % &                   &                 & GAPGC \cite{chen2022graphtta}          \\ \cline{4-4} 
                                                    % &                   &                      & GT3 \cite{wang2022test}            \\ \hline
\end{tabular}
}
\end{table}

\section{Graph Continual Learning}
Graph continual learning refers to the problem of learning from a stream of graph data that arrives over time and continuously evolves. In this setting, the goal is to develop models that can learn from new graphs as they arrive, without forgetting what has been learned from previous graphs.
The graph continual learning problem is particularly challenging because the distribution of graphs can change over time, and the model must adapt to these changes without degrading its performance as previously.

% Online learning involves updating the model parameters with each new graph, which allows the model to learn from the most recent data while retaining knowledge from previous graphs. Transfer learning involves using the knowledge learned from previous graphs to facilitate learning on new graphs. Memory-based approaches involve storing previously seen graphs in a memory buffer and using them to enhance the model's performance on future graphs.

Graph continual learning has applications in various fields, such as social network analysis, recommendation systems, and biological network analysis. In these fields, the graph data can arrive in a stream, and it is essential to have models that can adapt to the evolving graph data.

\begin{figure*}[!ht]
	\includegraphics[width=0.85\linewidth]{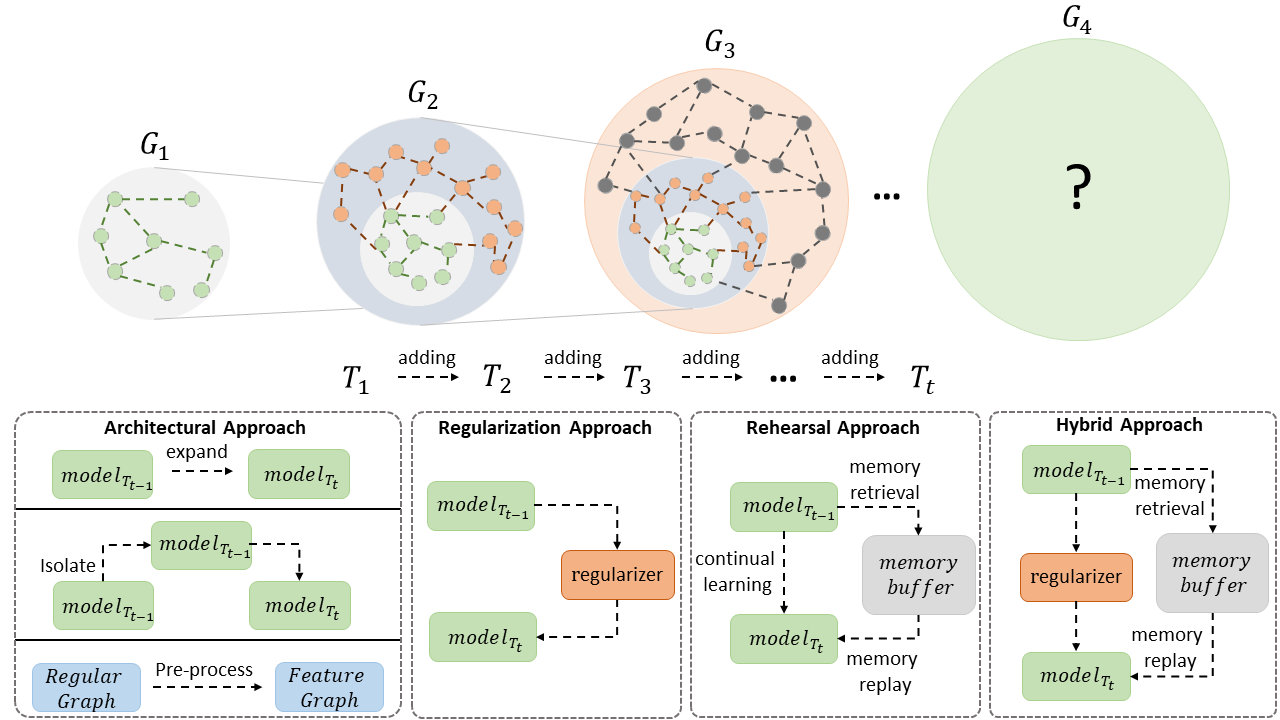}
		\centering
	\caption{Graph evolution and graph continual learning categorization. Graph lifelong learning aims to accommodate a new task $T_t$ based on the graph $G_t=G_{t-1}+\Delta   G_t$ while maintaining the performance in solving previous tasks $T_1,T_2,...,T_{t-1}$. the existing graph continual learning works can be divided into four categories: architectural, regularization, rehearsal, and hybrid.}
	\label{fig:CGNN_categorization}
\end{figure*}

The categorization of graph continual learning methods is based on how these methods consider graph structure data in the incremental learning process. 
%The hybrid approach is also added to categorize the current methods that combine more than one approach. 
Thus, the existing graph continual learning works can be divided into four categories: architectural, regularization, rehearsal, and hybrid, which are as shown in Fig. \ref{fig:CGNN_categorization} and follows:

\begin{itemize}
    \item 
    \textbf{Architectural Approach}. These approaches focus on modifying the specific architecture of networks, activation functions, or layers of algorithms to address a new task and prevent the forgetting of previous tasks.

    \item 
    \textbf{Regularization Approach}. These approaches consolidate the learned knowledge by adding a regularization item to the loss function, constraining the neural weights from updating in a direction that compromises performance on prior tasks.

    \item 
    \textbf{Rehearsal Approach}. These approaches maintain a memory buffer preserving the information of prior tasks and replay it when learning new tasks to mitigate catastrophic forgetting.

    \item 
    \textbf{Hybrid Approach}. These approaches combine more than one continual learning approach to take advantage of multiple approaches and improve the performance of models.
    
\end{itemize}

We summarize the methods of graph continual learning in Table \ref{tab:continual_graph_learning}. 

\begin{table}[!h]
\renewcommand\arraystretch{1.5}
\centering
 \caption{A summary of graph continual learning (GCL) methods.}
    \label{tab:continual_graph_learning}
\resizebox{0.48\textwidth}{!}{
\begin{tabular}{ll}
\toprule
\textbf{Category}   &\textbf{Method} \\ \midrule
\multirow{4}{*}{Architectural (Arch.)}   & FGN  \cite{wang2022lifelong} \\  
 & HPNs \cite{zhang2022hierarchical} \\
& CoST-GCN \cite{Hedegaard2023continual} \\ 
& PI-GNN \cite{Zhang2023continual}\\ \midrule

\multirow{6}{*}{Regularization (Reg.)} & DiCGRL \cite{kou-etal-2020-disentangle}  \\
& GPIL \cite{GPIL2022Tan}  \\
& TWP \cite{Liu2021OCFGN}  \\
& GraphSAIL \cite{xu2020GraphSAIL}  \\
& RieGrace \cite{Sun_Ye_Peng_Wang_Yu_2023}  \\
& LKGE \cite{cui2023lifelong}  \\  \midrule

\multirow{4}{*}{Rehearsal (Reh.)} &  ER-GNN \cite{zhou2021overcoming}  \\
& Incremental-GNNs \cite{Galke2020TGDS}  \\
& RBR and PBR \cite{Perini2022LSGER} \\
& Inverse Degree Sampling \cite{Ahrabian2021} \\ \midrule

\multirow{4}{*}{Hybrid} & ContinualGNN (Reg.\& Reh.) \cite{Wang2020SGNN}  \\
&LDANE (Arch.\& Reg.\& Reh.)~\cite{Wei2019LRL}  \\ 
& TrafficStream (Reg.\& Reh.) \cite{Xu2021TST} \\
& CMDGAT (Arch.\& Reh.) \cite{Anam2023cmdgat}
\\ \bottomrule
\end{tabular}
}
\end{table}

\subsection{Architectural Approach}
In the general continual learning setting, the architectural approaches mainly pay attention to altering the specific architecture of networks, activation functions, or layers of algorithms to address new tasks to prevent catastrophic forgetting of already seen tasks \cite{pmlrELLA2013,DBLP2016,Core502017}.
In graph continual learning scenario, the architectural approaches solve problems by techniques such as graph structure evolution, unit expansion, and compression. %For examples, FGN \cite{wang2022lifelong} convert graph data architecture into regular learning problems, and HPNs \cite{zhang2022hierarchical} which extract various level abstractions of prototypes to accommodate new knowledge.
%

% A general graph learning process needs the full adjacency matrix or the entire graph topology information to propagate node features across all layers. 
% %
% Before traning step, GCNs requires a integral pre-processing step which includes the entire graph dataset. The pre-processing step does not apply to continual incremental setting. Particularly, continual graph learning encounters insurmountable challenges provided by the topology of graphs. The above challenges motivated the proposal of a new suitable graph topology for continual graph learning.
%

\begin{figure}[!ht]
	\includegraphics[width=1\linewidth]{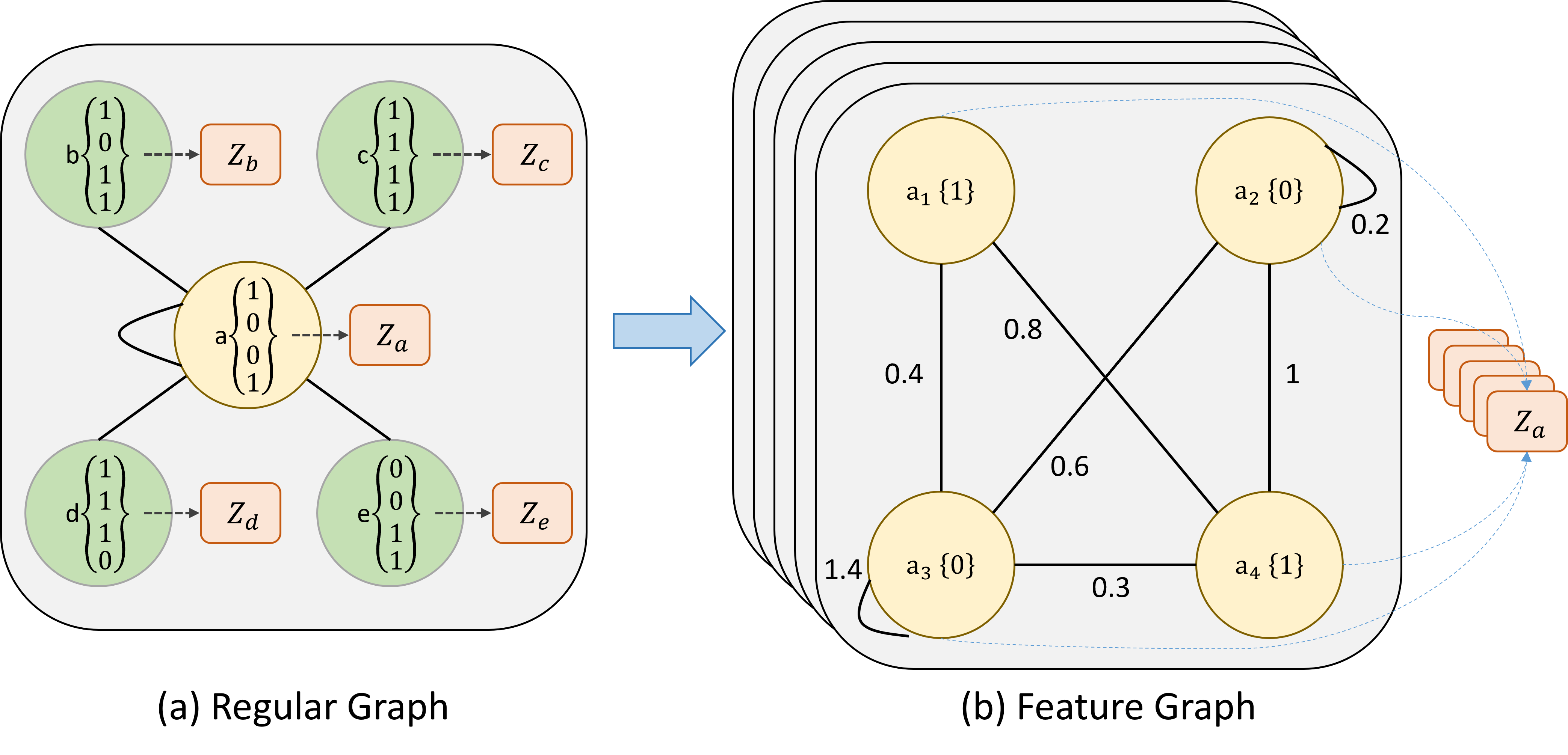}
		\centering
	\caption{The framework of FGN. FGN allows independent processing of graph nodes during knowledge acquisition and has implications for graph relationships.}
	\label{fig: FGN} 
\end{figure}

% \textbf{Feature Graph Networks (FGN)} in continual graph learning by Wang et al. \cite{wang2022lifelong} convert graph architectures into a representation that can be ingested by regular learning architectures. As shown in Figure \ref{fig:FGN}, The FGN architecture modification allows graph problems to be re-framed, by transforming a node classification task into a feature graph classification task. This then enables the feature graphs to be trained in a regular mini-batch manner. By converting graphs into a representation that suits common CNNs tasks, FGN enables the input of the graph’s node to be processed independently whilst new knowledge is being learned, without requiring the entire graph to be pre-processed. This also changes some graph relationships: fixed feature dimensions in the regular graph translate to a fixed number of nodes in the feature graph, and the addition of nodes to the adjacency matrix of the regular graph translates to the addition of adjacency matrices of feature graphs. The cross-correlation of connected feature vectors is accumulated to develop feature interactions that are useful for constructing feature adjacency.
\textbf{FGN} (Feature Graph Networks) \cite{wang2022lifelong} is a framework that converts graph architectures into a format compatible with learning structures. Illustrated in Fig. \ref{fig: FGN}, it allows independent processing of graph nodes during knowledge acquisition and has implications for graph relationships. In feature graphs, the fixed node count corresponds to unchanging feature dimensions in standard graphs. 

\textbf{HPNs} (Hierarchical Prototype Networks) \cite{zhang2022hierarchical} extract abstract knowledge from expanding graphs using prototypes. They encode attributes and relationships, generating prototypes at different levels. These prototypes capture shared attributes and relationships among nodes. HPNs enable the adaptive selection and composition of prototypes, facilitating the extraction of hierarchical knowledge.

% \begin{figure}[!ht]
% 	\includegraphics[width=1\linewidth]{figure/HPNs.png.png}
% 		\centering
% 	\caption{The framework of HPNs. HPNs enable the adaptive selection and composition of prototypes, facilitating the extraction of hierarchical knowledge. The figure is from reference \cite{zhang2022hierarchical}}
% 	\label{fig: HPNs}
% \end{figure}

\textbf{CoST-GCN} \cite{Hedegaard2023continual} is a variant of ST-GCN \cite{yu2018spatio} designed for graph-based human action recognition. It introduces a Continual Inference Network that enables step-by-step predictions without redundant frame processing. CoST-GCN reduces time complexity, accelerates on-hardware, and decreases memory usage during online inference while maintaining predictive accuracy.

\textbf{PI-GNN} (Parameter Isolation GNN) \cite{Zhang2023continual} is a model for continual learning on dynamic graphs. The method employs a strategy by using parameter isolation and expansion. The underlying rationale is rooted in the observation that distinct parameters play varying roles in learning diverse graph patterns. Building upon this insight, the approach expands model parameters to enable the assimilation of emerging graph patterns.

\textbf{Discussion.}
Both in general continual learning and graph continual learning scenarios, architectural approaches are key to managing the balance between learning new tasks and retaining old information. While the general continual learning setting often focuses on network-level adaptations, graph continual learning requires specific techniques tailored to the unique nature of graph-based data, such as evolving graph structures or managing graph complexity through expansion and compression. These approaches underscore the necessity for adaptable and flexible models capable of handling the dynamic nature of continuous learning environments.

\subsection{Regularization Approach}
This approach implements a single model and has a fixed capacity by leveraging the loss function using the loss term to help consolidate knowledge in the learning process for new tasks and retain previous knowledge \cite{Li2016Lwf, James2017OCF}. Prior knowledge of graph structures and tasks will be maintained to achieve stable performance while learning novel knowledge. %In graph-based learning, some current works that implement regularization approaches are DiCGRL \cite{kou-etal-2020-disentangle}, GPIL \cite{GPIL2022Tan}, TWP\cite{Liu2021OCFGN}. 

\begin{figure}[!h]
    \centering
    \subfigure[The disentangle module of DiCGRL.]{
        \label{fig:DiCGRL1}
        \includegraphics[width=0.85\linewidth]{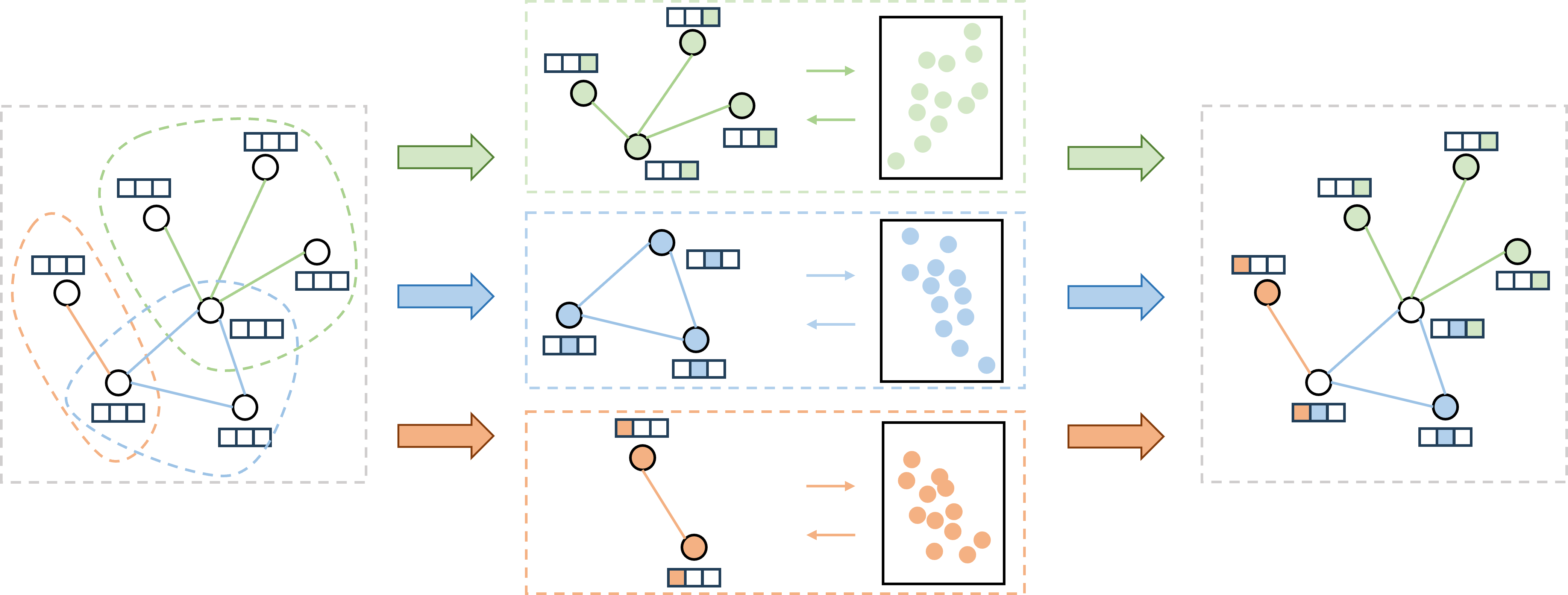}
    }
    \quad
    \subfigure[The updating module of DiCGRL.]{
        \label{fig:DiCGRL2}
        \includegraphics[width=0.85\linewidth]{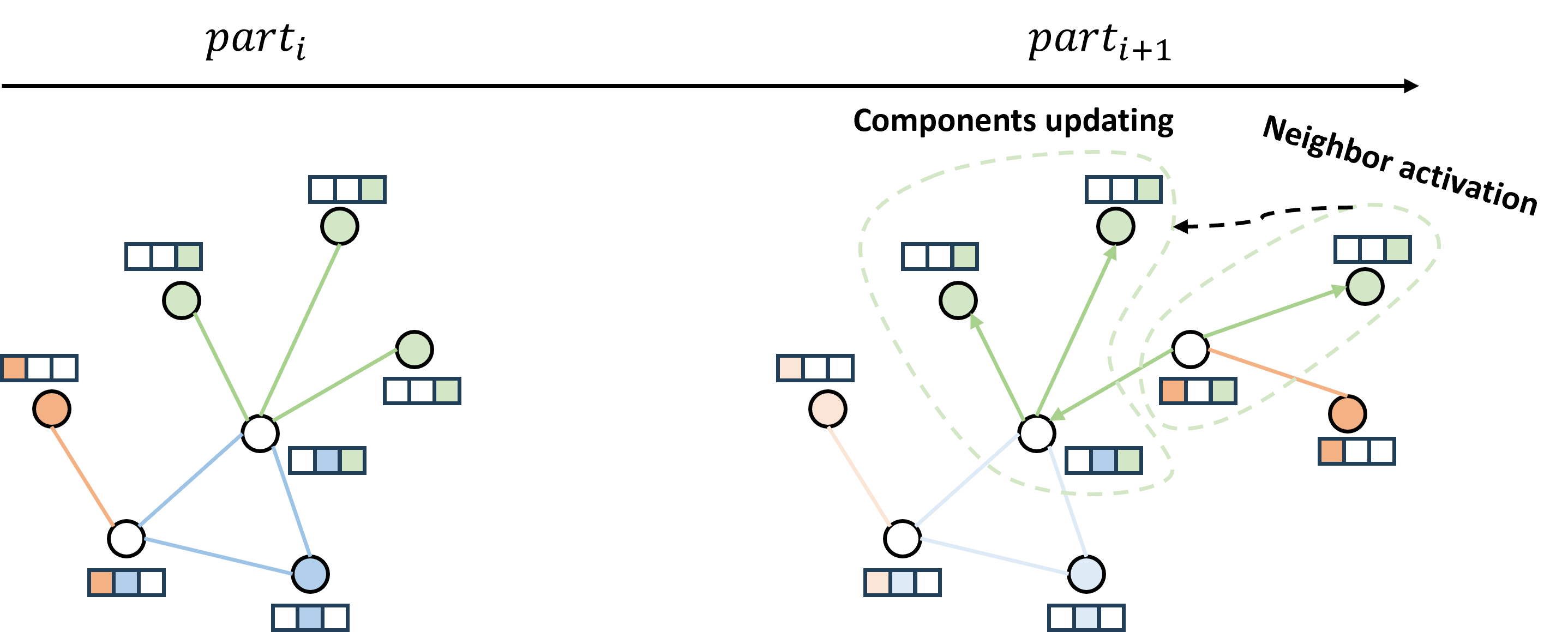}
    }
    \caption{The framework of DiCGRL. DiCGRL takes two approaches to continuously learn a new graph embedding whilst preventing the forgetting of old knowledge. }
    \label{fig:enter-label}
\end{figure}

\textbf{DiCGR} \cite{kou-etal-2020-disentangle} is an approach designed to support continual learning on graph data. It addresses the challenge of preserving previously acquired knowledge while learning new information. DiCGRL achieves it by breaking down relational triplets into individual components, refining the embeddings to better capture the graph structure, and incorporating a regularization mechanism to prevent forgetting of previously learned knowledge. DiCGRL takes two approaches to continuously learn a new graph embedding whilst preventing the forgetting of old knowledge, respectively the disentangle module shown in Fig. \ref{fig:DiCGRL1} and the updating module shown in Fig. \ref{fig:DiCGRL2}.

\textbf{GPIL} \cite{GPIL2022Tan} is a framework that aims at acquiring proficiency in different classes. It utilizes a recurrent strategy, pre-training the encoder on base classes and maintaining it during pseudo incremental learning. To mitigate catastrophic forgetting, this approach introduces a Hierarchical Attention Graph Meta Learning framework with augmented loss function using regularization techniques.

% \textbf{Topology-aware Weight Preserving (TWP)} by Liu et al. \cite{Liu2021OCFGN} implements a novel method to strengthen lifelong learning and minimize catastrophic forgetting in GNNs. TWP explicitly studies the local structure of the input graph and stabilizes the important parameters in topological aggregation. Given an input graph and its embedding feature of nodes, TWP estimates the important score of each network parameter based on their contribution to the topological structure and task-related performance. The methods used in TWP are to calculate the gradients of the task-wise objective and topological preserving with each parameter, then consider the gradient as an index for the parameter importance. After learning previous tasks, the model gets the optimal parameters by minimizing the loss on the task. As not all parameters contribute equally, it is important to preserve the minimized loss by considering highly influential parameters. The approximate contributions of each parameter are calculated based on an infinitesimal change in each parameter. Parameters that significantly contribute to minimizing loss must be kept stable when learning future tasks. Besides minimized loss preserving, topological structure preservation is conducted since structure information in the graph plays an important role. It aims to find the parameters with a substantial contribution to learning the topological information of the graph.

\begin{figure}[!ht]
	\includegraphics[width=1\linewidth]{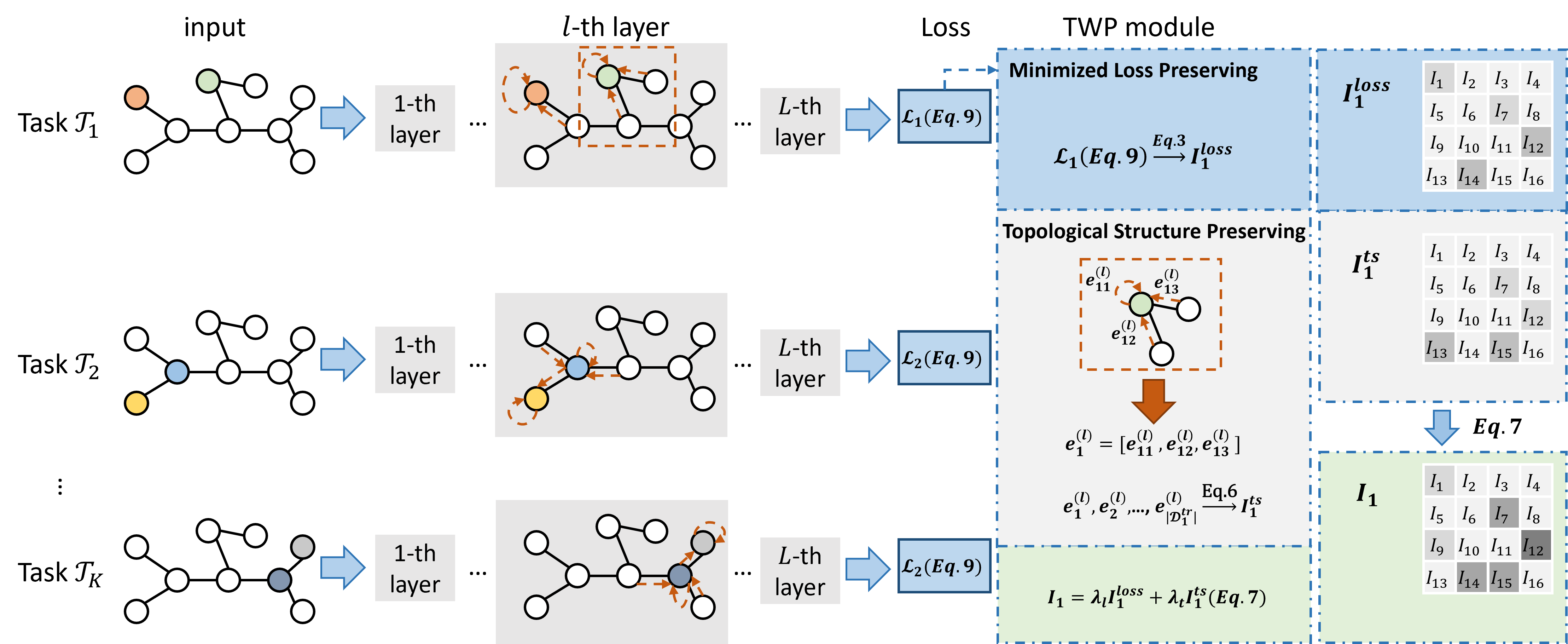}
		\centering
	\caption{The framework of TWP. TWP leverages the gradients of task-specific objectives alongside topological preservation in relation to each parameter.}
	\label{fig:TWP}
\end{figure}

\textbf{TWP} \cite{Liu2021OCFGN} is a framework designed to bolster continual learning while mitigating the risk of catastrophic forgetting within GNNs.As shown in Fig. \ref{fig:TWP}, TWP computes the importance score for each network parameter, rooted in its contribution to both the graph's topological architecture and its effectiveness in task execution. TWP leverages the gradients of task-specific objectives alongside topological preservation in relation to each parameter. 

\textbf{GraphSAIL} \cite{xu2020GraphSAIL} is an Incremental Learning framework for GNN-based recommender systems. It enables efficient updates, reducing computation time for more frequent recommendations. GraphSAIL implements inter-distillation to evaluate the agreement of high-level node encoding between the teacher and student models, enhancing the performance of its recommendation system.

\textbf{RieGrace} \cite{Sun_Ye_Peng_Wang_Yu_2023} is a self-supervised Riemannian Graph Continual Learner. It incorporates an Adaptive Riemannian GCN that shapes the Riemannian space using learned curvature adaptive to each graph. Additionally, it utilizes a Label-free Lorentz Distillation approach with teacher-student AdaRGCN models for the graph sequence. Overall, RieGrace enables graph continual learning in adaptive Riemannian spaces through self-supervision.

\textbf{LKGE} \cite{cui2023lifelong} is a model for lifelong learning in expanding knowledge graphs. It utilizes a masked KG autoencoder, embedding transfer strategy, and embedding regularization to efficiently transfer and retain knowledge. It streamlines the process of incorporating new data while preserving existing knowledge in KG embeddings.

\textbf{Discussion.}
 The regularization approach in continual learning, particularly for graph-based models, is a strategy that emphasizes the importance of retaining prior knowledge while acquiring new information. By carefully modifying the loss function, this approach seeks to maintain stable performance across a series of tasks without expanding the model's capacity. This is particularly crucial in graph-based applications where the relationships and structures in the data are complex and evolving.

\subsection{Rehearsal Approach}
This approach regulates retraining processes for previous tasks to strengthen the relationship between memory and performance on previously learned tasks \cite{Rebuffi2016iCaRLIC}. This graph learning approach enables the selection of appropriate samples of graph representation, such as nodes and edges, for retraining purposes. The number of samples is carefully considered to minimize the computational complexity. 
%Recently, graph lifelong learning methods using rehearsal approaches have been proposed, such as ER-GNN \cite{Zhou2020Continual}, and lifelong Open-world Node Classification \cite{Galke2020TGDS}.

% \textbf{Experience Replay GNN Framework}
% by Zhou et al. \cite{Zhou2020Continual} implements the concept of lifelong learning on graph data using graph-based experience replay called ER-GNN. The main objective is to mitigate the catastrophic forgetting problem in graph data while learning continuously over time. The model preserves past knowledge in an experienced buffer obtained from previous learning of tasks to be replayed while learning a new task. This model implements an experience node strategy to help to select the experience nodes to be stored in the experience buffer, Apart from those experience node strategies, ER-GNN proposed a new method called influence maximization (IM). The model’s experience node replay is influenced by Complementary Learning System (CLS) that supports the biological learning process in humans. CLS is a great example of a complementary contribution to the virtual experience mechanism of the hippocampus and neocortex in the human brain system. It helps consolidate knowledge by replaying memories in the hippocampus before being transferred into long-term memory using the neocortical system.

% \begin{figure*}[!ht]
% 	\includegraphics[width=0.85\linewidth]{figure/GPIL.png}
% 		\centering
% 	\caption{The updating module of DiCGRL.}
% 	\label{fig:GPIL}
% \end{figure*}

\textbf{ER-GNN} \cite{zhou2021overcoming} is a framework that implements a graph-based experience replay mechanism. This framework's core principle involves preserving historical knowledge within an experience buffer, drawn from the preceding task learning phases. ER-GNN prevents catastrophic forgetting by retaining previously learned knowledge using an experienced buffer. It selects and stores relevant nodes in the buffer using an experience node strategy.

% \textbf{Lifelong Open-world Node Classification}
% by Galke et al. \cite{Galke2020TGDS} implements continual open-world classification to implement incremental training on graph data. This model could rely on previous knowledge to help the learning process in a sequence of tasks. The learned knowledge is stored as historic data explicitly or within model parameters implicitly. The model then analyses the influence of stored knowledge that is helpful in refining the network. The model’s objective is to address the challenge in graph learning that nodes in graph data can not be processed independently because the relation of nodes needs to be processed through an embedding mechanism to share the information across the nodes. The second objective is to enable the concept of open-world classification to implement incremental training in graph data and accommodate new tasks that are completely different from the previous ones. The model introduces a new measurement to regulate the evolution in graph data with different characteristics to address that objective. The method has the capability to integrate an isotropic and anisotropic type of GNN.

\textbf{Incremental-GNNs} \cite{Galke2020TGDS} is a model specifically designed for incremental training on graph data. The model utilizes prior knowledge to enhance the learning process across a sequence of tasks. This acquired knowledge can be explicitly stored as historical data or implicitly encoded within the model’s parameters. The model then evaluates the impact of this accumulated knowledge, leveraging it to improve the performance of the network.
% \begin{figure*}[!ht]
% 	\includegraphics[width=0.85\linewidth]{figure/GPIL.png}
% 		\centering
% 	\caption{The updating module of DiCGRL.}
% 	\label{fig:GPIL}
% \end{figure*}

\textbf{RBR and PBR} (Random-Based Rehearsal and Priority-Based Rehearsal) 
\cite{Perini2022LSGER} are methods for training GNNs on streaming graph data. They use experience replay with a node sampling strategy that prioritizes higher loss samples. This allows for continuous learning and adaptation to changing dynamic graph streams, aiming for faster training, low latency, and high accuracy.

\textbf{Inverse Degree Sampling} \cite{Ahrabian2021} is an approach in graph-based recommender systems that focuses on selecting interaction subsets using an inverse degree experience sampling strategy. By redirecting the model's attention to interactions involving users with infrequent interactions, this strategy enhances the applicability of reservoir sampling in continual sampling contexts.

\textbf{Discussion.}
In graph continual learning, the rehearsal approach offers a viable method for reinforcing the model's memory of previous tasks while learning new ones. By judiciously retraining on selected graph samples, this approach helps maintain a balance between retaining past knowledge and acquiring new information, ensuring sustained performance across a series of evolving tasks. This strategy is particularly relevant in dynamic environments where graph data continuously changes or expands.

\subsection{Hybrid Approach}
The hybrid approach combines more than one continual learning approach to take advantage of each approach and maximize the performance of models \cite{Lopezpaz2017GEM, Maltoni2018CL}. %Examples of the graph lifelong learning models that implement hybrid approaches discussed in this section include ContinualGNN \cite{Wang2020SGNN}, LDANE \cite{Wei2019LRL}, and TrafficStream \cite{Xu2021TST}.

% \textbf{ContinualGNN} by Wang et al. \cite{Wang2020SGNN} implements hybrid approach to solve catastrophic forgetting. There is a concept of replaying strategy to refine the network as a complement to the regularization approach itself. It combines both approaches to mitigate catastrophic forgetting and maintain the existing learned pattern. The first main goal of this model is to detect a new pattern in graph structure that significantly influences all nodes in the network. Second, the model aims to consolidate the knowledge in the entire network. The ContinualGNN model captures new patterns in streaming graph data. The model proposes a method based on a propagation process to efficiently mine the information on affected nodes when learning new patterns. The existing knowledge from the propagation process is then maintained using a combination of both approaches of rehearsal and regularization strategies.

\textbf{ContinualGNN} \cite{Wang2020SGNN} is a framework that combines two distinct strategies: a replaying strategy and a regularization approach, functioning synergistically to mitigate catastrophic forgetting and maintain the established learned patterns. It identifies and incorporates novel patterns in the evolving graph structure while solidifying knowledge to maintain acquired insights throughout the network.

% \textbf{Lifelong Dynamic Attributed Network Embedding (LDANE)} by Wei et al. \cite{Wei2019LRL} aims to represent learning to produce a low-dimensional vector for each node in a growing size network. LDANE is developed based on lifelong learning that uses an architectural approach, namely Dynamic Expandable Networks (DEN) \cite{Lee2017LLDE}. DEN can dynamically determine network capacity to learn and share the structure among different tasks. Moreover, LDANE constructs attribute constraints that can be updated efficiently when the node attribute changes to restrict the learned embedding to satisfy the constraints. In this method, a Deep autoencoder is employed to learn the embedding of nodes.

\begin{figure}[!ht]
	\includegraphics[width=1\linewidth]{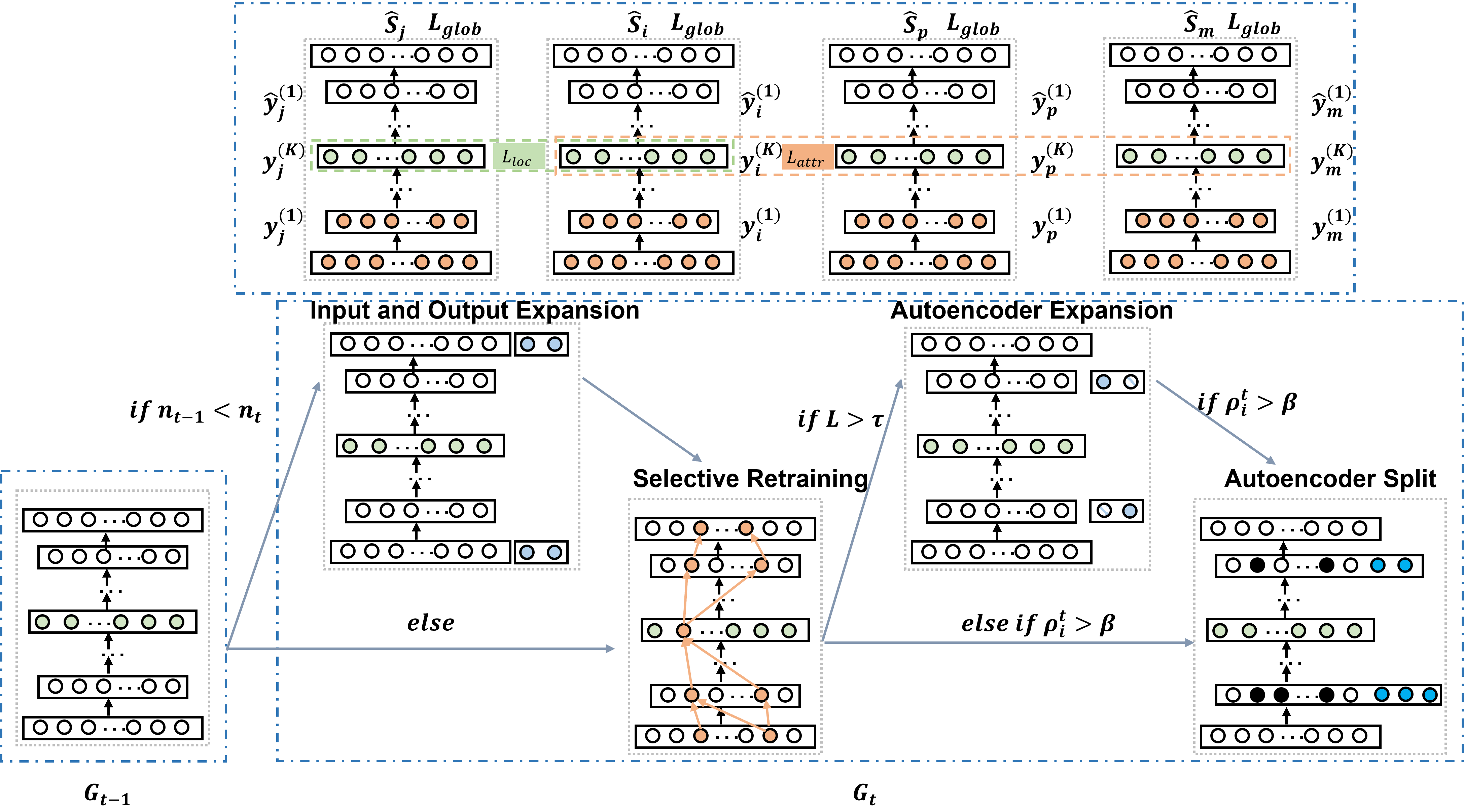}
		\centering
	\caption{The framework of LDANE. LDANE incorporates attribute constraints to align learned embeddings with specific attributes, adapting to changes in node attributes.}
	\label{fig: LDANE}
\end{figure}

\textbf{LDANE} \cite{Wei2019LRL} is a framework that acquires low-dimensional representations for nodes in an expanding network. Underpinned by an architectural paradigm known as Dynamic Expandable Networks \cite{Lee2017LLDE}, LDANE advances the field by integrating attribute constraints. As shown in Fig. \ref{fig: LDANE}, it incorporates attribute constraints to align learned embeddings with specific attributes, adapting to changes in node attributes.

\textbf{TrafficStream} \cite{Xu2021TST} is a framework that captures patterns for new graph structures and consolidates knowledge between previous and new ones. It uses a traffic flow forecasting model (SurModel) to represent intelligent transportation system data. It efficiently updates current network patterns through processes that consider network structure changes caused by factors like the addition of new sensors or stations.

\textbf{CMDGAT} \cite{Anam2023cmdgat} is a model that enhances the descriptive capacity of point clouds by using a novel graph continual network architecture. The architecture includes an attention mechanism that leverages knowledge from previous point cloud pairs to associate and aligns points in the current pair. By utilizing insights from past associations, the network optimally registers points, improving the overall representation of the point clouds.

\textbf{Discussion.}
The hybrid approach in graph continual learning offers a versatile and effective way to tackle the challenges of learning in dynamic and evolving environments. By combining the strengths of various continual learning methods, it aims to achieve robust performance, adaptability, and efficient use of computational resources. This approach is particularly relevant in complex learning scenarios where no single method can adequately address all challenges.

\section{Applications}
\subsection{AI-aided Drug Discovery.}
AI-aided drug discovery~\cite{ji2023drugood} is becoming increasingly prevalent when it offers the great potential to accelerate the development of new pharmaceuticals, making the process faster, more cost-effective, and more efficient. 
It has wide application scenarios across various domains, including generative chemistry~\cite{sliwoski2014computational}, protein structure prediction~\cite{gao2020deep,jumper2021highly}, virtual screening~\cite{lim2019predicting,karimi2019deepaffinity}, drug repurposing for emerging diseases~\cite{morselli2021network}, and many others.
Due to the distribution shift issues of graph data, the compatibility assumption between training and test sets does not usually hold in the drug discovery-related fields, especially when the learned knowledge from the training set may not smoothly provide accurate predicting signals for test samples~\cite{wang2022test,ji2023drugood,jin2022empowering,chen2022graphtta,li2022ood,li2022graphde}.

For instance, when performing virtual screening for hit finding~\cite{ji2023drugood}, the prediction model is typically trained on known target proteins. However, a `black swan' event like COVID-19 can occur, resulting in a new target with unseen data distribution. The performance on the new target will significantly degrade.
Moreover, for molecular prediction~\cite{li2022ood}, the GNN models trained on molecules with one group of scaffolds, i.e., the two-dimensional structural frameworks of molecules, might learn the spurious correlations between the scaffolds and labels, so that the GNN models would make incorrect predictions on real-world out-of-distribution testing molecules with different scaffolds.

Hence, dealing with distribution shifts on graph data is a critical way to address the performance degradation for practical AI-aided drug discovery. The potential solutions contain but are not limited to designing robust and generalizable graph learning models through domain adaption, OOD modeling, and continuing graph learning; Meanwhile, developing curated distribution shifted graph datasets and benchmarks to better understand distribution shift cases and characteristics for facilitating and promoting the development of drug discovery.

%DrugOOD: Out-of-Distribution (OOD) Dataset Curator and Benchmark for AI-aided Drug Discovery

\subsection{Personalized Intelligent Recommendation Systems.}
In the recommender system, the training samples with nodes and relations are usually observed and collected with a certain selection bias~\cite{zhang2021stable}, that is, the logged interactions collected from current undergoing systems may warp (graph) recommenders biased towards popular items, leading to potential distribution shifts on the deployed environment with caused unfairness towards less exposed items.
Moreover, considering male users often have fewer logged interactions on e-commerce platforms, the graph recommenders may be trained with a bias towards female users and thus deteriorate the experience of male users.

Hence, explicitly reducing the performance variances across different deployment environments with distribution shifts is the key to the graph learning based recommender system, for confronting the inconsistencies between the training and testing distributions, and
achieving fairness for different groups of users. Moreover, due to the continuous and frequent updating characteristic of recommender systems, tackling the catastrophic forgetting problem that occurs when training a model in an incremental pattern would benefit modeling a user's long-term preference (or an item’s long-term property) under the evolving distribution shifts~\cite{xu2020GraphSAIL,zhang2022hierarchical, GPIL2022Tan,Ahrabian2021}.

In this way, various graph learning models dealing with distribution shifts would benefit real-world recommender systems to make more adaptive and stable predictions of practical unseen users' preferences, leading to more accurate and intelligent recommendation~\cite{corrabs190510095,jin2022empowering,zhang2021stable}.

\subsection{Open-world Knowledge Exploration and Learning.}
Real-world information representation and related applications usually rely on knowledge-driven approaches that utilize knowledge graphs (KGs) with comprehensive knowledge exploration and learning.
Throughout the whole life-cycle of KG, new facts, unseen entities, and unseen relations continually emerge, evolve, and grow alongside
the development, leading to inevitable dynamic distribution shifts~\cite{das2020probabilistic,cui2023lifelong}. 
In this case, conducting open-world knowledge exploration and investigating lifelong KG learning under such temporal distribution shifts, would facilitate the learning and updating from growing new knowledge, facts, and relations, meanwhile, preserving the previously learned useful old knowledge and serving for various KG-related application scenarios~\cite{ji2021survey}, covering KG representation learning, KG question and answering, KG completion~\cite{wang2023survey}.

Moreover, such knowledge exploration in open-world KG under distribution shifts also provides a great chance to collaborate existing KG learning with the recent prevalent large language models (LLMs)~\cite{pan2023unifying,luo2023reasoning}, considering LLMs and KGs are inherently interconnected and can mutually enhance each other. For instance, DAGNN~\cite{wu2019domain} conducts cross-domain text classification by modeling documents as graph data structure, and combining the hierarchical structure and domain labels into a unified deep model. And ChatRule~\cite{luo2023chatrule} unleashes the power of LLMs for mining logical rules over knowledge graphs for enhancing the reasoning ability. 
It can be anticipated that open-world knowledge exploration and learning with KG related temporal shift graph learning would be capable of sophisticated reasoning and decision-making based on a deep understanding of real-world knowledge, ultimately driving forward the frontiers of AI research and its practical applications.

\subsection{Intelligent Transportation System Forecasting.}
Traffic flows typically show high-quantity, long-term streaming evolution, and gradual expansion characteristics in traffic networks. Accurately forecasting traffic flows plays an important role in improving the efficiency of intelligent transportation systems~\cite{Xu2021TST}. 
Considering the traffic networks would change flows of nodes and expanding structures in a long-term period indicated by spatial-temporal graphs~\cite{jin2022neural,jin2022multivariate}, the distribution shift in the inconsistency between patterns (context attribute shift and the spatial structure shift) of new data and historical data would negatively affect the accuracy of forecasting.
Hence, the key to conducting graph learning on intelligent transportation system forecasting under such distribution shifts is to jointly incorporate long-term stable patterns and temporarily disappearing patterns, for efficiently capturing patterns from new data while consolidating historical knowledge using graph continual learning approaches.

Hence, the ongoing development of graph learning models under distribution shifts with continual learning holds great promise for the future of transportation management and optimization. It can be anticipated that we can enhance the accuracy of traffic flow predictions and ultimately contribute to the efficiency and effectiveness of intelligent transportation systems. 

\section{Future Directions}
\subsection{Graph Data-centric Learning under Distribution Shifts.}
Very recently, there has been a growing interest in data-centric graph machine learning (DC-GML) for the recognition and prioritization of the importance of graph data~\cite{zheng2023towards}. 
Modeling graph data shifts in cross-domains and various distributions~\cite{guo2023data} is an important aspect of DC-GML, as the domain and distribution shifts of graph data would significantly affect the generalization ability and robustness of graph learning algorithms.
For one thing, there is some recent research focusing on exploiting the distinct graph distribution characteristics, emphasizing invariance, stability, evolution, etc.~\cite{wu2021handling,sui2023unleashing,jin2022empowering}.
For another thing, a very young research topic graph learning model evaluation problem~\cite{zheng2023gnnevaluator} is explored to observe and predict the GNN model performance under various graph data distribution shifts (e.g., domain changes) without test-time ground truth labels.

Therefore, emphasizing the investigation of distribution shifts in graph data-centric characteristics, while also tackling the co-development of distribution-shifted graph data and models, holds great potential as a future research avenue. 
This approach ensures that optimization and customization are applied to both elements, catering to a wide range of practical scenarios in graph learning and analysis. Consequently, it facilitates a comprehensive understanding of graph data distribution shifts and enhances the design and learning process of graph models, ultimately resulting in improved performance in data-centric graph machine learning.

\subsection{ Cross-modality Distribution Shift Exploration on Graph Learning.}
Data types in the real world exhibit various modalities and scales, e.g., images, texts, tabular data, audios, and graphs, which are crucial for the advancement of artificial intelligence and machine learning communities~\cite{ektefaie2023multimodal}.
Considering the complexity of cross-modal data, learning algorithms are used to make inferences for test data they have not encountered during training, leading to inevitable inductive bias. 
Meanwhile, there are natural and more general distribution shifts among modalities used during model training and deployment.
In such cases, as an important and commonly used structural data type, graph data might not be explicitly given for training and inference in real-world multi-modal model development~\cite{yoon2023multimodal}. 

Hence, constructing graphs from multi-modal data with different node representations, relationships, and attributes, can be a promising future research direction under potentially more complex distribution shifts, such as scene graph construction from images~\cite{guo2023visual}, and knowledge graph construction from texts~\cite{luo2023reasoning,wang2023knowledge,wu2019domain}. Meanwhile, another crucial future problem lies in conducting effective graph learning on these constructed graphs under distribution shifts. This involves thoroughly incorporating cross-modal dependencies. For example, this could entail learning about structural shifts using flexible message propagation strategies and adaptively transferring learned node or graph representations across various distributions and downstream tasks.

\subsection{Comprehensive Graph Learning Task-driven Evaluation Protocols.}
Given the complex and diverse nature of graph data distribution shifts, there is still potential to enhance the evaluation protocols used in various downstream tasks of graph learning for future exploration. These improved protocols should focus on two key aspects: at the model level, to measure performance across different tasks, and at the graph data level, to assess data distribution driven by both the graph learning models and the tasks they are applied to.
For instance, in terms of OOD generalization baselines, graph learning models should be evaluated on both near and far out-of-distribution graph datasets, which may originate from different domain sources. An instance of this would be training models on citation networks and then evaluating their performance on social networks~\cite{huang2023prodigy}.

Therefore, future research should concentrate on developing effective metrics to evaluate the distance or discrepancy among various graph data, considering both node-level analysis in a single graph and collective graph-level analysis in an entire dataset. Additionally, it is also important to focus on establishing fair and uniform methods to compare and evaluate graph learning models under varying degrees of distribution shifts~\cite{zheng2023gnnevaluator}. This should encompass diverse scenarios including graph domain adaptation, OOD generalization, detection, prediction, etc.

\subsection{Trustworthy Graph Learning Under Distribution Shifts.}
Graph data distribution shifts impose significant challenges to various aspects of trustworthiness in graph learning~\cite{zhang2022trustworthy}, including adversarial robustness under graph distribution related attacks, model explainability to diverse and complex graph distributions, as well as privacy and fairness guarantee under different distribution transfer or domain adaption learning.

From the \textit{adversarial robustness} point of view~\cite{wu2023graphguard}, when small but intentional adversarial perturbations are imposed on the input graphs, these alterations can lead to even more significant and unpredictable shifts in graph data distributions~\cite{zhang2023demystifying,zhang2022projective}. 
As a result, this makes graph learning models more vulnerable and less resistant to adversarial attacks. 
Hence, enhancing the robustness of graph learning models under adversarial attacks on graph data distributions could be a valuable direction to explore for future research.

Moreover, considering the \textit{explainability}~\cite{ying2019gnnexplainer}, future graph learning models are expected to have the ability to interpret or explain their decisions or outputs when models are dealing with different graph distributions with diverse characteristics~\cite{liu2023towards}.

Furthermore, in terms of the \textit{privacy and fairness}~\cite{jin2023survey,zhang2023interaction}, when graph data comes from sensitive personal sources or a certain graph learning model is trained on one specific distribution domain, how to maintain privacy across different domains within different graph data distributions is still a challenging aspect for privacy-guaranteed graph learning. Besides, how to ensure that the predictions of graph learning models do not systematically disadvantage any particular group or distribution with fairness is another promising future research question when deployed across varied graph data domains.
%, including adversarial robustness, explainability of different distributions, privacy in different domains, and fairness for distribution transferring.

%\noindent\textbf{Open-world Graph Distribution Shift Recognition and Exploration}
\section{Conclusion}
In this work, we present an in-depth review and synthesis of the forefront approaches and methodologies in graph learning under distribution shifts. We categorized graph learning methods into graph domain adaptation learning, graph out-of-distribution learning, and graph continual learning, based on their observability of distributions and the availability of supervision information, where each with a detailed taxonomy and discussion of the current advancements. The comprehensive analysis, along with the highlighted discussions on potential applications and future research directions, not only provides a clear road map of the state-of-the-art approaches, but also help shed light on the effective development of graph learning techniques with diverse and complex distribution shifts.
% \section*{Acknowledgment}
% This research is sponsored by the US National Science Foundation under grant Nos. IIS-1763452, CNS-1828181, and IIS-2027339. S. Pan is supported by an ARC Future Fellowship (No. FT210100097).
%The authors would like to thank...

% Can use something like this to put references on a page
% by themselves when using endfloat and the captionsoff option.
\ifCLASSOPTIONcaptionsoff
  \newpage
\fi

\balance
\bibliographystyle{IEEEtran}
\bibliography{bib}

% Generated by IEEEtran.bst, version: 1.14 (2015/08/26)
\begin{thebibliography}{100}
\providecommand{\url}[1]{#1}
\csname url@samestyle\endcsname
\providecommand{\newblock}{\relax}
\providecommand{\bibinfo}[2]{#2}
\providecommand{\BIBentrySTDinterwordspacing}{\spaceskip=0pt\relax}
\providecommand{\BIBentryALTinterwordstretchfactor}{4}
\providecommand{\BIBentryALTinterwordspacing}{\spaceskip=\fontdimen2\font plus
\BIBentryALTinterwordstretchfactor\fontdimen3\font minus
  \fontdimen4\font\relax}
\providecommand{\BIBforeignlanguage}[2]{{%
\expandafter\ifx\csname l@#1\endcsname\relax
\typeout{** WARNING: IEEEtran.bst: No hyphenation pattern has been}%
\typeout{** loaded for the language `#1'. Using the pattern for}%
\typeout{** the default language instead.}%
\else
\language=\csname l@#1\endcsname
\fi
#2}}
\providecommand{\BIBdecl}{\relax}
\BIBdecl

\bibitem{Wang2020SGNN}
J.~Wang, G.~Song, Y.~Wu, and L.~Wang, ``Streaming graph neural networks via
  continual learning,'' in \emph{Proceedings of the 29th ACM international
  conference on information \& knowledge management}, 2020, pp. 1515--1524.

\bibitem{Wei2019LRL}
H.~Wei, G.~Hu, W.~Bai, S.~Xia, and Z.~Pan, ``Lifelong representation learning
  in dynamic attributed networks,'' \emph{Neurocomputing}, vol. 358, 05 2019.

\bibitem{pilanci2020domain}
M.~Pilanci and E.~Vural, ``Domain adaptation on graphs by learning aligned
  graph bases,'' \emph{IEEE Transactions on Knowledge and Data Engineering},
  2020.

\bibitem{dai2019network}
Q.~Dai, X.-M. Wu, J.~Xiao, X.~Shen, and D.~Wang, ``Graph transfer learning via
  adversarial domain adaptation with graph convolution,'' \emph{IEEE
  Transactions on Knowledge and Data Engineering}, vol.~35, no.~5, pp.
  4908--4922, 2022.

\bibitem{corrabs190510095}
Y.~Ouyang, B.~Guo, X.~Tang, X.~He, J.~Xiong, and Z.~Yu, ``Learning cross-domain
  representation with multi-graph neural network,'' \emph{CoRR}, vol.
  abs/1905.10095, 2019.

\bibitem{wang2022test}
Y.~Wang, C.~Li, W.~Jin, R.~Li, J.~Zhao, J.~Tang, and X.~Xie, ``Test-time
  training for graph neural networks,'' \emph{arXiv preprint arXiv:2210.08813},
  2022.

\bibitem{chen2022graphtta}
G.~Chen, J.~Zhang, X.~Xiao, and Y.~Li, ``Graphtta: Test time adaptation on
  graph neural networks,'' \emph{arXiv preprint arXiv:2208.09126}, 2022.

\bibitem{jin2022empowering}
W.~Jin, T.~Zhao, J.~Ding, Y.~Liu, J.~Tang, and N.~Shah, ``Empowering graph
  representation learning with test-time graph transformation,'' \emph{arXiv
  preprint arXiv:2210.03561}, 2022.

\bibitem{li2022ood}
H.~Li, X.~Wang, Z.~Zhang, and W.~Zhu, ``Ood-gnn: Out-of-distribution
  generalized graph neural network,'' \emph{IEEE Transactions on Knowledge and
  Data Engineering}, 2022.

\bibitem{you2020graph}
Y.~You, T.~Chen, Y.~Sui, T.~Chen, Z.~Wang, and Y.~Shen, ``Graph contrastive
  learning with augmentations,'' \emph{Advances in neural information
  processing systems}, vol.~33, pp. 5812--5823, 2020.

\bibitem{li2022graphde}
Z.~Li, Q.~Wu, F.~Nie, and J.~Yan, ``Graphde: A generative framework for
  debiased learning and out-of-distribution detection on graphs,''
  \emph{Advances in Neural Information Processing Systems}, vol.~35, pp.
  30\,277--30\,290, 2022.

\bibitem{zheng2023spatio}
C.~Zheng, X.~Fan, S.~Pan, H.~Jin, Z.~Peng, Z.~Wu, C.~Wang, and S.~Y. Philip,
  ``Spatio-temporal joint graph convolutional networks for traffic
  forecasting,'' \emph{IEEE Transactions on Knowledge and Data Engineering},
  2023.

\bibitem{GPIL2022Tan}
Z.~Tan, K.~Ding, R.~Guo, and H.~Liu, ``Graph few-shot class-incremental
  learning,'' in \emph{Proceedings of the Fifteenth ACM International
  Conference on Web Search and Data Mining}, 2022, pp. 987--996.

\bibitem{zhang2022hierarchical}
X.~Zhang, D.~Song, and D.~Tao, ``Hierarchical prototype networks for continual
  graph representation learning,'' \emph{IEEE Transactions on Pattern Analysis
  and Machine Intelligence}, 2022.

\bibitem{mancini2019adagraph}
M.~Mancini, S.~R. Bulo, B.~Caputo, and E.~Ricci, ``Adagraph: Unifying
  predictive and continuous domain adaptation through graphs,'' in
  \emph{Proceedings of the IEEE/CVF Conference on Computer Vision and Pattern
  Recognition}, 2019, pp. 6568--6577.

\bibitem{xia2021graph}
F.~Xia, K.~Sun, S.~Yu, A.~Aziz, L.~Wan, S.~Pan, and H.~Liu, ``Graph learning:
  {A} survey,'' \emph{CoRR}, vol. abs/2105.00696, 2021.

\bibitem{ji2023drugood}
Y.~Ji, L.~Zhang, J.~Wu, B.~Wu, L.~Li, L.-K. Huang, T.~Xu, Y.~Rong, J.~Ren,
  D.~Xue \emph{et~al.}, ``Drugood: Out-of-distribution dataset curator and
  benchmark for ai-aided drug discovery--a focus on affinity prediction
  problems with noise annotations,'' in \emph{Proceedings of the AAAI
  Conference on Artificial Intelligence}, vol.~37, no.~7, 2023, pp. 8023--8031.

\bibitem{qiu2018deepinf}
J.~Qiu, J.~Tang, H.~Ma, Y.~Dong, K.~Wang, and J.~Tang, ``Deepinf: Social
  influence prediction with deep learning,'' in \emph{Proceedings of the 24th
  {ACM} {SIGKDD} International Conference on Knowledge Discovery {\&} Data
  Mining, {KDD} 2018, London, UK, August 19-23, 2018}, Y.~Guo and F.~Farooq,
  Eds.\hskip 1em plus 0.5em minus 0.4em\relax {ACM}, 2018, pp. 2110--2119.

\bibitem{luo2023normalizing}
L.~Luo, Y.-F. Li, G.~Haffari, and S.~Pan, ``Normalizing flow-based neural
  process for few-shot knowledge graph completion,'' \emph{arXiv preprint
  arXiv:2304.08183}, 2023.

\bibitem{wu2023graph}
\BIBentryALTinterwordspacing
S.~Wu, F.~Sun, W.~Zhang, X.~Xie, and B.~Cui, ``Graph neural networks in
  recommender systems: {A} survey,'' \emph{{ACM} Comput. Surv.}, vol.~55,
  no.~5, pp. 97:1--97:37, 2023. [Online]. Available:
  \url{https://doi.org/10.1145/3535101}
\BIBentrySTDinterwordspacing

\bibitem{zheng2022multi}
X.~Zheng, M.~Zhang, C.~Chen, C.~Li, C.~Zhou, and S.~Pan, ``Multi-relational
  graph neural architecture search with fine-grained message passing,'' in
  \emph{2022 IEEE International Conference on Data Mining (ICDM)}.\hskip 1em
  plus 0.5em minus 0.4em\relax IEEE, 2022, pp. 783--792.

\bibitem{wang2017graph}
\BIBentryALTinterwordspacing
Q.~Wang, Z.~Mao, B.~Wang, and L.~Guo, ``Knowledge graph embedding: {A} survey
  of approaches and applications,'' \emph{{IEEE} Trans. Knowl. Data Eng.},
  vol.~29, no.~12, pp. 2724--2743, 2017. [Online]. Available:
  \url{https://doi.org/10.1109/TKDE.2017.2754499}
\BIBentrySTDinterwordspacing

\bibitem{jin2023dual}
D.~Jin, L.~Wang, Y.~Zheng, G.~Song, F.~Jiang, X.~Li, W.~Lin, and S.~Pan, ``Dual
  intent enhanced graph neural network for session-based new item
  recommendation,'' in \emph{Proceedings of the ACM Web Conference 2023}, 2023,
  pp. 684--693.

\bibitem{yu2018spatio}
B.~Yu, H.~Yin, and Z.~Zhu, ``Spatio-temporal graph convolutional networks: A
  deep learning framework for traffic forecasting,'' in \emph{Proceedings of
  the 27th International Joint Conference on Artificial Intelligence (IJCAI)},
  2018.

\bibitem{li2022out}
H.~Li, X.~Wang, Z.~Zhang, and W.~Zhu, ``Out-of-distribution generalization on
  graphs: A survey,'' \emph{arXiv preprint arXiv:2202.07987}, 2022.

\bibitem{pomeroy2020dynamics}
C.~Pomeroy, R.~M. Bond, P.~J. Mucha, and S.~J. Cranmer, ``Dynamics of social
  network emergence explain network evolution,'' \emph{Scientific Reports},
  vol.~10, no.~1, p. 21876, 2020.

\bibitem{bardoscia2021physics}
M.~Bardoscia, P.~Barucca, S.~Battiston, F.~Caccioli, G.~Cimini,
  D.~Garlaschelli, F.~Saracco, T.~Squartini, and G.~Caldarelli, ``The physics
  of financial networks,'' \emph{Nature Reviews Physics}, vol.~3, no.~7, pp.
  490--507, 2021.

\bibitem{acemoglu2015systemic}
D.~Acemoglu, A.~Ozdaglar, and A.~Tahbaz-Salehi, ``Systemic risk and stability
  in financial networks,'' \emph{American Economic Review}, vol. 105, no.~2,
  pp. 564--608, 2015.

\bibitem{wu2023molecular}
H.~Wu, C.~M. Eckhardt, and A.~A. Baccarelli, ``Molecular mechanisms of
  environmental exposures and human disease,'' \emph{Nature Reviews Genetics},
  vol.~24, no.~5, pp. 332--344, 2023.

\bibitem{ercan2017public}
T.~Ercan, N.~C. Onat, O.~Tatari, and J.-D. Mathias, ``Public transportation
  adoption requires a paradigm shift in urban development structure,''
  \emph{Journal of cleaner production}, vol. 142, pp. 1789--1799, 2017.

\bibitem{alam2018domain}
F.~Alam, S.~Joty, and M.~Imran, ``Domain adaptation with adversarial training
  and graph embeddings,'' in \emph{Proceedings of the 56th Annual Meeting of
  the Association for Computational Linguistics (Volume 1: Long Papers)}, 2018,
  pp. 1077--1087.

\bibitem{wu2019domain}
M.~Wu, S.~Pan, X.~Zhu, C.~Zhou, and L.~Pan, ``Domain-adversarial graph neural
  networks for text classification,'' in \emph{2019 IEEE International
  Conference on Data Mining (ICDM)}.\hskip 1em plus 0.5em minus 0.4em\relax
  IEEE, 2019, pp. 648--657.

\bibitem{chen2022learning}
Y.~Chen, Y.~Zhang, Y.~Bian, H.~Yang, M.~Kaili, B.~Xie, T.~Liu, B.~Han, and
  J.~Cheng, ``Learning causally invariant representations for
  out-of-distribution generalization on graphs,'' \emph{Advances in Neural
  Information Processing Systems}, vol.~35, pp. 22\,131--22\,148, 2022.

\bibitem{li2022learning}
H.~Li, Z.~Zhang, X.~Wang, and W.~Zhu, ``Learning invariant graph
  representations for out-of-distribution generalization,'' in \emph{Advances
  in Neural Information Processing Systems}, 2022.

\bibitem{baranwal2021graph}
A.~Baranwal, K.~Fountoulakis, and A.~Jagannath, ``Graph convolution for
  semi-supervised classification: Improved linear separability and
  out-of-distribution generalization,'' in \emph{International Conference on
  Machine Learning}.\hskip 1em plus 0.5em minus 0.4em\relax PMLR, 2021, pp.
  684--693.

\bibitem{MALTONI201956}
D.~Maltoni and V.~Lomonaco, ``Continuous learning in single-incremental-task
  scenarios,'' \emph{Neural Networks}, vol. 116, pp. 56--73, 2019.

\bibitem{biesialska-etal-2020-continual}
M.~Biesialska, K.~Biesialska, and M.~R. Costa-juss{\`a}, ``Continual lifelong
  learning in natural language processing: A survey,'' in \emph{Proceedings of
  the 28th International Conference on Computational Linguistics}.\hskip 1em
  plus 0.5em minus 0.4em\relax Barcelona, Spain (Online): International
  Committee on Computational Linguistics, Dec. 2020, pp. 6523--6541.

\bibitem{pmlrELLA2013}
\BIBentryALTinterwordspacing
P.~Ruvolo and E.~Eaton, ``{ELLA}: An efficient lifelong learning algorithm,''
  in \emph{Proceedings of the 30th International Conference on Machine
  Learning}, ser. Proceedings of Machine Learning Research, S.~Dasgupta and
  D.~McAllester, Eds., vol.~28, no.~1.\hskip 1em plus 0.5em minus 0.4em\relax
  Atlanta, Georgia, USA: PMLR, 17--19 Jun 2013, pp. 507--515. [Online].
  Available: \url{https://proceedings.mlr.press/v28/ruvolo13.html}
\BIBentrySTDinterwordspacing

\bibitem{DBLP2016}
\BIBentryALTinterwordspacing
A.~A. Rusu, N.~C. Rabinowitz, G.~Desjardins, H.~Soyer, J.~Kirkpatrick,
  K.~Kavukcuoglu, R.~Pascanu, and R.~Hadsell, ``Progressive neural networks,''
  \emph{CoRR}, vol. abs/1606.04671, 2016. [Online]. Available:
  \url{http://arxiv.org/abs/1606.04671}
\BIBentrySTDinterwordspacing

\bibitem{Core502017}
\BIBentryALTinterwordspacing
V.~Lomonaco and D.~Maltoni, ``Core50: a new dataset and benchmark for
  continuous object recognition,'' \emph{CoRR}, vol. abs/1705.03550, 2017.
  [Online]. Available: \url{http://arxiv.org/abs/1705.03550}
\BIBentrySTDinterwordspacing

\bibitem{das2018unsupervised}
D.~Das and C.~G. Lee, ``Unsupervised domain adaptation using regularized
  hyper-graph matching,'' in \emph{2018 25th IEEE International Conference on
  Image Processing (ICIP)}.\hskip 1em plus 0.5em minus 0.4em\relax IEEE, 2018,
  pp. 3758--3762.

\bibitem{zhang2018structural}
Y.~Zhang, S.~Miao, and R.~Liao, ``Structural domain adaptation with latent
  graph alignment,'' in \emph{2018 25th IEEE International Conference on Image
  Processing (ICIP)}.\hskip 1em plus 0.5em minus 0.4em\relax IEEE, 2018, pp.
  3753--3757.

\bibitem{wu2020unsupervised}
M.~Wu, S.~Pan, C.~Zhou, X.~Chang, and X.~Zhu, ``Unsupervised domain adaptive
  graph convolutional networks,'' in \emph{Proceedings of The Web Conference
  2020}, 2020, pp. 1457--1467.

\bibitem{yang2021oodsurvey}
J.~Yang, K.~Zhou, Y.~Li, and Z.~Liu, ``Generalized out-of-distribution
  detection: A survey,'' \emph{arXiv preprint arXiv:2110.11334}, 2021.

\bibitem{yuan2023continual}
Q.~Yuan, S.-U. Guan, P.~Ni, T.~Luo, K.~L. Man, P.~Wong, and V.~Chang,
  ``Continual graph learning: A survey,'' \emph{arXiv preprint
  arXiv:2301.12230}, 2023.

\bibitem{febrinanto2023graph}
F.~G. Febrinanto, F.~Xia, K.~Moore, C.~Thapa, and C.~Aggarwal, ``Graph lifelong
  learning: A survey,'' \emph{IEEE Computational Intelligence Magazine},
  vol.~18, no.~1, pp. 32--51, 2023.

\bibitem{liu2024generalization}
S.~Liu and K.~Ding, ``Beyond generalization: A survey of out-of-distribution
  adaptation on graphs,'' \emph{arXiv:2402.11153}, 2024.

\bibitem{yu2017open-holder38}
Y.~Yu, W.-Y. Qu, N.~Li, and Z.~Guo, ``Open-category classification by
  adversarial sample generation,'' in \emph{International Joint Conference on
  Artificial Intelligence}, 2017, pp. 3357--3363.

\bibitem{luo2020progressive}
Y.~Luo, Z.~Wang, Z.~Huang, and M.~Baktashmotlagh, ``Progressive graph learning
  for open-set domain adaptation,'' in \emph{International Conference on
  Machine Learning}.\hskip 1em plus 0.5em minus 0.4em\relax PMLR, 2020, pp.
  6468--6478.

\bibitem{ma2019gcan}
X.~Ma, T.~Zhang, and C.~Xu, ``Gcan: Graph convolutional adversarial network for
  unsupervised domain adaptation,'' in \emph{Proceedings of the IEEE/CVF
  Conference on Computer Vision and Pattern Recognition}, 2019, pp. 8266--8276.

\bibitem{yang2020heterogeneous}
X.~Yang, C.~Deng, T.~Liu, and D.~Tao, ``Heterogeneous graph attention network
  for unsupervised multiple-target domain adaptation,'' \emph{IEEE Transactions
  on Pattern Analysis and Machine Intelligence}, vol.~44, no.~4, pp.
  1992--2003, 2020.

\bibitem{shen2020network}
X.~Shen, Q.~Dai, S.~Mao, F.-l. Chung, and K.-S. Choi, ``Network together: Node
  classification via cross-network deep network embedding,'' \emph{IEEE
  Transactions on Neural Networks and Learning Systems}, vol.~32, no.~5, pp.
  1935--1948, 2020.

\bibitem{shen2020adversarial}
X.~Shen, Q.~Dai, F.-l. Chung, W.~Lu, and K.-S. Choi, ``Adversarial deep network
  embedding for cross-network node classification,'' in \emph{Proceedings of
  the AAAI conference on artificial intelligence}, vol.~34, no.~03, 2020, pp.
  2991--2999.

\bibitem{shen2023domain}
X.~Shen, S.~Pan, K.-S. Choi, and X.~Zhou, ``Domain-adaptive message passing
  graph neural network,'' \emph{Neural Networks}, vol. 164, pp. 439--454, 2023.

\bibitem{shen2023CNEC}
X.~Shen, M.~Shao, S.~Pan, L.~T. Yang, and X.~Zhou, ``Domain-adaptive graph
  attention-supervised network for cross-network edge classification,''
  \emph{IEEE Transactions on Neural Networks and Learning Systems}, 2023.

\bibitem{DBLP:conf/aaai/0003LNW0S21}
T.~Zhao, Y.~Liu, L.~Neves, O.~Woodford, M.~Jiang, and N.~Shah, ``Data
  augmentation for graph neural networks,'' in \emph{Proceedings of the aaai
  conference on artificial intelligence}, vol.~35, no.~12, 2021, pp.
  11\,015--11\,023.

\bibitem{shen2023neighbor}
X.~Shen, D.~Sun, S.~Pan, X.~Zhou, and L.~T. Yang, ``Neighbor contrastive
  learning on learnable graph augmentation,'' in \emph{Proceedings of the AAAI
  Conference on Artificial Intelligence}, vol.~37, no.~8, 2023, pp. 9782--9791.

\bibitem{feng2020graph}
W.~Feng, J.~Zhang, Y.~Dong, Y.~Han, H.~Luan, Q.~Xu, Q.~Yang, E.~Kharlamov, and
  J.~Tang, ``Graph random neural networks for semi-supervised learning on
  graphs,'' \emph{Advances in neural information processing systems}, vol.~33,
  pp. 22\,092--22\,103, 2020.

\bibitem{mcpherson2001birds}
M.~McPherson, L.~Smith-Lovin, and J.~M. Cook, ``Birds of a feather: Homophily
  in social networks,'' \emph{Annual review of sociology}, vol.~27, no.~1, pp.
  415--444, 2001.

\bibitem{ding2021closer}
M.~Ding, K.~Kong, J.~Chen, J.~Kirchenbauer, M.~Goldblum, D.~Wipf, F.~Huang, and
  T.~Goldstein, ``A closer look at distribution shifts and out-of-distribution
  generalization on graphs,'' in \emph{NeurIPS 2021 Workshop on Distribution
  Shifts: Connecting Methods and Applications}, 2021.

\bibitem{ma2019disentangled}
J.~Ma, P.~Cui, K.~Kuang, X.~Wang, and W.~Zhu, ``Disentangled graph
  convolutional networks,'' in \emph{International conference on machine
  learning}.\hskip 1em plus 0.5em minus 0.4em\relax PMLR, 2019, pp. 4212--4221.

\bibitem{kuang2018stable}
K.~Kuang, P.~Cui, S.~Athey, R.~Xiong, and B.~Li, ``Stable prediction across
  unknown environments,'' in \emph{proceedings of the 24th ACM SIGKDD
  international conference on knowledge discovery \& data mining}, 2018, pp.
  1617--1626.

\bibitem{zhang2021stable}
S.~Zhang, K.~Kuang, J.~Qiu, J.~Yu, Z.~Zhao, H.~Yang, Z.~Zhang, and F.~Wu,
  ``Stable prediction on graphs with agnostic distribution shift,'' \emph{arXiv
  preprint arXiv:2110.03865}, 2021.

\bibitem{hu2019strategies}
W.~Hu, B.~Liu, J.~Gomes, M.~Zitnik, P.~Liang, V.~Pande, and J.~Leskovec,
  ``Strategies for pre-training graph neural networks,'' \emph{arXiv preprint
  arXiv:1905.12265}, 2019.

\bibitem{zhao2020uncertainty}
X.~Zhao, F.~Chen, S.~Hu, and J.-H. Cho, ``Uncertainty aware semi-supervised
  learning on graph data,'' \emph{Advances in Neural Information Processing
  Systems}, vol.~33, pp. 12\,827--12\,836, 2020.

\bibitem{stadler2021graph}
M.~Stadler, B.~Charpentier, S.~Geisler, D.~Z{\"u}gner, and S.~G{\"u}nnemann,
  ``Graph posterior network: Bayesian predictive uncertainty for node
  classification,'' \emph{Advances in Neural Information Processing Systems},
  vol.~34, pp. 18\,033--18\,048, 2021.

\bibitem{huang2022end}
T.~Huang, D.~Wang, Y.~Fang, and Z.~Chen, ``End-to-end open-set semi-supervised
  node classification with out-of-distribution detection,'' in
  \emph{Proceedings of the Thirty-First International Joint Conference on
  Artificial Intelligence, {IJCAI-22}}, L.~D. Raedt, Ed.\hskip 1em plus 0.5em
  minus 0.4em\relax International Joint Conferences on Artificial Intelligence
  Organization, 7 2022, pp. 2087--2093.

\bibitem{liu2023good}
Y.~Liu, K.~Ding, H.~Liu, and S.~Pan, ``Good-d: On unsupervised graph
  out-of-distribution detection,'' in \emph{Proceedings of the Sixteenth ACM
  International Conference on Web Search and Data Mining}, 2023, pp. 339--347.

\bibitem{zhang2023g2pxy}
Q.~Zhang, Z.~Shi, X.~Zhang, X.~Chen, P.~Fournier-Viger, and S.~Pan, ``G2pxy:
  generative open-set node classification on graphs with proxy unknowns,'' in
  \emph{Proceedings of the Thirty-Second International Joint Conference on
  Artificial Intelligence}, 2023, pp. 4576--4583.

\bibitem{zhang2022dynamic}
Q.~Zhang, Q.~Li, X.~Chen, P.~Zhang, S.~Pan, P.~Fournier-Viger, and J.~Z. Huang,
  ``A dynamic variational framework for open-world node classification in
  structured sequences,'' in \emph{2022 IEEE International Conference on Data
  Mining (ICDM)}.\hskip 1em plus 0.5em minus 0.4em\relax IEEE, 2022, pp.
  703--712.

\bibitem{wu2020openwgl}
M.~Wu, S.~Pan, and X.~Zhu, ``Openwgl: Open-world graph learning,'' in
  \emph{2020 IEEE international conference on data mining (icdm)}.\hskip 1em
  plus 0.5em minus 0.4em\relax IEEE, 2020, pp. 681--690.

\bibitem{wu2021towards}
Q.~Wu, C.~Yang, and J.~Yan, ``Towards open-world feature extrapolation: An
  inductive graph learning approach,'' \emph{Advances in Neural Information
  Processing Systems}, vol.~34, pp. 19\,435--19\,447, 2021.

\bibitem{mancini2022learning}
M.~Mancini, M.~F. Naeem, Y.~Xian, and Z.~Akata, ``Learning graph embeddings for
  open world compositional zero-shot learning,'' \emph{IEEE Transactions on
  pattern analysis and machine intelligence}, 2022.

\bibitem{shi2018open}
B.~Shi and T.~Weninger, ``Open-world knowledge graph completion,'' in
  \emph{Proceedings of the AAAI conference on artificial intelligence},
  vol.~32, no.~1, 2018.

\bibitem{das2020probabilistic}
R.~Das, A.~Godbole, N.~Monath, M.~Zaheer, and A.~McCallum, ``Probabilistic
  case-based reasoning for open-world knowledge graph completion,'' in
  \emph{Findings of the Association for Computational Linguistics: EMNLP 2020},
  2020, pp. 4752--4765.

\bibitem{niu2021open}
L.~Niu, C.~Fu, Q.~Yang, Z.~Li, Z.~Chen, Q.~Liu, and K.~Zheng, ``Open-world
  knowledge graph completion with multiple interaction attention,'' \emph{World
  Wide Web}, vol.~24, pp. 419--439, 2021.

\bibitem{wang2022lifelong}
C.~Wang, Y.~Qiu, D.~Gao, and S.~Scherer, ``Lifelong graph learning,'' in
  \emph{Proceedings of the IEEE/CVF conference on computer vision and pattern
  recognition}, 2022, pp. 13\,719--13\,728.

\bibitem{Hedegaard2023continual}
L.~Hedegaard, N.~Heidari, and A.~Iosifidis, ``Continual spatio-temporal graph
  convolutional networks,'' \emph{Pattern Recogn.}, vol. 140, no.~C, may 2023.

\bibitem{Zhang2023continual}
P.~Zhang, Y.~Yan, C.~Li, S.~Wang, X.~Xie, G.~Song, and S.~Kim, ``Continual
  learning on dynamic graphs via parameter isolation,'' in \emph{Proceedings of
  the 46th International ACM SIGIR Conference on Research and Development in
  Information Retrieval}, ser. SIGIR '23.\hskip 1em plus 0.5em minus
  0.4em\relax New York, NY, USA: Association for Computing Machinery, 2023, p.
  601–611.

\bibitem{kou-etal-2020-disentangle}
X.~Kou, Y.~Lin, S.~Liu, P.~Li, J.~Zhou, and Y.~Zhang, ``{D}isentangle-based
  {C}ontinual {G}raph {R}epresentation {L}earning,'' in \emph{Proceedings of
  the 2020 Conference on Empirical Methods in Natural Language Processing
  (EMNLP)}.\hskip 1em plus 0.5em minus 0.4em\relax Online: Association for
  Computational Linguistics, Nov. 2020, pp. 2961--2972.

\bibitem{Liu2021OCFGN}
H.~Liu, Y.~Yang, and X.~Wang, ``Overcoming catastrophic forgetting in graph
  neural networks,'' in \emph{Proceedings of the AAAI conference on artificial
  intelligence}, vol.~35, no.~10, 2021, pp. 8653--8661.

\bibitem{xu2020GraphSAIL}
Y.~Xu, Y.~Zhang, W.~Guo, H.~Guo, R.~Tang, and M.~Coates, ``Graphsail: Graph
  structure aware incremental learning for recommender systems,'' in
  \emph{Proceedings of the 29th ACM International Conference on Information \&
  Knowledge Management}.\hskip 1em plus 0.5em minus 0.4em\relax Association for
  Computing Machinery, 2020, p. 2861–2868.

\bibitem{Sun_Ye_Peng_Wang_Yu_2023}
L.~Sun, J.~Ye, H.~Peng, F.~Wang, and P.~S. Yu, ``Self-supervised continual
  graph learning in adaptive riemannian spaces,'' \emph{Proceedings of the AAAI
  Conference on Artificial Intelligence}, vol.~37, no.~4, pp. 4633--4642, 2023.

\bibitem{cui2023lifelong}
Y.~Cui, Y.~Wang, Z.~Sun, W.~Liu, Y.~Jiang, K.~Han, and W.~Hu, ``Lifelong
  embedding learning and transfer for growing knowledge graphs,'' in
  \emph{AAAI}, 2023.

\bibitem{zhou2021overcoming}
F.~Zhou and C.~Cao, ``Overcoming catastrophic forgetting in graph neural
  networks with experience replay,'' in \emph{Proceedings of the AAAI
  Conference on Artificial Intelligence}, vol.~35, no.~5, 2021, pp. 4714--4722.

\bibitem{Galke2020TGDS}
L.~Galke, I.~Vagliano, and A.~Scherp, ``Incremental training of graph neural
  networks on temporal graphs under distribution shift,'' \emph{CoRR}, vol.
  abs/2006.14422, 2020.

\bibitem{Perini2022LSGER}
M.~Perini, G.~Ramponi, P.~Carbone, and V.~Kalavri, ``Learning on streaming
  graphs with experience replay,'' in \emph{Proceedings of the 37th ACM/SIGAPP
  Symposium on Applied Computing}, 2022, pp. 470--478.

\bibitem{Ahrabian2021}
K.~Ahrabian, Y.~Xu, Y.~Zhang, J.~Wu, Y.~Wang, and M.~Coates, ``Structure aware
  experience replay for incremental learning in graph-based recommender
  systems,'' in \emph{Proceedings of the 30th ACM International Conference on
  Information \& Knowledge Management}, 2021, pp. 2832--2836.

\bibitem{Xu2021TST}
X.~Chen, J.~Wang, and K.~Xie, ``Trafficstream: {A} streaming traffic flow
  forecasting framework based on graph neural networks and continual
  learning,'' \emph{CoRR}, vol. abs/2106.06273, 2021.

\bibitem{Anam2023cmdgat}
A.~Zaman, F.~Yangyu, M.~S. Ayub, M.~Irfan, L.~Guoyun, and L.~Shiya, ``Cmdgat:
  Knowledge extraction and retention based continual graph attention network
  for point cloud registration,'' \emph{Expert Syst. Appl.}, vol. 214, no.~C,
  mar 2023.

\bibitem{Li2016Lwf}
Z.~Li and D.~Hoiem, ``Learning without forgetting,'' in \emph{Computer Vision
  -- ECCV 2016}, B.~Leibe, J.~Matas, N.~Sebe, and M.~Welling, Eds.\hskip 1em
  plus 0.5em minus 0.4em\relax Cham: Springer International Publishing, 2016,
  pp. 614--629.

\bibitem{James2017OCF}
J.~Kirkpatrick, R.~Pascanu, N.~Rabinowitz, J.~Veness, G.~Desjardins, A.~A.
  Rusu, K.~Milan, J.~Quan, T.~Ramalho, A.~Grabska-Barwinska, D.~Hassabis,
  C.~Clopath, D.~Kumaran, and R.~Hadsell, ``Overcoming catastrophic forgetting
  in neural networks,'' \emph{Proceedings of the National Academy of Sciences},
  vol. 114, no.~13, pp. 3521--3526, 2017.

\bibitem{Rebuffi2016iCaRLIC}
S.-A. Rebuffi, A.~Kolesnikov, G.~Sperl, and C.~H. Lampert, ``icarl: Incremental
  classifier and representation learning,'' \emph{2017 IEEE Conference on
  Computer Vision and Pattern Recognition (CVPR)}, pp. 5533--5542, 2016.

\bibitem{Lopezpaz2017GEM}
\BIBentryALTinterwordspacing
D.~Lopez{-}Paz and M.~Ranzato, ``Gradient episodic memory for continuum
  learning,'' \emph{CoRR}, vol. abs/1706.08840, 2017. [Online]. Available:
  \url{http://arxiv.org/abs/1706.08840}
\BIBentrySTDinterwordspacing

\bibitem{Maltoni2018CL}
\BIBentryALTinterwordspacing
D.~Maltoni and V.~Lomonaco, ``Continuous learning in single-incremental-task
  scenarios,'' \emph{CoRR}, vol. abs/1806.08568, 2018. [Online]. Available:
  \url{http://arxiv.org/abs/1806.08568}
\BIBentrySTDinterwordspacing

\bibitem{Lee2017LLDE}
J.~Lee, J.~Yoon, E.~Yang, and S.~J. Hwang, ``Lifelong learning with dynamically
  expandable networks,'' \emph{CoRR}, vol. abs/1708.01547, 2017.

\bibitem{sliwoski2014computational}
G.~Sliwoski, S.~Kothiwale, J.~Meiler, and E.~W. Lowe, ``Computational methods
  in drug discovery,'' \emph{Pharmacological reviews}, vol.~66, no.~1, pp.
  334--395, 2014.

\bibitem{gao2020deep}
W.~Gao, S.~P. Mahajan, J.~Sulam, and J.~J. Gray, ``Deep learning in protein
  structural modeling and design,'' \emph{Patterns}, vol.~1, no.~9, 2020.

\bibitem{jumper2021highly}
J.~Jumper, R.~Evans, A.~Pritzel, T.~Green, M.~Figurnov, O.~Ronneberger,
  K.~Tunyasuvunakool, R.~Bates, A.~{\v{Z}}{\'\i}dek, A.~Potapenko
  \emph{et~al.}, ``Highly accurate protein structure prediction with
  alphafold,'' \emph{Nature}, vol. 596, no. 7873, pp. 583--589, 2021.

\bibitem{lim2019predicting}
J.~Lim, S.~Ryu, K.~Park, Y.~J. Choe, J.~Ham, and W.~Y. Kim, ``Predicting
  drug--target interaction using a novel graph neural network with 3d
  structure-embedded graph representation,'' \emph{Journal of chemical
  information and modeling}, vol.~59, no.~9, pp. 3981--3988, 2019.

\bibitem{karimi2019deepaffinity}
M.~Karimi, D.~Wu, Z.~Wang, and Y.~Shen, ``Deepaffinity: interpretable deep
  learning of compound--protein affinity through unified recurrent and
  convolutional neural networks,'' \emph{Bioinformatics}, vol.~35, no.~18, pp.
  3329--3338, 2019.

\bibitem{morselli2021network}
D.~Morselli~Gysi, {\'I}.~Do~Valle, M.~Zitnik, A.~Ameli, X.~Gan, O.~Varol, S.~D.
  Ghiassian, J.~Patten, R.~A. Davey, J.~Loscalzo \emph{et~al.}, ``Network
  medicine framework for identifying drug-repurposing opportunities for
  covid-19,'' \emph{Proceedings of the National Academy of Sciences}, vol. 118,
  no.~19, p. e2025581118, 2021.

\bibitem{ji2021survey}
S.~Ji, S.~Pan, E.~Cambria, P.~Marttinen, and S.~Y. Philip, ``A survey on
  knowledge graphs: Representation, acquisition, and applications,'' \emph{IEEE
  transactions on neural networks and learning systems}, vol.~33, no.~2, pp.
  494--514, 2021.

\bibitem{wang2023survey}
J.~Wang, B.~Wang, M.~Qiu, S.~Pan, B.~Xiong, H.~Liu, L.~Luo, T.~Liu, Y.~Hu,
  B.~Yin \emph{et~al.}, ``A survey on temporal knowledge graph completion:
  Taxonomy, progress, and prospects,'' \emph{arXiv preprint arXiv:2308.02457},
  2023.

\bibitem{pan2023unifying}
S.~Pan, L.~Luo, Y.~Wang, C.~Chen, J.~Wang, and X.~Wu, ``Unifying large language
  models and knowledge graphs: A roadmap,'' \emph{IEEE Transactions on
  Knowledge and Data Engineering}, 2023.

\bibitem{luo2023reasoning}
L.~Luo, Y.-F. Li, G.~Haffari, and S.~Pan, ``Reasoning on graphs: Faithful and
  interpretable large language model reasoning,'' \emph{arXiv preprint
  arXiv:2310.01061}, 2023.

\bibitem{luo2023chatrule}
L.~Luo, J.~Ju, B.~Xiong, Y.-F. Li, G.~Haffari, and S.~Pan, ``Chatrule: Mining
  logical rules with large language models for knowledge graph reasoning,''
  \emph{arXiv preprint arXiv:2309.01538}, 2023.

\bibitem{jin2022neural}
M.~Jin, Y.-F. Li, and S.~Pan, ``Neural temporal walks: Motif-aware
  representation learning on continuous-time dynamic graphs,'' \emph{Advances
  in Neural Information Processing Systems}, vol.~35, pp. 19\,874--19\,886,
  2022.

\bibitem{jin2022multivariate}
M.~Jin, Y.~Zheng, Y.-F. Li, S.~Chen, B.~Yang, and S.~Pan, ``Multivariate time
  series forecasting with dynamic graph neural odes,'' \emph{IEEE Transactions
  on Knowledge and Data Engineering}, 2022.

\bibitem{zheng2023towards}
X.~Zheng, Y.~Liu, Z.~Bao, M.~Fang, X.~Hu, A.~W.-C. Liew, and S.~Pan, ``Towards
  data-centric graph machine learning: Review and outlook,'' \emph{arXiv
  preprint arXiv:2309.10979}, 2023.

\bibitem{guo2023data}
Y.~Guo, C.~Yang, Y.~Chen, J.~Liu, C.~Shi, and J.~Du, ``A data-centric framework
  to endow graph neural networks with out-of-distribution detection ability,''
  2023.

\bibitem{wu2021handling}
Q.~Wu, H.~Zhang, J.~Yan, and D.~Wipf, ``Handling distribution shifts on graphs:
  An invariance perspective,'' in \emph{International Conference on Learning
  Representations}, 2021.

\bibitem{sui2023unleashing}
Y.~Sui, Q.~Wu, J.~Wu, Q.~Cui, L.~Li, J.~Zhou, X.~Wang, and X.~He, ``Unleashing
  the power of graph data augmentation on covariate distribution shift,'' in
  \emph{Thirty-seventh Conference on Neural Information Processing Systems},
  2023.

\bibitem{zheng2023gnnevaluator}
X.~Zheng, M.~Zhang, C.~Chen, S.~Molaei, C.~Zhou, and S.~Pan, ``Gnnevaluator:
  Evaluating gnn performance on unseen graphs without labels,'' in
  \emph{Thirty-seventh Conference on Neural Information Processing Systems},
  2023.

\bibitem{ektefaie2023multimodal}
Y.~Ektefaie, G.~Dasoulas, A.~Noori, M.~Farhat, and M.~Zitnik, ``Multimodal
  learning with graphs,'' \emph{Nature Machine Intelligence}, vol.~5, no.~4,
  pp. 340--350, 2023.

\bibitem{yoon2023multimodal}
M.~Yoon, J.~Y. Koh, B.~Hooi, and R.~Salakhutdinov, ``Multimodal graph learning
  for generative tasks,'' 2023.

\bibitem{guo2023visual}
Y.~Guo, F.~Yin, X.-h. Li, X.~Yan, T.~Xue, S.~Mei, and C.-L. Liu, ``Visual
  traffic knowledge graph generation from scene images,'' in \emph{Proceedings
  of the IEEE/CVF International Conference on Computer Vision}, 2023, pp.
  21\,604--21\,613.

\bibitem{wang2023knowledge}
Y.~Wang, N.~Lipka, R.~A. Rossi, A.~Siu, R.~Zhang, and T.~Derr, ``Knowledge
  graph prompting for multi-document question answering,'' \emph{arXiv preprint
  arXiv:2308.11730}, 2023.

\bibitem{huang2023prodigy}
Q.~Huang, H.~Ren, P.~Chen, G.~Kr{\v{z}}manc, D.~Zeng, P.~Liang, and
  J.~Leskovec, ``Prodigy: Enabling in-context learning over graphs,''
  \emph{arXiv preprint arXiv:2305.12600}, 2023.

\bibitem{zhang2022trustworthy}
H.~Zhang, B.~Wu, X.~Yuan, S.~Pan, H.~Tong, and J.~Pei, ``Trustworthy graph
  neural networks: Aspects, methods and trends,'' \emph{arXiv preprint
  arXiv:2205.07424}, 2022.

\bibitem{wu2023graphguard}
B.~Wu, H.~Zhang, X.~Yang, S.~Wang, M.~Xue, S.~Pan, and X.~Yuan, ``Graphguard:
  Detecting and counteracting training data misuse in graph neural networks,''
  \emph{arXiv preprint arXiv:2312.07861}, 2023.

\bibitem{zhang2023demystifying}
H.~Zhang, B.~Wu, S.~Wang, X.~Yang, M.~Xue, S.~Pan, and X.~Yuan, ``Demystifying
  uneven vulnerability of link stealing attacks against graph neural
  networks,'' in \emph{International Conference on Machine Learning}.\hskip 1em
  plus 0.5em minus 0.4em\relax PMLR, 2023, pp. 41\,737--41\,752.

\bibitem{zhang2022projective}
H.~Zhang, X.~Yuan, C.~Zhou, and S.~Pan, ``Projective ranking-based gnn evasion
  attacks,'' \emph{IEEE Transactions on Knowledge and Data Engineering}, 2022.

\bibitem{ying2019gnnexplainer}
Z.~Ying, D.~Bourgeois, J.~You, M.~Zitnik, and J.~Leskovec, ``Gnnexplainer:
  Generating explanations for graph neural networks,'' \emph{Advances in neural
  information processing systems}, vol.~32, 2019.

\bibitem{liu2023towards}
Y.~Liu, K.~Ding, Q.~Lu, F.~Li, L.~Y. Zhang, and S.~Pan, ``Towards
  self-interpretable graph-level anomaly detection,'' in \emph{Thirty-seventh
  Conference on Neural Information Processing Systems}, 2023.

\bibitem{jin2023survey}
D.~Jin, L.~Wang, H.~Zhang, Y.~Zheng, W.~Ding, F.~Xia, and S.~Pan, ``A survey on
  fairness-aware recommender systems,'' \emph{arXiv preprint arXiv:2306.00403},
  2023.

\bibitem{zhang2023interaction}
H.~Zhang, X.~Yuan, Q.~V.~H. Nguyen, and S.~Pan, ``On the interaction between
  node fairness and edge privacy in graph neural networks,'' \emph{arXiv
  preprint arXiv:2301.12951}, 2023.

\end{thebibliography}

\end{document}